\documentclass{article}

\PassOptionsToPackage{numbers, compress}{natbib}

 \usepackage[main,final]{neurips_2026}

\usepackage[utf8]{inputenc} %
\usepackage[T1]{fontenc}    %
\usepackage{hyperref}       %
\usepackage{url}            %
\usepackage{booktabs}       %
\usepackage{amsfonts}       %
\usepackage{nicefrac}       %
\usepackage{microtype}      %
\usepackage{xcolor}         %

\definecolor{linkblue}{RGB}{30,70,140}
\hypersetup{
    colorlinks=true,
    allcolors=linkblue
}

\usepackage{graphicx}
\usepackage{amsmath}
\usepackage{cleveref}
\usepackage{threeparttable}
\usepackage{wrapfig}
\usepackage[export]{adjustbox}

\usepackage[varqu,varl,scaled=0.95]{inconsolata} %
\usepackage{upquote}                             %
\usepackage{fvextra}
\usepackage[most]{tcolorbox}

\ifdefined\pdfgentounicode                       %
  \input{glyphtounicode}\pdfgentounicode=1
\fi

\DefineVerbatimEnvironment{PromptVerb}{Verbatim}{
  fontsize=\small,
  breaklines, breakanywhere,
  breaksymbolleft={}, breakindent=0pt, breakautoindent=false,
}

\newcounter{prompt}             %
\crefname{prompt}{Prompt}{Prompts}
\Crefname{prompt}{Prompt}{Prompts}

\newcommand{\promptheading}[1]{%
  \textbf{Prompt~\theprompt}\ifx\relax#1\relax\else\ #1\fi}

\newenvironment{prompt}[2]      %
  {\refstepcounter{prompt}%
   \ifx\relax#2\relax\else\label{#2}\fi
   \VerbatimEnvironment
   \begin{tcolorbox}[enhanced, breakable,
     colback=gray!3, colframe=gray!45,
     colbacktitle=gray!12, coltitle=black,
     boxrule=0.4pt, arc=1.5pt,
     titlerule=0.4pt, fonttitle=\normalfont,
     title={\promptheading{#1}}]
   \begin{PromptVerb}}
  {\end{PromptVerb}\end{tcolorbox}}

\newcommand{\unif}{\textsc{Copilot-Uniform}}
\newcommand{\thumbs}{\textsc{Copilot-Thumbs}}
\newcommand{\iwa}[1]{\emph{#1}}
\newcommand{\wa}[1]{\iwa{#1}}

\newcommand{\appcref}[1]{%
    \Cref{#1}%
}

\newcommand{\Appcref}[1]{%
    \Cref{#1}%
}

\title{Working with AI: Measuring the Applicability of Generative AI to Occupations}

\author{%
  Kiran Tomlinson \\
  Microsoft Research 
  \And
  Sonia Jaffe \\
  Microsoft Research 
  \AND
  Will Wang \\
  Microsoft Research 
  \And
  Scott Counts \\
  Microsoft
  \And
  Siddharth Suri \\
  Microsoft Research 
}

\begin{document}

\maketitle

\begin{abstract}
With generative AI emerging as a general-purpose technology, understanding its economic effects is among society’s most pressing questions. Earlier work on AI impact largely relied on predictions of AI capabilities or focused narrowly on individual firms. Drawing instead on real-world AI usage, we analyze a dataset of 200k anonymized conversations with Microsoft Bing Copilot to measure AI applicability to occupations. We use an LLM-based pipeline to classify the O*NET work activities assisted or performed by AI in each conversation. We find that the most common and successful AI-assisted work activities involve information work---the creation, processing, and communication of information. At the occupation level, we find widespread AI applicability cutting across sectors, as most occupations have information work components. Our methodology also allows us to predict which occupations are more likely to delegate tasks to AI and which are more likely to use AI to assist existing workflows. Finally, we see an increase in occupation-level AI applicability over our 9-month data period, driven primarily by people using AI for more varied tasks.

AI applicability scores and component usage and success metrics are available at \url{https://github.com/microsoft/working-with-ai}.
\end{abstract}

\section{Introduction}

General purpose technologies~\cite{bresnahan1995general}, such as the steam engine and the computer, have historically been strong drivers of economic growth, impacting a broad range of sectors. 
In the last several years, generative artificial intelligence (AI) has come to the fore as the next  general purpose technology~\cite{goldfarb2023could,eloundou2024gpts}, capable of improving or speeding up tasks as varied as medical diagnosis~\cite{mcduff2025towards} and software development~\cite{cui2024effects}. These capabilities are reflected in the astounding rate of AI adoption: nearly 40\% of Americans  reported using generative AI in late 2024, outpacing the early diffusion of the personal computer and the internet~\cite{bick2024rapid}. 
Given this widespread adoption and potential for economic impact, a crucial question is \emph{which} work activities AI is most useful for and which occupations perform them, as this will shape where any productivity benefits of AI fall.

We answer these questions by identifying the work activities performed in a dataset of 200k real-world conversations with a mainstream large language model (LLM)-powered generative AI system, Microsoft Bing Copilot (now Microsoft Copilot) from January -- September 2024.
To identify the occupations where AI might be most useful, we draw from a common economic methodology~\cite{autor2003skill} that decomposes an occupation into its common tasks and aggregates task-level measurements to occupation-level metrics \cite{ eloundou2024gpts, shao2025futureworkaiagents, felten2023will, brynjolfsson2018machines,frey2017employment,manyika2017future,tolan2017measure,felten2018ai,web2020impact}. As is standard in this research area, we use the O*NET database~\cite{onet29} developed under U.S.\ Department of Labor sponsorship, which decomposes occupations hierarchically into their work activities. We develop an LLM-based pipeline to map the \emph{user goals} and \emph{AI actions} in each conversation to the set of work activities being performed. Separating user goals (what task is the user trying to accomplish?) from AI actions (what action does the AI take to assist the user?) allows us to identify which occupations are more likely to delegate tasks to AI and which are more likely to engage collaboratively with AI.  Additionally, we measure how successfully different work activities are assisted or performed by AI, using both explicit user feedback (thumbs up and down) and a task completion classifier. To distinguish between broad and narrow AI contributions towards work activities, we also classify the scope of AI capability towards each work activity matched to a conversation. 
From these classifications, we compute an \emph{AI applicability score} for each occupation. This score captures if there is a non-trivial amount of AI usage that successfully completes activities relevant to an occupation.

\paragraph{Results.} 
Bing Copilot usage data reveals that AI is most often and most successfully used for information work, that is, activities involving creating, gathering, and disseminating information. These activities are particularly relevant to computer and mathematical, sales, and administrative support occupations, resulting in high AI applicability scores for those groups. Since information work tasks are so widespread, most occupations have some overlap with work activities for which there is non-trivial AI usage. Our separate analysis of user goals and AI actions indicates AI may be used in more assistive capacities in occupational groups with larger physical work components (like food preparation and installation, maintenance, and repair) and may directly perform more tasks in media and financial operations occupations, changing the task distribution of those occupations. 
Furthermore, we see AI usage increasing in variety over time: from the first to last month of our data, the number of user goals covering 75\% of activity increases from 20 to 28, resulting in increasing AI applicability estimates.
Comparing to earlier predictions of occupational AI exposure by \citet{eloundou2024gpts}, our AI applicability scores reveal similar trends (employment-weighted $r = 0.73$ between the measures), but suggest they underestimated the relevance of AI to certain information work occupations like proofreaders and software developers. Our analysis also finds only a weak correlation between AI applicability and occupational wage (employment-weighted $r = 0.13$), challenging narratives that AI exposure is strongly associated with higher-wage occupations~\cite{felten2023will,eloundou2024gpts}.

\section{Related work}
Our work connects to research predicting occupational AI impact using various proxies for AI capabilities, including expert forecasts \cite{eloundou2024gpts, shao2025futureworkaiagents}, progress on AI benchmarks~\cite{felten2023will}, AI patents~\cite{septiandri2024potential}, and surveys of workers paired with AI predictions~\cite{gmyrek2025generative}. Before the advent of generative AI, similar techniques were used to predict the occupational impacts of robotics, machine learning, and computerization more broadly~\cite{autor2003skill,frey2017employment,manyika2017future,brynjolfsson2018machines,tolan2017measure,felten2018ai,web2020impact,felten2021occupational}.
We contribute to this literature by analyzing actual human--LLM conversations, identifying the work activities being assisted or performed and, by extension, the applicability of AI to associated occupations.

The studies most similar to ours are the Anthropic Economic Index~\cite{handa2025economic} and Google's AI \& Economy ATLAS~\cite{iscenko2026googlesaieconomy}, which analyze Claude and Gemini usage by mapping conversations (or usage clusters) to the O*NET taxonomy. 
These studies (and follow-up Anthropic Economic Index reports~\cite{appelmccrorytamkin2025geoapi,anthropic2026aeiv5,anthropic2026economicprimitives,massenkoffmccrory2026labor}) map AI usage to occupation-specific O*NET \emph{tasks},  limiting identification of cross-occupation AI capabilities.
Our research goal is to go beyond usage to an estimate of occupational applicability of AI. Thus, 
we exhaustively classify a conversation into all relevant O*NET \emph{work activities}, which apply across occupations. In addition, our measures of conversational success allow us to approximate the relative efficacy of AI across work activities, and our user goal vs.\ AI action distinction provides further insight into which work activities are being assisted or performed by AI. 
 Following an initial release of our paper, subsequent iterations of the Anthropic Economic Index line of work incorporated some of these ideas, including measures of success~\cite{anthropic2026economicprimitives} and occupation-level exposure scores~\cite{massenkoffmccrory2026labor}.
Other recent work analyzing ChatGPT data~\cite{chatterji2025chatgpt,chatterji2026organizationsuseaievidence} adopted our approach of classifying by work activities, but does not use them to measure relevance for occupations. 
 
Another prominent line of related work involves AI benchmarks for measuring performance on occupational tasks~\cite{patwardhan2026gdpval,jimenez2024swebench,miserendino2025swelancer,xu2025theagentcompany,drouin2024workarena,vishwakarma-etal-2025-llms}. These provide more thorough evaluations of success than our proxy measures, but on hand-crafted sets of tasks that may not represent the true distribution of AI usage. In a meta-analysis, \citet{wang2026doesagentdevelopmentreflect} argue that these benchmarks tend to skew towards software development and do not represent the full spectrum of work. On the methods side, our multi-step classification pipeline is similar to \texttt{Infer-Retrieve-Rank}~\citep{doosterlinck2024extreme}.

\section{Data}
\subsection{Bing Copilot conversations}

We analyze anonymized and privacy-scrubbed U.S. conversation data from Microsoft Bing Copilot (henceforth, Copilot) gathered from January 1, 2024 to September 30, 2024. We focus only on conversations in the United States to align with occupation and work activity information from O*NET. 
Our main dataset, \unif{}, is a uniform sample of approximately 100k conversations, 
providing a representative view of what tasks users perform with a mainstream, publicly available, free-to-use generative AI chatbot. 
We use a supporting dataset, \thumbs{}, of 100k conversations uniformly sampled from those that received at least one thumbs up or thumbs down reaction from the user. This dataset allows us to investigate what activities are performed more or less successfully, as measured by explicit user feedback. Due to the potential for strong selection bias in user--task pairs for which  we observe thumbs feedback, we perform most of our analysis on \unif{} and use \thumbs{} only for its user feedback signal. 
Our use of Copilot data was reviewed and approved by the Microsoft IRB (under ID \# 11028).
While we cannot release conversational data due to legal and privacy obligations, all of our aggregated work-activity- and occupation-level metrics are publicly available at \url{https://github.com/microsoft/working-with-ai}.

\subsection{O*NET and Bureau of Labor Statistics data}
We use the O*NET 29.0 Database~\citep{onet29}, which provides a hierarchical decomposition of occupations into their tasks and work activities.  
Each of the ${\sim} 1$k O*NET occupations is mapped to a set of job-specific \emph{tasks} (${\sim} 19$k total), which are each mapped to a set of cross-occupational \emph{detailed work activities} (DWAs; ${\sim} 2$k total). DWAs are clustered into \emph{intermediate work activities} (IWAs; 332 total), which are in turn clustered into \emph{generalized work activities} (GWAs; 41 total). We chose IWAs as the appropriate level of granularity for our analysis, balancing enough specificity with cross-occupational generality. We combine O*NET with data on wages and employment from the 2023 Occupational Employment and Wage Statistics (OEWS) data published by the U.S.\ Bureau of Labor Statistics (BLS)~\cite{oews2023}. We combine OEWS data with task frequency and relevance data from O*NET to get a rough estimate of how often each IWA is done by the U.S.~labor force (details in \appcref{app:real-world-freq}). We also use wage, education, and demographic data from  the Current Population Survey (CPS) data from 2024~\cite{CPS_BLS}, which is gathered by the U.S.\ Census Bureau for the BLS. All of our occupational analysis is performed on occupations from the 2018 Standard Occupational Classification (SOC)~\cite{BLS_SOC_2018}. \Appcref{app:onet-bls} describes how we map the O*NET and CPS occupational taxonomies to SOC codes and \appcref{app:mising-data} describes which SOC codes we drop due to missing data.

\section{Methods}\label{sec:methods}

\subsection{Work activity classification}
For each conversation in our datasets, we use an LLM classification pipeline to identify \emph{all} IWAs that match the user goal and the AI action. In most conversations, there are multiple IWA matches on both sides (on average, about 3 user goal IWAs and 6 AI action IWAs out of 332 total; see~\Cref{fig:per-convo-matches}). If the user goal or AI action is not related to any work activity, then it should  match with zero IWAs.

Our pipeline consists of two stages of LLM prompts. In the first-stage prompts, we give GPT-4o the entire conversation and ask it to separately summarize the user goal and the AI action in the style of an O*NET IWA, as well as four rewordings of each statement (for these ``generate'' prompts, we use default temperature 1; the allowed range is $[0, 2]$).\footnote{GPT-4o very likely includes O*NET data in its pretraining corpus, as it exhibits strong knowledge of O*NET~\cite{achananuparp2026multi}.} We then use these summaries to sort all IWA titles in order of relevance to the user goal and AI action (creating two rankings) through cosine similarity of their OpenAI \texttt{text-embedding-3-large} embeddings. More specifically, we sort by average similarity between IWA title embeddings and the five alternate phrasings of the LLM-generated summaries to average out differences caused by word choice rather than meaning.

In the second stage, we use GPT-4o to do a binary classification for every IWA as to whether it matches the user goal and/or AI action in the conversation (for these ``classify'' prompts, we use temperature 0). The user and AI classifications are done in separate sequences of prompts, starting by classifying the 20 IWAs ranked highest in stage one and each successive prompt classifying the next 20 highest-ranked IWAs. In validation against human labels, we found that GPT-4o could perform 20 IWA classifications in one prompt without degrading accuracy, but that more led to worse classification; we also found that grouping IWAs by level of similarity as described led to higher classification reliability. As another measure to improve agreement between human and LLM labels, we provide the first GPT-4o-generated summary from stage one as an additional ``IWA'' in each prompt, which serves as a point of reference against which other IWAs are measured. Compared to alternative approaches (e.g., hierarchical classification~\cite{handa2025economic}), our pipeline sacrifices efficiency for thoroughness.

We validated work activity classifications against human annotations for 195 conversations, finding human--human inter-rater reliability $\kappa \approx 0.51$ and human--LLM pipeline $\kappa \approx 0.40$ on a held-out test set not used for the pipeline tuning described above (due to the extreme sparsity of labels, the overall agreement of human and LLM labels is well over 90\%). These moderate $\kappa$ values are driven by the unclear boundaries between IWAs; for many instances, there is no clear ``right answer.''  All prompts, model details, and human validation details are provided in \appcref{app:iwa-and-scope}.

\subsubsection{Activity share and coverage}
Since each conversation can be assigned multiple IWAs, instead of counting raw IWA occurrences in our conversation data, we focus on each IWA's \emph{activity share}, where we allocate an equal fraction of each conversation to its labeled IWAs, separately on the user and AI sides. 
We consider the work activities that appear at non-trivial frequencies in \unif{} to be \emph{covered}. We define non-trivial usage using a threshold of 0.05\% activity share and use this as a signal that AI can potentially assist or perform that IWA. We chose the threshold 0.05\% to minimize the number of occupations with no or all IWAs covered, thereby maximizing the usefulness of the measure for relative comparisons between occupations (\Cref{fig:threshold-selection}A). The ordering of occupations induced by our AI applicability score is robust to the chosen coverage threshold (\Cref{fig:threshold-selection}B). We define an occupation's \emph{coverage} to be the weighted fraction of its work activities that are covered (using the occupation--IWA weights defined in \Cref{sec:ai-app-score}; see \Cref{fig:coverage-by-threshold,fig:soc-coverage-hists}).

\subsubsection{Metrics of success: thumbs feedback, completion, and scope}
We use multiple measures of the success of a conversation and what it demonstrates about Copilot's ability to assist with or perform work. The most direct measure of success comes directly from users: for each IWA, we define its \emph{positive feedback share} as the fraction of conversations mapping to that IWA in \thumbs{} for which the feedback was positive. This metric can take values from 0 (thumbs were always down) to 1 (thumbs were always up).

We also use a conversation-level LLM completion classification (details and prompts in \appcref{app:completion}), to measure each IWA's \emph{completion rate}, the share of conversations labeled with that IWA where the AI completed the user's task in the conversation (as estimated by the classifier). IWA completion rates are highly correlated with IWA positive feedback shares (weighted $r > 0.75$; \Cref{fig:thumbs-vs-completion}), and have the advantage of being available for all conversations, so they are not subject to the (potentially strong) selection bias of thumbs feedback. This high correlation acts as validation of the completion classifier's ability to measure real signal, but it is an imperfect proxy for true task completion.\footnote{\citet{hicke2026adoptneqadaptlongitudinal} performed human validation of our completion classifier on WildChat~\cite{wildchat}, finding human--LLM agreement ($\kappa = 0.29$, $0.12$) in the same range as human--human agreement ($\kappa = 0.25$), although both were low.}

For each relevant IWA in a conversation, we also perform an LLM classification of the fraction of work in the IWA that Copilot demonstrates the ability to assist or perform, which we call the \emph{scope}, measured on a six-point ordinal scale: none, minimal, limited, moderate, significant, complete. The goal of scope is to distinguish between cases where Copilot assists with a large or small fraction of the work in an IWA. Scope classification and validation are detailed in \appcref{app:iwa-and-scope}. To aggregate these ordinal ratings, we measure the fraction of conversations mapping to an IWA where the scope rating is moderate or higher, which we call the \emph{scope share}.

\subsection{Occupational AI applicability score}\label{sec:ai-app-score}
We aggregate the IWA-level measures into an \emph{AI applicability score} for each occupation in the Standard Occupational Classification (SOC)~\cite{BLS_SOC_2018}. 
The score combines whether Copilot assists or performs an occupation's associated work activities (frequency $\geq0.05\%$) successfully (completion rate) and in ways that cover a broad share of the work activity (scope $\geq$ moderate). We also use task-level ratings in O*NET to weight each IWA in an occupation by importance. We compute AI applicability score separately for user goals and AI actions, averaging the two when we need a single summary metric. 
Treating all work activities with activity share of at least 0.05\% in the same way mitigates potential distributional bias from considering only consumer Copilot data. 
However, using a frequency threshold means relative comparisons are more meaningful than absolute scores, as scores depend strongly on the threshold (\Cref{fig:depth-vs-threshold}) but relative comparisons are robust (\Cref{fig:threshold-selection}B).

\paragraph{AI applicability score definition.}
Formally, the user goal AI applicability score is 
\begin{equation}\label{eq:ai-score}
    a_i^{\text{user}} = \sum_{j \in \text{IWAs}(i)}    \mathbf{1}[f_j^\text{user} \ge 0.0005]  c_j^\text{user}  s_j^\text{user} w_{ij},
\end{equation}
where IWAs$(i)$ is the set of IWAs performed by occupation $i$,  $f_j^\text{user}$ 
is the user goal activity share of $j$,  $c_j^\text{user}${} is the task completion rate of conversations with IWA $j$ as a user goal,  $s_j^\text{user}${} is the fraction of conversations with user goal $j$ in which the scope classification is moderate or higher, and $w_{ij}$ is a weight derived from O*NET data which captures how important IWA $j$ is to occupation $i$ based on task-level ratings (see Appendix \ref{app:iwa_weights} for more details).
We define $a_i^\text{AI}$ similarly for AI actions, using AI action IWA measures. For $a_i^\text{AI}$ only, we account for the fact that some IWAs map to both nonphysical and physical tasks, the latter of which  a text-based AI cannot itself perform. To do this, we adjust the occupation--IWA weights based on a physical task classifier, replacing $w_{ij}$ with $w_{ij}^\text{nonphys}$ (see Appendix \ref{app:iwa_weights} for more details). We report $a_i = (a_i^\text{user} + a_i^\text{AI})/2$ unless otherwise specified. 
We find that coverage is the primary contributor to AI applicability scores, but both completion and scope provide meaningful adjustments (\Cref{app:ai-app-components}).

\section{Results}
\begin{figure*}[h]
    \centering
\includegraphics[width=0.47\linewidth]{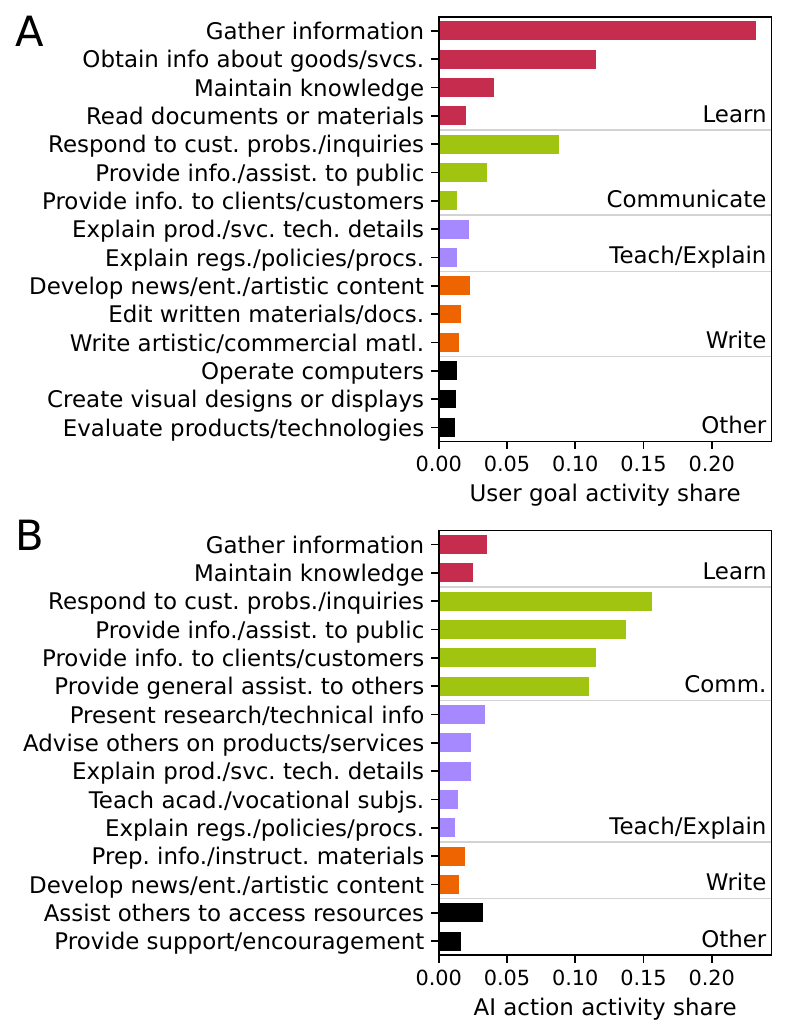}
\includegraphics[width=0.515\linewidth]{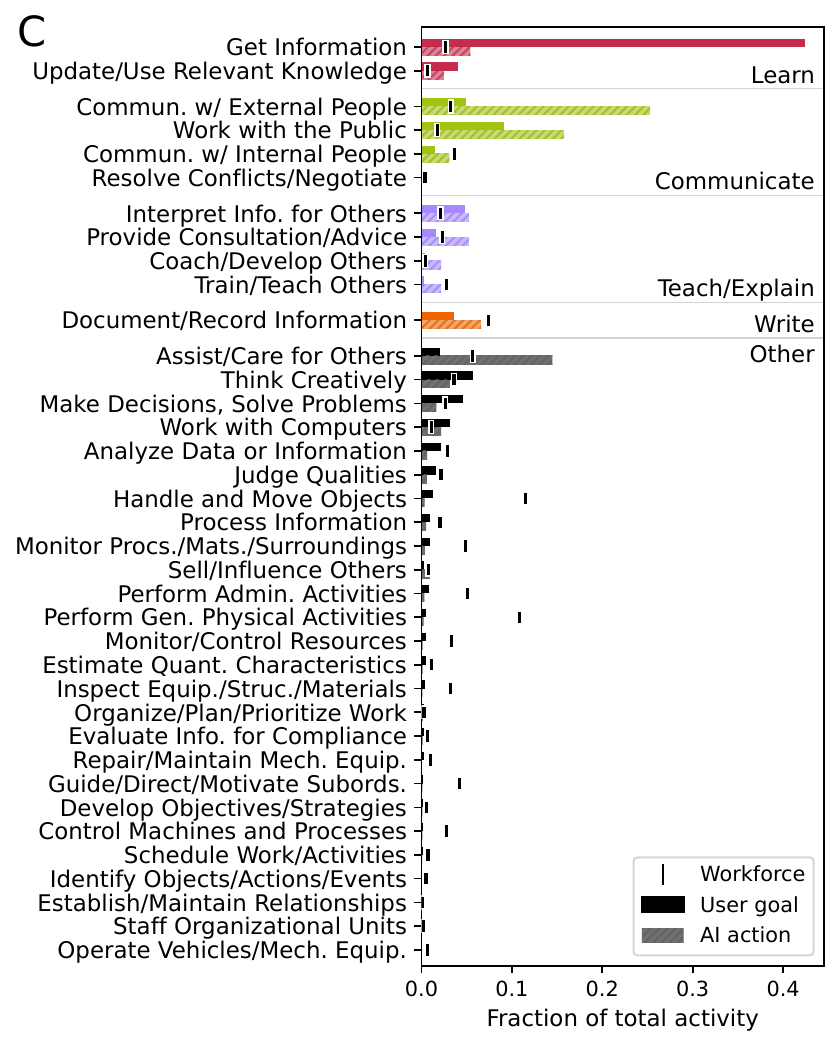}
    \caption{\textbf{Frequency of work activities in Bing Copilot usage.}
    (\textbf{A}) The 15 IWAs that occur most frequently as user goals.
    (\textbf{B}) The 15 IWAs that occur most frequently as AI actions.
    (\textbf{C}) The frequencies of GWAs as user goals and AI actions, compared to a rough estimate of GWA workforce frequency (described in \appcref{app:real-world-freq}).
    All panels show researcher-generated IWA groupings.}
    \label{fig:iwas}  
\end{figure*}

\subsection{Work activities}

The most common Copilot user goals fall into four broad categories: learning, communicating, teaching/explaining, and writing---with learning and communicating being the most prominent (\Cref{fig:iwas}A).
The IWAs reflected in AI actions tell a complementary story.  
\Cref{fig:iwas}B shows that the AI plays a service role: common IWA verbs include \iwa{Provide}, \iwa{Explain}, \iwa{Teach}, \iwa{Assist}, and \iwa{Respond}. A majority  of AI-side activity is devoted to communicating and teaching/explaining information to the user (\Cref{fig:user-bot-counts} and \Cref{tab:user-bot-ratios} highlight additional differences between user goals and AI actions).

The conversation-level asymmetry between user goals and AI actions is pronounced: these sets of IWAs are disjoint in 40\% of conversations, and in 96\%,  there are more IWAs unique to a side than in common.
Furthermore,  
for every work activity matched to a user goal the AI performs an average of two work activities in service of the user's request (\Cref{fig:per-convo-matches}).

Recall that IWAs are grouped into \emph{generalized work activities} (GWAs); we use GWAs to identify how Copilot usage compares to the overall frequencies of work activities across the U.S.\ workforce. 
\Cref{fig:iwas}C shows the activity shares in Copilot aggregated to GWAs and a roughly estimated fraction of total U.S.\ labor falling under each GWA (described in \appcref{app:real-world-freq}).  
GWAs focused on learning and communication are most prevalent, echoing the results in \Cref{fig:iwas}A,B.
The GWAs more prevalent in Copilot usage than in the workforce include \wa{Getting Information}, \wa{Updating and Using Knowledge}, \wa{Communicate with External People}, and \wa{Interpreting Information for Others}.
These align with \emph{information work}~\cite{huvila2009ecological,huvila2016information}, concerning the creation, processing, and communication of information, and the related concept of \emph{knowledge work}, which typically refers to work requiring non-routine and complex problem-solving \cite{drucker1959landmarks,pyoria2005concept,mckercher2008knowledge, reinhardt2011knowledge, suri2024use}.  
Our data show that the user goals that people use Copilot for, and the activities AI performs in service of those goals, center around information work, and these activities occur disproportionately often relative to the fraction of information work in the workforce.

\subsubsection{Metrics of success}

\begingroup 
\setlength{\intextsep}{0pt}
\begin{wrapfigure}{r}{0.45\textwidth}
    \centering
    \includegraphics[width=0.45\textwidth]{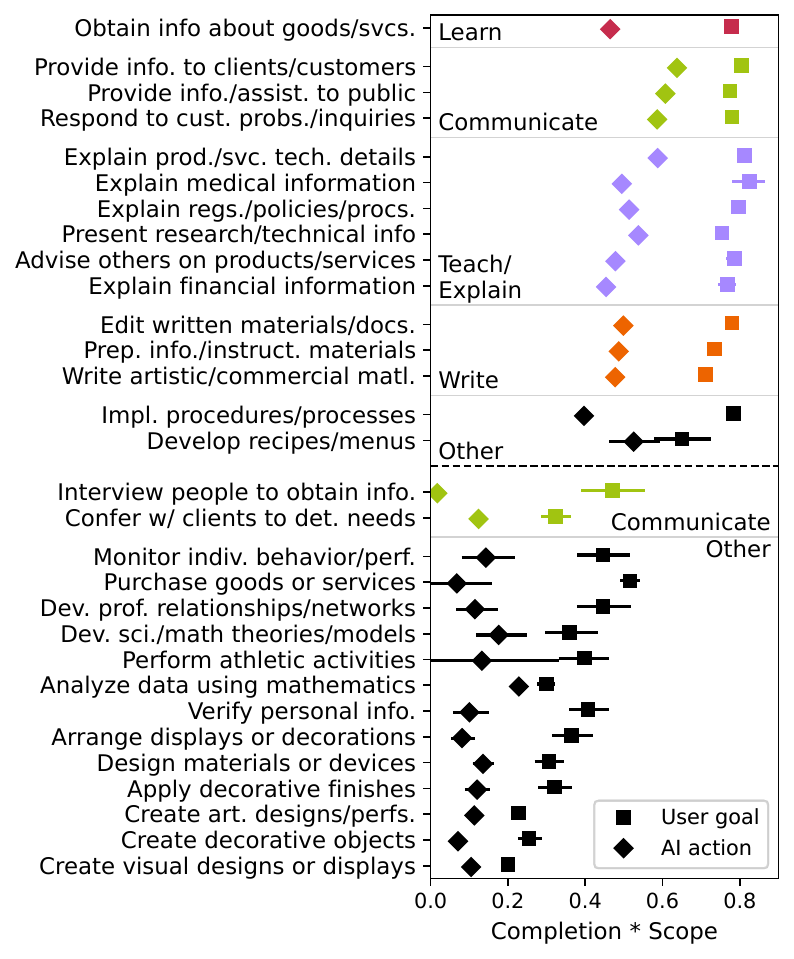}
    \caption{\textbf{Success of Bing Copilot work activities.} The top and bottom 15 IWAs (with user goal or AI action activity share at least 0.05\%) by the product of their completion rate and the share of conversations where Bing Copilot demonstrated at least a moderate scope of capability.}
    \label{fig:success-metrics}
\end{wrapfigure}

Beyond identifying which activities generative AI is being used for, we also analyze how successfully it performs and assists different work activities using our three measures of success: task completion, user feedback, and scope. We focus primarily on completion and scope due to concerns about selection bias in thumbs feedback, although the patterns observed in user feedback are similar. 
\Cref{fig:success-metrics} shows the top and bottom IWAs ranked by the product of completion rate and scope share. 
Copilot performs best at activities related to communicating, teaching/explaining, and writing, largely corresponding to the most frequent AI actions in \Cref{fig:iwas}B. 
On the other hand, it performs worse at image generation and data analysis tasks (these patterns are confirmed by thumbs feedback; \Cref{fig:thumb-rate-top-bottom-iwas,fig:thumb-rate-top-bottom-gwas}). 
Scope in particular is highly correlated  with (log) share of user activity ($r = 0.64$; \Cref{fig:scope-vs-activity-share}). 
This suggests that users have discovered and are exploiting the informational tasks where AI can have broadest impact. Conversely, we see lower correlation between scope and completion (weighted $r = 0.45$ for user goals and $r = 0.22$ for AI actions; \Cref{fig:scope-vs-completion}), indicating that they are measuring different dimensions, as intended. 
AI action completion $\times$ scope is consistently lower than on the user goal side in \Cref{fig:success-metrics}, indicating that AI can help users with a broader fraction of their work than it can perform directly.  
\par
\endgroup

\subsection{AI applicability to occupations}

\begin{figure}[h]
    \centering
    \includegraphics[width=0.9\linewidth]{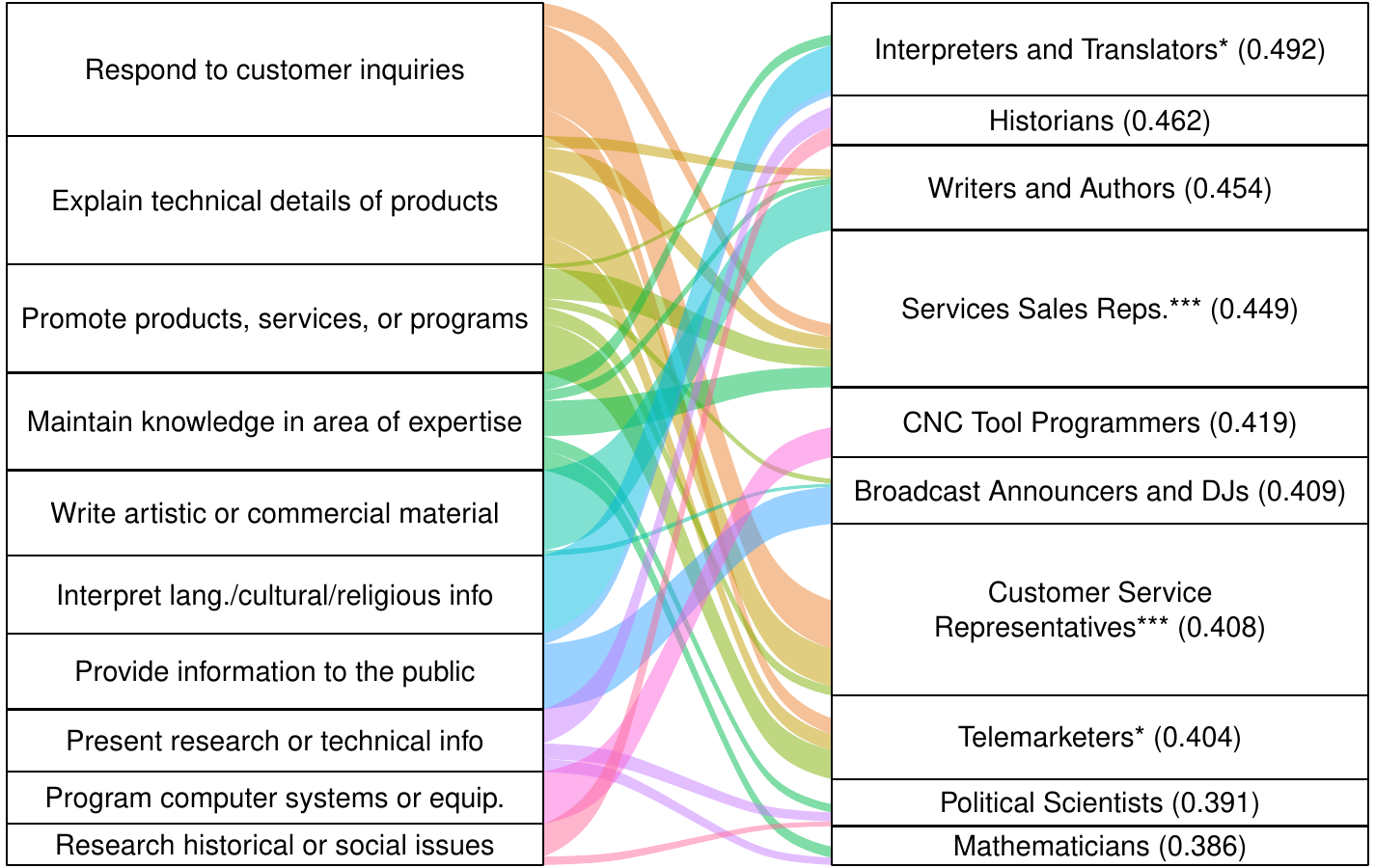}
    \caption{\textbf{Top occupations by AI applicability score and their contributing IWAs.} The 10 SOC occupations with the highest AI applicability scores (right) along with the 10 IWAs that provide the greatest contributions to those scores (left). Asterisks indicate 2023 U.S.\ occupational employment (none: up to 50k, one: 50k to 100k, two: 100k to 500k, three: more than 500k); the heights of each stratum indicate the same. Occupations' applicability scores are in parentheses, and occupations are sorted by their score in decreasing order. 
    Portions of the occupational strata not connected to an IWA by a colored flow represent IWAs not present in the Figure that still contribute to the occupation's applicability score. Both occupation and IWA titles have been shortened for space.
    }
    \label{fig:sankey}
\end{figure}

The use of AI for information work activities observed in \Cref{fig:iwas} translates directly to the occupations where AI is most applicable.
\Cref{fig:sankey} shows the 10 occupations with the highest AI applicability score and the work activities that contribute the most to those occupations' scores (Figure \ref{fig:sankey25} shows the analogous figure with the top 25 occupations). 
The top IWAs involve the creation (e.g.,~\iwa{Write commercial material, Program computer systems}), gathering (e.g.,~\iwa{Maintain knowledge}), and dissemination (e.g.,~\iwa{Respond to customer inquiries, Explain technical details}) of information. In fact, almost all of the top-contributing IWAs fall into one or more of these categories.
These IWAs contribute to the AI applicability scores of occupations involving information creation and processing, such as Interpreters and Translators, Writers and Authors, and Mathematicians.
Occupations involving the communication and dissemination of information also have high AI applicability, such as Sales and Customer Service Representatives.

\begin{figure*}[h]
    \centering
    \includegraphics[width=0.49\linewidth,valign=t]{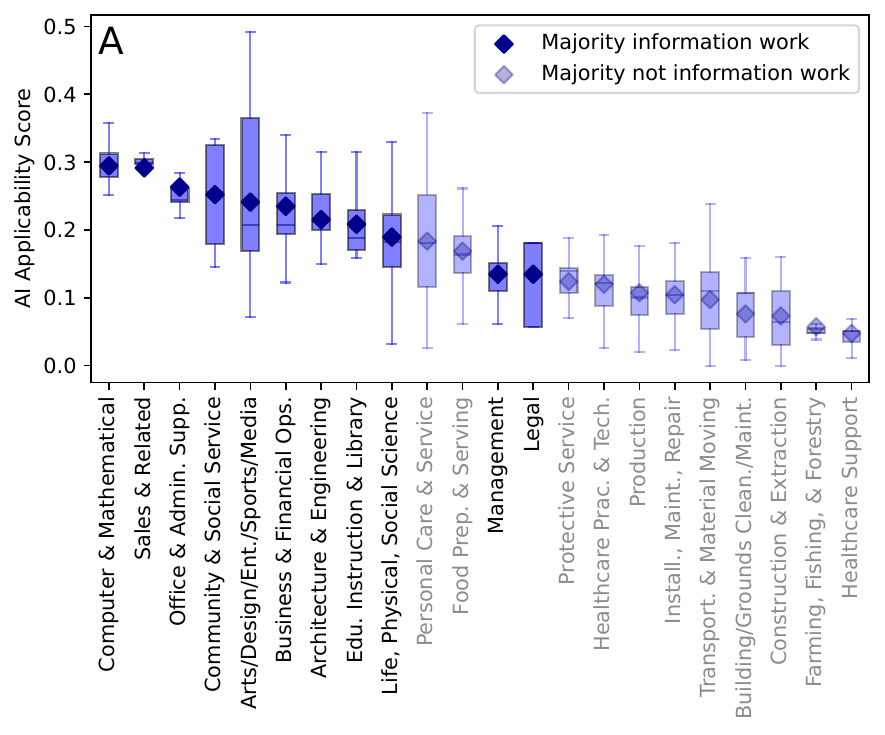}
    \includegraphics[width=0.49\linewidth,valign=t]{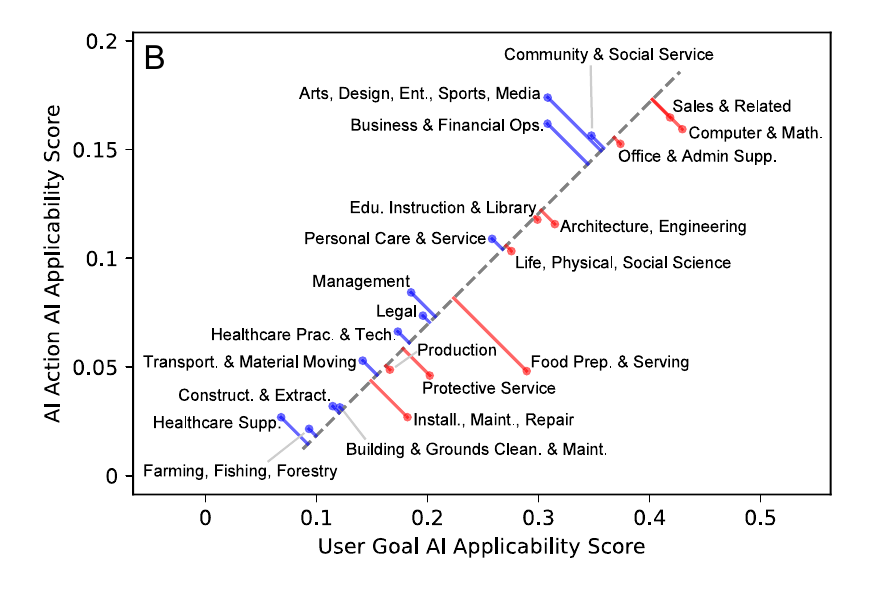}
    \caption{\textbf{AI applicability of occupational groups.}
    \textbf{(A)} Employment-weighted boxplots of major group overall AI applicability (average of user goal and AI action scores), highlighting high scores for information work.
    (\textbf{B}) Scatterplot of each major occupation group's (employment-weighted) average AI action applicability score and average user goal applicability score. The dashed gray line shows the first principal component, capturing overall applicability. The blue and red lines show the extent to which a major group has higher applicability from one metric versus the other.
    }   
    \label{fig:soc-major-applicability}
\end{figure*}

To get a broader view of all occupations in the economy, 
\Cref{fig:soc-major-applicability}A reports the AI applicability of all SOC major groups (clusters of occupations), highlighting groups containing a majority of workers in information work occupations (\appcref{app:information-work} describes how we classify occupations as information work). Similarly, \Cref{tab:soc-minor-group-all} shows the AI applicability scores of all SOC minor groups. The top SOC major groups are predominantly information work, with Computer \& Mathematical, Sales, and Office \& Administrative Support at the very top.
The groups with the lowest scores include occupations that require physically working with people, operating machinery, and other manual labor. However, we find that most occupations have at least some AI applicability, reflecting that most work has some information processing component~\cite{huvila2009ecological}.

\begin{table}[h]
\begin{threeparttable}
\centering
        \caption{\textbf{SOC minor groups by AI applicability score.} }
        \label{tab:soc-minor-group-all}
\tiny
\setlength{\tabcolsep}{2pt}
\renewcommand{\arraystretch}{0.6}
\begin{tabular}{lr|lr|lr}
\toprule
\bfseries Minor Group Title (Abbr.) &\bfseries   Score & \bfseries  Minor Group Title (Abbr.) & \bfseries  Score & \bfseries  Minor Group Title (Abbr.) &\bfseries  Score \\
\midrule
\color{black}Media and Communication Workers\textsuperscript{*} & 0.38 & \color{gray}Food and Beverage Serving Workers\textsuperscript{**} & 0.20 & \color{gray}Woodworkers\textsuperscript{} & 0.12 \\
\color{black}Sales Representatives, Services\textsuperscript{**} & 0.35 & \color{black}Other Educational and Library Workers\textsuperscript{*} & 0.19 & \color{black}Top Executives\textsuperscript{**} & 0.11 \\
\color{black}Information and Record Clerks\textsuperscript{**} & 0.33 & \color{gray}Material Recording and Distrib. Workers\textsuperscript{**} & 0.19 & \color{black}Construction and Extraction Sups.\textsuperscript{*} & 0.11 \\
\color{black}Mathematical Science Occupations\textsuperscript{} & 0.32 & \color{black}Prim., Second., and Special Ed. Teachers\textsuperscript{**} & 0.18 & \color{gray}Assemblers and Fabricators\textsuperscript{*} & 0.11 \\
\color{black}Tour and Travel Guides\phantom{\textsuperscript{*}} & 0.32 & \color{black}Media and Comms. Equipment Workers\textsuperscript{} & 0.18 & \color{gray}Printing Workers\textsuperscript{} & 0.11 \\
\color{black}Postsecondary Teachers\textsuperscript{*} & 0.31 & \color{black}Advertising, Mktg., PR, and Sales Managers\textsuperscript{*} & 0.18 & \color{gray}Health Technologists and Technicians\textsuperscript{**} & 0.10 \\
\color{black}Sales Reps., Wholesale and Manufacturing\textsuperscript{*} & 0.31 & \color{gray}Air Transportation Workers\textsuperscript{} & 0.18 & \color{gray}Metal Workers and Plastic Workers\textsuperscript{*} & 0.10 \\
\color{black}Communications Equipment Operators\textsuperscript{} & 0.30 & \color{black}Lawyers, Judges, and Related Workers\textsuperscript{*} & 0.17 & \color{gray}Vehicle and Mobile Equipment Mechs.\textsuperscript{*} & 0.10 \\
\color{black}Baggage Porters, Bellhops, and Concierges\textsuperscript{} & 0.30 & \color{black}Transportation and Material Moving Sups.\textsuperscript{*} & 0.17 & \color{gray}Other Maintenance Workers\textsuperscript{**} & 0.10 \\
\color{gray}Retail Sales Workers\textsuperscript{**} & 0.30 & \color{black}Architects, Surveyors, and Cartographers\textsuperscript{} & 0.17 & \color{gray}Water Transportation Workers\textsuperscript{} & 0.09 \\
\color{black}Other Sales and Related Workers\textsuperscript{} & 0.30 & \color{gray}Electrical and Electronic Equipment Mechs.\textsuperscript{} & 0.17 & \color{gray}Other Production Occupations\textsuperscript{**} & 0.08 \\
\color{black}Computer Occupations\textsuperscript{**} & 0.29 & \color{gray}Other Healthcare Pracs. and Tech. Occs.\textsuperscript{} & 0.16 & \color{gray}Bldg. Cleaning and Pest Ctrl. Workers\textsuperscript{**} & 0.08 \\
\color{black}Personal Care and Service Sups.\textsuperscript{} & 0.27 & \color{black}Sports Entertainers and Related\textsuperscript{*} & 0.16 & \color{gray}Firefighting and Prevention Workers\textsuperscript{} & 0.07 \\
\color{gray}Entertainment Attendants and Related\textsuperscript{*} & 0.27 & \color{gray}Other Personal Care and Service Workers\textsuperscript{*} & 0.16 & \color{gray}Protective Service Sups.\textsuperscript{} & 0.07 \\
\color{black}Religious Workers\textsuperscript{} & 0.26 & \color{gray}Cooks and Food Preparation Workers\textsuperscript{**} & 0.15 & \color{gray}Material Moving Workers\textsuperscript{**} & 0.07 \\
\color{black}Social Scientists and Related\textsuperscript{} & 0.26 & \color{gray}Funeral Service Workers\textsuperscript{} & 0.14 & \color{gray}Construction Trades Workers\textsuperscript{**} & 0.07 \\
\color{black}Librarians, Curators, and Archivists\textsuperscript{} & 0.25 & \color{black}Other Management Occupations\textsuperscript{**} & 0.14 & \color{gray}Personal Appearance Workers\textsuperscript{*} & 0.06 \\
\color{black}Counselors, Social Workers, and Related\textsuperscript{**} & 0.25 & \color{gray}Building, Grounds, and Maintenance Sups.\textsuperscript{} & 0.14 & \color{gray}Agricultural Workers\textsuperscript{} & 0.06 \\
\color{black}Supervisors of Production Workers\textsuperscript{*} & 0.25 & \color{gray}Other Protective Service Workers\textsuperscript{*} & 0.14 & \color{gray}Other Construction and Related Workers\textsuperscript{} & 0.06 \\
\color{black}Other Office and Admin. Support Workers\textsuperscript{**} & 0.25 & \color{black}Operations Specialties Managers\textsuperscript{**} & 0.14 & \color{black}Legal Support Workers\textsuperscript{} & 0.06 \\
\color{black}Office and Administrative Support Sups.\textsuperscript{*} & 0.25 & \color{black}Occupational Health and Safety Specialists\textsuperscript{} & 0.14 & \color{gray}Other Healthcare Support Occupations\textsuperscript{*} & 0.06 \\
\color{black}Financial Clerks\textsuperscript{**} & 0.24 & \color{gray}Supervisors of Sales Workers\textsuperscript{*} & 0.14 & \color{gray}Helpers, Construction Trades\textsuperscript{} & 0.06 \\
\color{black}Secretaries and Administrative Assistants\textsuperscript{**} & 0.24 & \color{gray}Law Enforcement Workers\textsuperscript{*} & 0.13 & \color{black}Farming, Fishing, and Forestry Sups.\textsuperscript{} & 0.05 \\
\color{black}Business Operations Specialists\textsuperscript{**} & 0.24 & \color{black}Life, Phys., and Social Science Tech'ns\textsuperscript{} & 0.13 & \color{gray}Occupational and Physical Therapy Assts.\textsuperscript{} & 0.05 \\
\color{gray}Animal Care and Service Workers\textsuperscript{} & 0.24 & \color{gray}Motor Vehicle Operators\textsuperscript{**} & 0.13 & \color{gray}Textile, Apparel, and Furnishings Workers\textsuperscript{} & 0.05 \\
\color{black}Financial Specialists\textsuperscript{**} & 0.23 & \color{gray}Healthcare Diagnosing or Treating Pracs.\textsuperscript{**} & 0.13 & \color{gray}Extraction Workers\textsuperscript{} & 0.05 \\
\color{black}Other Teachers and Instructors\textsuperscript{*} & 0.23 & \color{black}Installation, Maintenance, and Repair Sups.\textsuperscript{*} & 0.13 & \color{gray}Home Health Aides and Nursing Assts.\textsuperscript{**} & 0.04 \\
\color{black}Engineers\textsuperscript{*} & 0.22 & \color{gray}Rail Transportation Workers\textsuperscript{} & 0.12 & \color{gray}Grounds Maintenance Workers\textsuperscript{*} & 0.04 \\
\color{black}Physical Scientists\textsuperscript{} & 0.21 & \color{gray}Other Food Prep. and Serving Workers\textsuperscript{*} & 0.12 & \color{gray}Plant and System Operators\textsuperscript{} & 0.04 \\
\color{black}Drafters and Eng. and Mapping Tech'ns\textsuperscript{*} & 0.21 & \color{gray}Food Preparation and Serving Sups.\textsuperscript{*} & 0.12 & \color{gray}Forest and Conservation Workers\textsuperscript{} & 0.03 \\
\color{black}Life Scientists\textsuperscript{} & 0.20 & \color{gray}Food Processing Workers\textsuperscript{*} & 0.12 & \color{black}\textsuperscript{} &  \\
\color{black}Art and Design Workers\textsuperscript{*} & 0.20 & \color{gray}Other Transportation Workers\textsuperscript{} & 0.12 & \color{black}\textsuperscript{} &  \\
\bottomrule 
\end{tabular}
\begin{tablenotes}
    \item  \tiny{Score is the employment-weighted average AI applicability score for each specific occupation in the SOC minor group, averaging the mean of the user goal and AI action scores. Asterisks indicate U.S.\ employment (none: 0 to 500k, one: 500k to 2M, two: greater than 2M). Minor groups in black have a majority of workers performing information work (classified as described in \appcref{app:information-work}); minor groups in gray have only a minority. Titles have been abbreviated for space (Sups: Supervisors, Pracs: Practitioners, Tech'ns: Technicians).}
\end{tablenotes}

\end{threeparttable}
\end{table}

Our separation of user goals from AI actions allows us to not only identify where AI is most useful, but also differentiate how AI may change the way work is done. 
There are at least two different ways occupations could change as a result of AI: (1) delegating some tasks to AI, freeing up workers to focus on different parts of the job, corresponding to high AI action applicability, and (2) performing the same tasks but in collaboration with AI,  corresponding to high user goal applicability score.

\Cref{fig:soc-major-applicability}B plots the average AI-action and user-goal applicability scores for each major SOC group, using principal component analysis to highlight the occupational groups that lean towards either side of this split. As the user goal and AI scores are strongly correlated, the first PC captures overall applicability, while the second measures skew towards AI actions or user goals. The lengths of the red (resp.\ blue) lines indicate the extent to which the occupations in that group are more associated with user goals (resp.\ AI actions); these are projections onto the second principal component (PC) of the applicability scores.   This figure indicates that media and financial operations occupations are relatively likelier to delegate tasks to AI. Conversely, to the extent that AI is used for more physical occupations like those in the food preparation and serving category, it is likely to be in a more assistive role. \Cref{fig:tripartite} highlights which work activities contribute most to this skew.

At a high level, our observations of AI applicability based on real-world usage mostly align with predictions of AI capabilities.
Eloundou et al.~\cite{eloundou2024gpts} used six annotators to predict the share of tasks within each occupation for which access to an LLM would lead to time savings of 50\% or more. The employment-weighted occupation-level correlation between their predictions and our AI applicability score is $r = 0.73$ (\Cref{fig:prediction}).
But our data do not fully confirm expectations: for instance, 
we find only a weak relationship between AI applicability score and wage (employment-weighted $r = 0.13$; \Cref{fig:wages-education}A), while previous work found strong positive correlations~\cite{felten2023will,eloundou2024gpts} (some of this may be due to a lack of employment-weighting in prior work; see \appcref{app:employment-weighting}). This challenges the narrative that AI is only relevant to high-paying white-collar occupations. Similarly, we find a broad range of AI applicability across educational requirements (\Cref{fig:wages-education}B). 
Given these differences, and the widespread usage of the predictive scores of \citet{eloundou2024gpts} in research and policy-making, we advocate for these downstream applications to use our data-driven measurements of AI applicability as a real-world complement.\footnote{Since our initial paper release, the BLS has begun using AI applicability scores as part of their AI exposure categories~\cite{BLS_AI_exposure}.} 
Beyond wage and education, AI may have differential effects across demographic groups to the extent that some demographics are  more or less likely to be employed in AI-applicable occupations (\Cref{fig:demographics-ai-app}).

\begin{figure}
    \centering
    \includegraphics[width=\linewidth]{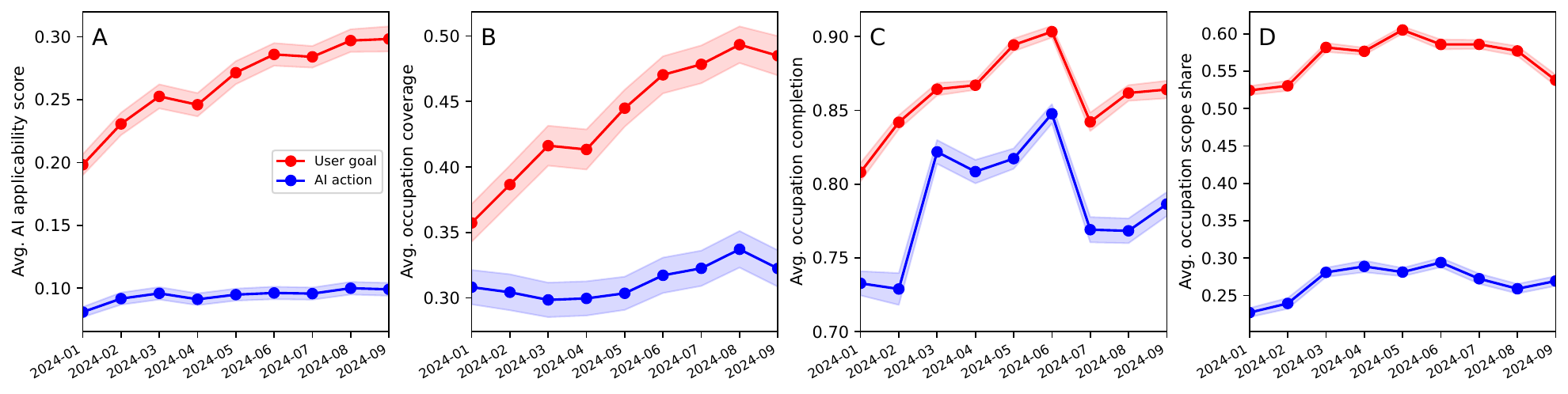}
    \caption{\textbf{Monthly trends for occupation-level metrics.} (\textbf{A}) Average AI applicability scores. (\textbf{B}) Average occupational coverage.  (\textbf{C}) Average occupation-aggregated completion. (\textbf{D}) Average occupation-aggregated scope share. All four metrics are employment-weighted averages over occupations, with IWA-level metrics in B--D aggregated to occupations via weights $w_{ij}$. The shaded regions show 95\% confidence intervals for the weighted mean.
    }
    \label{fig:monthly}
\end{figure}

\subsection{Temporal trends}
We explore how our occupation-level metrics changed over the nine-month data period. \Cref{fig:monthly}A shows AI applicability score, particularly on the user goal side, increasing substantially and almost monotonically. \Cref{fig:monthly}B--D show the components that make up AI applicability score: coverage, completion, and scope. These reveal that the increase in AI applicability score is driven primarily by users performing a wider breadth of tasks with Copilot (hence increasing coverage), accompanied by a smaller increase in completion, while scope remained relatively flat. This increase in diversity of AI usage is clearly visible in both the user goal and AI action monthly IWA distributions, which we see increasing in entropy by about 0.5 bits over the 9 months ($\approx 40\%$ increase in perplexity) and becoming more tail-heavy (\Cref{fig:monthly-coverage-entropy,fig:monthly-share-ratio}). This trend of AI usage diversifying over time is consistent with similar recent trends in Claude usage~\cite{anthropic2026aeiv5}. This is likely driven by changes in the user pool over time, as another analysis of Bing Copilot data over the same period shows that individual users reduce task diversity over time~\cite{hicke2026adoptneqadaptlongitudinal}. 
Conversation-level metrics show the same patterns as the occupation-aggregated metrics in \Cref{fig:monthly} (\Cref{fig:monthly-convo-level}).

\section{Discussion}
This work provides a signal indicating where and how AI is likely to change the way people work. Our data show that people use AI to do a variety of work activities related to the creation, analysis, and processing of information, i.e., information work.  Furthermore, people are using AI for an increasing variety of user goals over time.
Most occupations have at least some information work component, reflected in our finding that AI is at least somewhat relevant to most occupations. This helps to explain the rapid rate of adoption of AI in the workplace~\cite{bick2024rapid}. 

Our analysis also highlights the different ways AI may be used in different roles.
Occupations with high AI applicability to user goals, such as computer and mathematical occupations, may see more active AI usage by workers as they use AI to improve existing workflows. Meanwhile, occupations with high AI action applicability, like those in business and financial operations, may delegate tasks to AI and focus more on the parts of their jobs that require a human touch. 
To enable further research in this area, we have made our AI applicability scores and component metrics available at \url{https://github.com/microsoft/working-with-ai}. See \Cref{app:broader-impacts} for an extended discussion of the broader impacts of our work.

Our user goal versus AI action distinction relates to a key question in the literature and public discourse around AI: to what extent is AI automating versus augmenting work activities? 
It is tempting to conclude that occupations that have high AI action applicability score will be automated and thus experience job or wage loss, and that occupations with high user goal applicability score will be augmented and raise wages. 
This would be a mistake, as downstream consequences of new technologies are very hard to predict and can be counterintuitive~\cite{autor2015there}. 
Take the example of ATMs, which automated a core task of bank tellers, but led to an \emph{increase} in the number of bank teller jobs as banks opened more branches at lower costs and tellers focused on more valuable relationship-building rather than processing deposits and withdrawals~\cite{bessen2015toil}.
Our measures of AI applicability indicate where AI is likely to change work, which is a part of the automation vs.\ augmentation question, but AI's effects on employment and wages will depend on hard-to-predict business decisions as well as the emergence of new types of work. Early evidence on the labor market effects of AI is mixed, with some analyses pointing to little or no effect on earnings, hours, and unemployment rates~\cite{humlum2025still,massenkoffmccrory2026labor} and others finding that hiring of younger workers has declined in particular roles~\cite{brynjolfsson2026canaries}.

There are some natural limitations to the conclusions that can be drawn from our data. 
First and foremost, our data on AI usage are from 2024; models have improved dramatically since then. For follow-up analyses on M365 Copilot data from early 2026 using our methodology, see \citet{counts2026aienterprisepeopleuse}. The development of powerful agent harnesses is a particularly important advance, as is the increasing ease of use of AI tools within information workflows. 
Beyond the limitations imposed by the passage of time and technological advances, our methods themselves also have limitations. Our task completion and scope metrics are measures of AI's utility towards an IWA, but they do not capture the productivity impact on people doing that IWA.
Our data also represents only one slice of the AI market: there are many other AI platforms, including more task- or occupation-specific LLMs, which are not represented in our data. 
For instance, ChatGPT usage shows a higher prevalence of writing tasks~\citep{chatterji2025chatgpt}, while Claude usage skews towards coding and mathematics~\citep{handa2025economic}; our coverage thresholding approach provides a degree of robustness against this variance in usage patterns.
Next, while decomposing an occupation into its work activities is standard practice in the literature, it does not provide a complete representation of every occupation: the connecting glue between tasks also contributes to the value of work. Our use of O*NET means our results are shaped by its U.S.-centric view, may lag behind current workplace activities, and do not capture valuable tasks performed outside of occupations (e.g., work in the home or volunteering).
Modernizing our understanding of workplace activities will be crucial as generative AI continues to change how work is done.

We measured how AI capabilities overlap with work activities, but it remains to be seen how different occupations refactor their work responsibilities in response to AI's rapid progress. 
In addition, entirely new occupations may emerge, performing new types of work activities~\cite{acemoglu2019automation,capps2025AIjobs}. This is not a new phenomenon: the \emph{majority} of employment  today is in occupations that arose in the last 100 years as a result of new technologies~\cite{autor2024new}. Which new jobs emerge, and how old ones are reconstituted, is a crucial future research direction.

 \clearpage

\section*{Acknowledgments}
We thank Hancheng Cao, David Holtz, Kristina McElheran, Jennifer Neville, Ashish Sharma, Carolyn Tsao, the Microsoft Research AI Interaction and Learning Group, and the Microsoft Research Computational Social Science Working Group for helpful discussions and feedback, and Patrick Bourke, Nick Caurvina, Jonathan McLean, David Tittsworth, and Bryan Tower for software and data engineering support. This work has benefited from discussion with attendees of the 2025 Wharton Business \& Generative AI Conference, the 2025 NABE Tech Economics Conference, and the 2026 NBER Conference on AI and Economic Measurement. We thank the anonymous reviewers for their helpful suggestions.

\textit{Competing Interests.} At the time they performed this work, the authors were all employed by Microsoft. The views expressed in this paper are theirs alone. 

\bibliographystyle{unsrtnat}
\bibliography{references}

@misc{onet29,
  author       = {{National Center for O*NET Development}},
  title        = {{O*NET Database Version 29.0}},
  year         = {2024},
  url          = {https://www.onetcenter.org/db_releases.html},
  note         = {Accessed: 2025-05-29}
}

@misc{oews2023,
  author       = {{U.S. Bureau of Labor Statistics}},
  title        = {{Occupational Employment and Wage Statistics (OEWS), May 2023}},
  year         = {2024},
  url          = {https://www.bls.gov/oes/tables.htm},
  note         = {Accessed: 2025-10-17}
}

@misc{iscenko2026googlesaieconomy,
      title={{Google's} {AI} \& Economy {ATLAS} v1.0: Mapping {Gemini} Usage in the Economy}, 
      author={Zanna Iscenko and Scott Strand and Yiyuan Chen and Guillaume Aimard and Mihai Codreanu and Vivek Sampathkumar and Alex Imas and Julian Jacobs and Evalyne Muiruri and Juan Mateos-Garcia and Jia Jen Ng and Samirah Javed and Josh Martin and Omar Ajmeri and Denis Calin and Andrew Kim and Fabien Curto Millet and James Manyika},
      year={2026},
      eprint={2608.00038},
      archivePrefix={arXiv},
      primaryClass={econ.GN},
      url={https://arxiv.org/abs/2608.00038} 
}

@misc{bick2024rapid,
  title={The rapid adoption of generative {AI}},
  author={Bick, Alexander and Blandin, Adam and Deming, David J},
  year={2024},
  howpublished={NBER 32966},
url={https://doi.org/10.3386/w32966}
}

@misc{CPS_BLS,
  author       = {{U.S. Census Bureau} and {U.S. Bureau of Labor Statistics}},
  title        = {Current Population Survey {(CPS)}},
  year         = {2024},
  howpublished = {\url{https://www.bls.gov/cps/tables.htm}},
  note         = {Accessed 2025-08-18}
}

@misc{BLS_AI_exposure,
author = {{U.S. Bureau of Labor Statistics}},
title = {Artificial Intelligence ({AI}) exposure categories},
year={2026},
howpublished={\url{https://www.bls.gov/emp/publications/ai-exposure-categories.htm}},
note={Accessed 2026-10-08}
}

@article{eloundou2024gpts,
  title={{GPTs are GPTs}: Labor market impact potential of {LLMs}},
  author={Eloundou, Tyna and Manning, Sam and Mishkin, Pamela and Rock, Daniel},
  journal={Science},
  volume={384},
  number={6702},
  pages={1306--1308},
  year={2024},
  publisher={American Association for the Advancement of Science},
  url={https://www.science.org/doi/10.1126/science.adj0998}
}

@article{huvila2009ecological,
  title={Ecological framework of information interactions and information infrastructures},
  author={Huvila, Isto},
  journal={Journal of Information Science},
  volume={35},
  number={6},
  pages={695--708},
  year={2009},
  publisher={Sage publications Sage UK: London, England},
url={https://doi.org/10.1177/0165551509336705}
}

@article{huvila2016information,
  title={Information work in information science research and practice},
  author={Huvila, Isto and Lloyd, Annemaree and Budd, John M and Palmer, Carole and Toms, Elaine},
  journal={Proceedings of the Association for Information Science and Technology},
  volume={53},
  number={1},
  pages={1--5},
  year={2016},
  publisher={Wiley Online Library},
  url={https://doi.org/10.1002/pra2.2016.14505301004}
}

@inproceedings{hicke2026adoptneqadaptlongitudinal,
      title={Adopt $\neq$ Adapt: Longitudinal Analyses of {LLM} Conversations in the Wild}, 
      author={Rebecca M. M. Hicke and Kiran Tomlinson},
      year={2026},
      booktitle={Proceedings of the 2026 Conference on Empirical Methods in Natural Language Processing},
      url={https://arxiv.org/abs/2605.29018}, 
      note={In press}
}

@article{pyoria2005concept,
  title={The concept of knowledge work revisited},
  author={Py{\"o}ri{\"a}, Pasi},
  journal={Journal of Knowledge Management},
  volume={9},
  number={3},
  pages={116--127},
  year={2005},
  publisher={Emerald Group Publishing Limited},
  url={https://doi.org/10.1108/13673270510602818}
}

@article{goldfarb2023could,
  title={Could machine learning be a general purpose technology? A comparison of emerging technologies using data from online job postings},
  author={Goldfarb, Avi and Taska, Bledi and Teodoridis, Florenta},
  journal={Research Policy},
  volume={52},
  number={1},
  pages={104653},
  year={2023},
  publisher={Elsevier},
  url={https://doi.org/10.1016/j.respol.2022.104653}
}

@misc{handa2025economic,
      title={Which Economic Tasks are Performed with {AI}? Evidence from Millions of {Claude} Conversations}, 
      author={Kunal Handa and Alex Tamkin and Miles McCain and Saffron Huang and Esin Durmus and Sarah Heck and Jared Mueller and Jerry Hong and Stuart Ritchie and Tim Belonax and Kevin K. Troy and Dario Amodei and Jared Kaplan and Jack Clark and Deep Ganguli},
      year={2025},
      eprint={2503.04761},
      archivePrefix={arXiv},
      primaryClass={cs.CY},
      url={https://arxiv.org/abs/2503.04761}, 
}

@misc{counts2026aienterprisepeopleuse,
      title={{AI} in the Enterprise: How People Use {M365 Copilot Chat}}, 
      author={Scott Counts and Yan Chen and Jing Dong and Himanshu Sharma and Andrey Zaikin and Rui Hu and Alperen Kok and Gorkem Ozer Yilmaz and Siddharth Suri and Kiran Tomlinson and Sonia Jaffe and Will Wang},
      year={2026},
      eprint={2605.23958},
      archivePrefix={arXiv},
      primaryClass={cs.CY},
      url={https://arxiv.org/abs/2605.23958}, 
}

@misc{suri2024use,
      title={The Use of Generative Search Engines for Knowledge Work and Complex Tasks}, 
      author={Siddharth Suri and Scott Counts and Leijie Wang and Chacha Chen and Mengting Wan and Tara Safavi and Jennifer Neville and Chirag Shah and Ryen W. White and Reid Andersen and Georg Buscher and Sathish Manivannan and Nagu Rangan and Longqi Yang},
      year={2024},
      eprint={2404.04268},
      archivePrefix={arXiv},
      primaryClass={cs.IR},
      url={https://arxiv.org/abs/2404.04268}, 
}

@article{autor2003skill,
  title={The skill content of recent technological change: An empirical exploration},
  author={Autor, David H and Levy, Frank and Murnane, Richard J},
  journal={The Quarterly Journal of Economics},
  volume={118},
  number={4},
  pages={1279--1333},
  year={2003},
  publisher={MIT Press},
  url={https://doi.org/10.1162/003355303322552801}
}

@inproceedings{achananuparp2026multi,
  title={A multi-stage framework with taxonomy-guided reasoning for occupation classification using large language models},
  author={Achananuparp, Palakorn and Lim, Ee-Peng},
  booktitle={Proceedings of the International AAAI Conference on Web and Social Media},
  volume={20},
  year={2026},
  url={https://ojs.aaai.org/index.php/ICWSM/article/view/42622/50182}
}

@article{septiandri2024potential,
  title={The potential impact of {AI} innovations on {US} occupations},
  author={Septiandri, Ali Akbar and Constantinides, Marios and Quercia, Daniele},
  journal={PNAS Nexus},
  volume={3},
  number={9},
  pages={pgae320},
  year={2024},
  publisher={Oxford University Press US},
  url={https://doi.org/10.1093/pnasnexus/pgae320}
}

@inproceedings{wildchat,
 author = {Zhao, Wenting and Ren, Xiang and Hessel, Jack and Cardie, Claire and Choi, Yejin and Deng, Yuntian},
 booktitle = {International Conference on Learning Representations},
 pages = {34590--34605},
 title = {{WildChat: 1M ChatGPT} Interaction Logs in the Wild},
 url = {https://proceedings.iclr.cc/paper_files/paper/2024/file/9421261e06f1a63a352b068f1ac90609-Paper-Conference.pdf},
 year = {2024}
}

@techreport{brynjolfsson2026canaries,
  author      = {Brynjolfsson, Erik and Chandar, Bharat and Chen, Ruyu},
  title       = {Canaries in the Coal Mine? {Six} Facts about the Recent Employment Effects of Artificial Intelligence},
  year        = {2026},
  month       = {August},
  institution = {Stanford Digital Economy Lab},
  url         = {https://digitaleconomy.stanford.edu/publications/canaries-in-the-coal-mine/},
  note        = {Revised version; original published 2025}
}

@misc{humlum2025still,
  title={Still waters, rapid currents: Early labor market transformation under generative {AI}},
  author={Humlum, Anders and Vestergaard, Emilie},
  year={2026},
  institution={National Bureau of Economic Research},
  howpublished={NBER Working Paper No.\ w33777},
  url={https://www.nber.org/papers/w33777}
}

@misc{cui2024effects,
    author = {Cui, Zheyuan Kevin and Demirer, Mert and Jaffe, Sonia and Musolff, Leon and Peng, Sida and Salz, Tobias},
    title = {The effects of generative {AI} on high skilled work: Evidence from three field experiments with software developers},
    howpublished = {SSRN 4945566},
    year = {2024},
url={http://doi.org/10.2139/ssrn.4945566}
}

@misc{shao2025futureworkaiagents,
      title={Future of Work with {AI} Agents: Auditing Automation and Augmentation Potential across the {U.S.} Workforce}, 
      author={Yijia Shao and Humishka Zope and Yucheng Jiang and Jiaxin Pei and David Nguyen and Erik Brynjolfsson and Diyi Yang},
      year={2026},
      eprint={2506.06576},
      archivePrefix={arXiv},
      primaryClass={cs.CY},
      url={https://arxiv.org/abs/2506.06576}, 
}

@techreport{gmyrek2025generative,
  title={Generative {AI} and jobs: A refined global index of occupational exposure},
  author={Gmyrek, Pawe{\l} and Berg, Janine and Kami{\'n}ski, Karol and Konopczy{\'n}ski, Filip and {\L}adna, Agnieszka and Nafradi, Balint and Ros{\l}aniec, Konrad and Troszy{\'n}ski, Marek},
  year={2025},
  institution={International Labour Organization},
  howpublished={ILO Working Paper No. 140},
  url={https://www.ilo.org/publications/generative-ai-and-jobs-refined-global-index-occupational-exposure}
}

@article{bresnahan1995general,
  title={General purpose technologies: ``Engines of growth''?},
  author={Bresnahan, Timothy F and Trajtenberg, Manuel},
  journal={Journal of Econometrics},
  volume={65},
  number={1},
  pages={83--108},
  year={1995},
  publisher={Elsevier},
  url={https://doi.org/10.1016/0304-4076(94)01598-T}
}

@article{mcduff2025towards,
  title={Towards accurate differential diagnosis with large language models},
  author={McDuff, Daniel and Schaekermann, Mike and Tu, Tao and Palepu, Anil and Wang, Amy and Garrison, Jake and Singhal, Karan
and Sharma, Yash and Azizi, Shekoofeh and Kulkarni, Kavita and Hou, Le and Cheng, Yong and Liu, Yun and Mahdavi, S. Sara and Prakash, Sushant and Pathak, Anupam and Semturs, Christopher and Patel, Shwetak and Webster, Dale R. and Dominowska, Ewa and Gottweis, Juraj and Barral, Joelle and Chou, Katherine and Corrado, Greg S. and Matias, Yossi and Sunshine, Jake and Karthikesalingam, Alan and Natarajan, Vivek},
  journal={Nature},
  pages={451--457},
    volume={642},
  year={2025},
  publisher={Nature Publishing Group UK London},
  url={https://www.nature.com/articles/s41586-025-08869-4}
}

@article{frey2017employment,
title = {The future of employment: How susceptible are jobs to computerisation?},
journal = {Technological Forecasting and Social Change},
volume = {114},
pages = {254-280},
year = {2017},
issn = {0040-1625},
author = {Carl Benedikt Frey and Michael A. Osborne},
url={https://doi.org/10.1016/j.techfore.2016.08.019}
}

@misc{capps2025AIjobs,
  author       = {Robert Capps},
  title        = {{A.I. might take your job. Here are 22 new ones it could give you}},
  howpublished      = {\textit{The New York Times Magazine} [Internet]},
  year         = {2025},
month = {6},
day={17},
date         = {2025-06-17},
  url          = {https://www.nytimes.com/2025/06/17/magazine/ai-new-jobs.html}
}

@article{bessen2015toil,
  title={Toil and technology: Innovative technology is displacing workers to new jobs rather than replacing them entirely},
  author={Bessen, James},
  journal={Finance \& Development},
  volume={52},
  number={1},
  pages={16-19},
  year={2015},
  publisher={International Monetary Fund},
  url={https://doi.org/10.5089/9781498351942.022.A007}
}

@manual{BLS_SOC_2018,
  title        = {Standard Occupational Classification Manual, 2018},
  author       = {{U.S.\ Office of Management and Budget}},
  year         = {2017},
  url          = {https://www.bls.gov/soc/2018/soc_2018_manual.pdf},
note = {Accessed: 2025-07-07}
}

@article{autor2015there,
  title={Why are there still so many jobs? The history and future of workplace automation},
  author={Autor, David H},
  journal={Journal of Economic Perspectives},
  volume={29},
  number={3},
  pages={3--30},
  year={2015},
  publisher={American Economic Association 2014 Broadway, Suite 305, Nashville, TN 37203-2418},
  url={https://www.aeaweb.org/articles?id=10.1257/jep.29.3.3}
}

@article{autor2024new,
  title={New frontiers: The origins and content of new work, 1940--2018},
  author={Autor, David and Chin, Caroline and Salomons, Anna and Seegmiller, Bryan},
  journal={The Quarterly Journal of Economics},
  volume={139},
  number={3},
  pages={1399--1465},
  year={2024},
  publisher={Oxford University Press},
  url={https://doi.org/10.1093/qje/qjae008}
}

@article{james1984estimating,
  title={Estimating within-group interrater reliability with and without response bias.},
  author={James, Lawrence R and Demaree, Robert G and Wolf, Gerrit},
  journal={Journal of Applied Psychology},
  volume={69},
  number={1},
  pages={85-98},
  year={1984},
  publisher={American Psychological Association},
  url={https://doi.org/10.1037/0021-9010.69.1.85}
}

@article{acemoglu2019automation,
Author = {Acemoglu, Daron and Restrepo, Pascual},
Title = {Automation and new tasks: How technology displaces and reinstates labor},
Journal = {Journal of Economic Perspectives},
Volume = {33},
Number = {2},
Year = {2019},
Month = {May},
Pages = {3–30},
url={https://www.aeaweb.org/articles?id=10.1257/jep.33.2.3}
}

@article{felten2018ai,
Author = {Felten, Edward W. and Raj, Manav and Seamans, Robert},
Title = {A method to link advances in artificial intelligence to occupational abilities},
Journal = {AEA Papers and Proceedings},
Volume = {108},
Year = {2018},
Month = {May},
Pages = {54–57},
url={https://www.aeaweb.org/articles?id=10.1257/pandp.20181021}
}

@misc{felten2023will,
      title={How will Language Modelers like {ChatGPT} Affect Occupations and Industries?}, 
      author={Ed Felten and Manav Raj and Robert Seamans},
      year={2023},
      eprint={2303.01157},
      archivePrefix={arXiv},
      primaryClass={econ.GN},
      url={https://arxiv.org/abs/2303.01157}, 
}

@article{felten2021occupational,
  title={Occupational, industry, and geographic exposure to artificial intelligence: A novel dataset and its potential uses},
  author={Felten, Edward and Raj, Manav and Seamans, Robert},
  journal={Strategic Management Journal},
  volume={42},
  number={12},
  pages={2195--2217},
  year={2021},
  publisher={Wiley Online Library},
  url={https://doi.org/10.1002/smj.3286}
}

@article{brynjolfsson2018machines,
Author = {Brynjolfsson, Erik and Mitchell, Tom and Rock, Daniel},
Title = {What can machines learn, and what does it mean for occupations and the economy?},
Journal = {AEA Papers and Proceedings},
Volume = {108},
Year = {2018},
Month = {May},
Pages = {43–47},
url={https://www.aeaweb.org/articles?id=10.1257/pandp.20181019}
}

@techreport{manyika2017future,
    author = {James Manyika and Michael Chui and Mehdi Miremadi and Jacques Bughin and Katy George and Paul Willmott and Martin Dewhurst},
    title = {A future that works: Automation, employment, and productivity},
    institution = {McKinsey Global Institute},
    year = {2017},
    url={https://www.mckinsey.com/featured-insights/digital-disruption/harnessing-automation-for-a-future-that-works}
}

@book{drucker1959landmarks,
  address = {New York},
  author = {Drucker, Peter Ferdinand},
  edition = {1st},
  language = {eng},
  pages = 270,
  publisher = {Harper},
  title = {Landmarks of tomorrow: a report on the new ``post-modern'' world},
  year = 1959
}

@book{mckercher2008knowledge,
  title={Knowledge Workers in the Information Society},
  author={McKercher, C. and Mosco, V.},
  isbn={9780739117811},
  lccn={2007035318},
  year={2008},
  publisher={Lexington Books}
}

@article{reinhardt2011knowledge,
author = {Reinhardt, Wolfgang and Schmidt, Benedikt and Sloep, Peter and Drachsler, Hendrik},
title = {Knowledge worker roles and actions---Results of two empirical studies},
journal = {Knowledge and Process Management},
volume = {18},
number = {3},
pages = {150-174},
eprint = {https://onlinelibrary.wiley.com/doi/pdf/10.1002/kpm.378},
year = {2011},
url={https://doi.org/10.1002/kpm.378}
}

@InProceedings{drouin2024workarena,
  title = 	 {{W}ork{A}rena: How Capable are Web Agents at Solving Common Knowledge Work Tasks?},
  author =       {Drouin, Alexandre and Gasse, Maxime and Caccia, Massimo and Laradji, Issam H. and Del Verme, Manuel and Marty, Tom and Vazquez, David and Chapados, Nicolas and Lacoste, Alexandre},
  booktitle = 	 {Proceedings of the 41st International Conference on Machine Learning},
  pages = 	 {11642--11662},
  year = 	 {2024},
  editor = 	 {Salakhutdinov, Ruslan and Kolter, Zico and Heller, Katherine and Weller, Adrian and Oliver, Nuria and Scarlett, Jonathan and Berkenkamp, Felix},
  volume = 	 {235},
  series = 	 {Proceedings of Machine Learning Research},
  month = 	 {21--27 Jul},
  publisher =    {PMLR},
  url = 	 {https://proceedings.mlr.press/v235/drouin24a.html},
}

@misc{doosterlinck2024extreme,
      title={In-Context Learning for Extreme Multi-Label Classification}, 
      author={Karel D'Oosterlinck and Omar Khattab and François Remy and Thomas Demeester and Chris Develder and Christopher Potts},
      year={2024},
      eprint={2401.12178},
      archivePrefix={arXiv},
      primaryClass={cs.CL},
      url={https://arxiv.org/abs/2401.12178}, 
}

@misc{anthropic2026aeiv5,
        author = {Maxim Massenkoff and Eva Lyubich and Peter McCrory and Ruth Appel and Ryan Heller},
        title = {Anthropic Economic Index report: Learning curves},
        date = {2026-03-24},
        year = {2026},
        url = {https://www.anthropic.com/research/economic-index-march-2026-report},
}

@misc{appelmccrorytamkin2025geoapi,

author = {Ruth Appel and Peter McCrory and Alex Tamkin and Michael Stern and Miles McCain and Tyler Neylon},

title = {Anthropic Economic Index report: Uneven geographic and enterprise {AI} adoption},

date = {2025-09-15},

year = {2025},

url = {https://www.anthropic.com/research/anthropic-economic-index-september-2025-report},

}

@misc{anthropic2026economicprimitives,
        author = {Ruth Appel and Maxim Massenkoff and Peter McCrory and Miles McCain and Ryan Heller and Tyler Neylon and Alex Tamkin},
        title = {Anthropic Economic Index report: economic primitives},
        date = {2026-01-15},
        year = {2026},
        url = {https://www.anthropic.com/research/anthropic-economic-index-january-2026-report},
}

@misc{massenkoffmccrory2026labor,
 author = {Maxim Massenkoff and Peter McCrory},
 title = {Labor market impacts of {AI}: A new measure and early evidence},
 date = {2026-03-05},
 year = {2026},
 url = {https://www.anthropic.com/research/labor-market-impacts},
}

@inproceedings{vishwakarma-etal-2025-llms,
    title = "Can {LLM}s Help You at Work? A Sandbox for Evaluating {LLM} Agents in Enterprise Environments",
    author = "Vishwakarma, Harsh  and
      Agarwal, Ankush  and
      Patil, Ojas  and
      Devaguptapu, Chaitanya  and
      Chandran, Mahesh",
    booktitle = "Proceedings of the 2025 Conference on Empirical Methods in Natural Language Processing",
    year = "2025",
    publisher = "Association for Computational Linguistics",
    url = "https://aclanthology.org/2025.emnlp-main.466/",
    pages = "9167--9201",
}

@misc{wang2026doesagentdevelopmentreflect,
      title={How Well Does Agent Development Reflect Real-World Work?}, 
      author={Zora Zhiruo Wang and Sanidhya Vijayvargiya and Aspen Chen and Hanmo Zhang and Venu Arvind Arangarajan and Jett Chen and Valerie Chen and Diyi Yang and Daniel Fried and Graham Neubig},
      year={2026},
      eprint={2603.01203},
      archivePrefix={arXiv},
      primaryClass={cs.AI},
      url={https://arxiv.org/abs/2603.01203}, 
}

@inproceedings{
xu2025theagentcompany,
title={{TheAgentCompany}: Benchmarking {LLM} Agents on Consequential Real World Tasks},
author={Frank F. Xu and Yufan Song and Boxuan Li and Yuxuan Tang and Kritanjali Jain and Mengxue Bao and Zora Zhiruo Wang and Xuhui Zhou and Zhitong Guo and Murong Cao and Mingyang Yang and Hao Yang Lu and Amaad Martin and Zhe Su and Leander Melroy Maben and Raj Mehta and Wayne Chi and Lawrence Keunho Jang and Yiqing Xie and Shuyan Zhou and Graham Neubig},
booktitle={The Thirty-ninth Annual Conference on Neural Information Processing Systems Datasets and Benchmarks Track},
year={2025},
url={https://openreview.net/forum?id=LZnKNApvhG}
}

@inproceedings{
miserendino2025swelancer,
title={{SWE-Lancer}: Can Frontier {LLM}s Earn \$1 Million from Real-World Freelance Software Engineering?},
author={Samuel Miserendino and Michele Wang and Tejal Patwardhan and Johannes Heidecke},
booktitle={Forty-second International Conference on Machine Learning},
year={2025},
url={https://openreview.net/forum?id=xZXhFg43EI}
}

@article{tolan2017measure,
 author = {Song{\"u}l Tolan and  Annarosa Pesole and  Fernando Mart{\'i}nez-Plumed and  Enrique Fern{\'a}ndez-Mac{\'i}as and Jos{\'e} Hernández-Orallo and Emilia G{\'o}mez},
 journal = {Journal of Artificial Intelligence Research},
 volume = {71}, 
 year = {2021},
 title = {Measuring the occupational impact of {AI}: tasks, cognitive abilities and {AI} benchmarks},
 pages = {191-236},
 url={https://doi.org/10.1613/jair.1.12647}
}

@misc{web2020impact,
 author = {Michael Webb},
 title = {The impact of artificial intelligence on the labor market},
 year = {2020},
 howpublished = {SSRN 3482150},
url={https://doi.org/10.2139/ssrn.3482150}
}

@misc{chatterji2026organizationsuseaievidence,
      title={How Organizations Use {AI}: Evidence from {ChatGPT}}, 
      author={Aaron Chatterji and David Holtz and Neel Rakholia and Prasanna Tambe and Gawesha Weeratunga},
      year={2026},
      eprint={2608.12236},
      archivePrefix={arXiv},
      primaryClass={econ.GN},
      url={https://arxiv.org/abs/2608.12236}, 
}

@misc{chatterji2025chatgpt,
 title = {How People Use {ChatGPT}},
  author={Chatterji, Aaron and Cunningham, Thomas and Deming, David J and Hitzig, Zoe and Ong, Christopher and Shan, Carl Yan and Wadman, Kevin},
 year = {2025}, 
 howpublished = {NBER Working Paper 34255},
 url={https://www.nber.org/papers/w34255}
}

@inproceedings{
patwardhan2026gdpval,
title={{GDP}val: Evaluating {AI} Model Performance on Real-World Economically Valuable Tasks},
author={Tejal Patwardhan and Rachel Dias and Elizabeth Proehl and Grace Kim and Michele Wang and Olivia Watkins and Simon Posada Fishman and Marwan Aljubeh and Phoebe Thacker and Laurance Fauconnet and Natalie S. Kim and Samuel Miserendino and Gildas Chabot and David Li and Patrick Chao and Michael Sharman and Alexandra Barr and Amelia Glaese and Jerry Tworek},
booktitle={The Fourteenth International Conference on Learning Representations},
year={2026},
url={https://openreview.net/forum?id=hcuEdq6eKD}
}

@inproceedings{
jimenez2024swebench,
title={{SWE}-bench: Can Language Models Resolve Real-world {GitHub} Issues?},
author={Carlos E Jimenez and John Yang and Alexander Wettig and Shunyu Yao and Kexin Pei and Ofir Press and Karthik R Narasimhan},
booktitle={The Twelfth International Conference on Learning Representations},
year={2024},
url={https://openreview.net/forum?id=VTF8yNQM66}
}

\clearpage

\appendix
\renewcommand{\thefigure}{\thesection.\arabic{figure}}
\renewcommand{\thetable}{\thesection.\arabic{table}}

\section{Merging O*NET with BLS and CPS data}\label{app:onet-bls}

The BLS Occupational Employment and Wage Statistics data identifies occupations by Standard Occupational Classification (SOC) codes, which differ slightly from the O*NET-SOC codes used in O*NET data. We use the BLS-provided mapping between the codes (\url{https://www.bls.gov/emp/documentation/crosswalks.htm}) and present all of our results in terms of SOC occupations. When multiple O*NET-SOC occupations share the same SOC code (e.g., Tour Guides and Travel Guides share the SOC code for ``Tour and travel guides''), we take the union over O*NET data mapping to the SOC Code (i.e., tasks and DWA/IWA/GWAs) \footnote{When comparing our AI applicability scores to E1 exposure scores from Eloundou et al.~\cite{eloundou2024gpts}, we take the mean of O*NET-SOC occupational exposure scores mapped to the same SOC code.}.

To measure how AI applicability varies across demographic groups, we use the Current Population Survey (CPS) data from 2024~\cite{CPS_BLS}, which is gathered by the U.S.\ Census Bureau for the BLS. In particular, we use tables 11 (``Employed persons by detailed occupation, sex, race, and Hispanic or Latino ethnicity'') and 11b (``Employed persons by detailed occupation and age'') from the 2024 CPS annual averages. These provide demographic statistics for detailed occupation codes, although CPS uses a different occupational taxonomy than either SOC or O*NET. We use the BLS-provided crosswalk from SOC codes to CPS codes (also at the above link) to propagate SOC-level AI applicability scores to CPS-level AI applicability scores. When multiple CPS codes map to the same SOC code, we apply the SOC-level score to each CPS code. When multiple SOC codes map to the same CPS code (which is much more common), we use the employment-weighted average of the SOC-level AI applicability scores.

\subsection{Handling missing data}\label{app:mising-data}
We omit all military occupations (SOC Codes 55-xxxx), as they have no task data in O*NET, no employment data in the BLS OEWS data, and are not included in the O*NET-SOC to SOC crosswalk. We also omit fishing and hunting workers (SOC Code 45-3031), as they are missing from the 2023 OEWS data. Finally, we omit 74 SOC codes mapping to O*NET occupations for which there is no task data. This leaves us with 785 SOC codes covering 149.8 million workers in the 2023 OEWS data (total US employment in 2023 OEWS data is 151.9 million).

\section{Estimating real-world IWA frequency}\label{app:real-world-freq}
For each task, O*NET provides the overall share of survey respondents in an occupation who say the task is relevant to their job, as well as the share of respondents who report performing that task at each of various frequency levels (e.g., hourly, weekly, yearly). To convert these into annual total workforce counts for each IWA, we perform the following procedure:
\begin{enumerate}
    \item Convert O*NET task frequency categories into annual counts based on 260 workdays/year and 8 hours/workday: ``Yearly or less'': 1, ``More than yearly'': 4, ``More than monthly'': 24, ``more than weekly'': 104, ``Daily'': 260, ``Several times daily'': 780, ``Hourly or more'': 2080.
    \item For each task, compute its average annual frequency by averaging the above counts (weighted by surveyed percentages) and multiplying by relevance. 
    \item To get IWA-level frequencies, sum over tasks mapping to the same IWA. 
    \item To compute the total annual counts of an IWA in the workforce, sum over all occupations performing the IWA, multiplying by employment of each occupation. 
\end{enumerate}

\section{Work activities and scope classifier}\label{app:iwa-and-scope}
For work activity and scope classification, we used \texttt{gpt-4o-2024-08-06}, API version \texttt{2024-08-01-preview}. We used temperature 1 for the generate prompt and temperature 0 for the classify prompt.  We had initially intended to only classify the top-$k$ most relevant IWAs in the sorted order generated by the stage one generate prompt, but decided to classify every IWA for completeness. We kept the stage one sorting since we found that grouping IWAs by similarity to the generated summaries led to better agreement with human labels.

\subsection{Validation}
We tuned and validated our prompts using independent annotations by three of the authors on a sample of 195 anonymized English conversations which were already automatically scrubbed of personally identifiable information (the sensitive nature of the data precluded external annotators). For each conversation, the three annotators were shown the conversation text, 20 candidate user goal IWAs, and 20 candidate AI action IWAs. These sets of 20 consisted of the 10 most similar according to cosine similarity to stage one summaries (where matches are dramatically more likely) and 10 uniformly sampled from the next 90 most similar IWAs, all shuffled together; the same IWAs were sampled across annotators. The annotators independently listed all matching IWAs for the user goal and all matching IWAs for the AI action. We randomly split the conversations into a validation set of 95 used for prompt and pipeline tuning and a test set of 100, which was not touched until all full-scale pipeline runs had completed. The binary classification task over IWA matches was challenging but still had moderate agreement, with Cohen's kappa inter-rater reliabilities of 0.51, 0.58, 0.41 between the three pairs of annotators for user goal classification and 0.50, 0.49, 0.57 for AI action (on the test set, $n=100$). 
The inter-rater reliabilities of our human annotators were higher than the $\kappa = 0.27$ for the pair of human annotators in \cite{chatterji2025chatgpt} on a similar task.
Our final classification pipeline achieves Cohen's kappas with our three annotators on the test set that are only slightly lower: 0.44, 0.35, 0.38 for user goal and 0.53, 0.34, 0.39 for AI action (with very similar scores on validation, $0.48, 0.38, 0.42$ for user goals and $0.52, 0.38, 0.32$ for AI actions, indicating that our prompt tuning did not result in overfitting). These kappa scores are generally low due to the high degree of uncertainty around whether a particular IWA accurately describes the intent of the user or the action of the AI; in many cases, it is easy to make compelling arguments both that an IWA does and does not apply to a conversation, so we found even moderate agreement encouraging. Additionally, the overall match rate is very low (single-digit percentages), so the overall accuracies of all raters (including our LLM pipeline) with respect to each other are well over 90\%.    Our final prompts are included below.

In addition to validating the IWA classifications, we also validated the scope of impact ordinal classification. Note that the sample size for scope of impact validation is limited by only being able to compare scope classifications on IWAs that both raters labeled as a match. The test set user goal and AI action $r_{wg}$ scores~\cite{james1984estimating} (where 1 indicates perfect agreement and 0 only agreement due to chance) between the human rater pairs were 0.48 ($n=55$), 0.32 ($n = 66$), 0.55 ($n=51$) and 0.66 ($n = 74$), 0.12 ($n = 89$), 0.30 ($n=88$), respectively. This indicates some agreement, but with substantial variance. Between human raters and LLM classifications, the corresponding $r_{wg}$ scores were 0 ($n=43$), 0.35 ($n=44$), 0.49 ($n=34$) and 0.19 ($n = 90$), 0.04 ($n = 64$), 0.28 ($n=61$). As another measure, the average mean absolute error (MAE) between pairs of human raters on the test set was 0.94 for user goals and 1.05 for AI actions, while the average human-LLM MAEs were 1.06 and 1.23. In comparison, the expected MAE against uniform random ratings on a six-point scale is $35/18 \approx 1.94$. This indicates that the scope of impact LLM classifications have some agreement with human ratings, although the gap between human-human and human-LLM agreement is larger for scope classification than for IWA classification. An independent indication that, while noisy, the scope of impact classifier captures real signal is its correlation with IWA share ($r = 0.64$ and $r = 0.54$ for user goals and AI action; \Cref{fig:scope-vs-activity-share}).

\subsection{Prompts}
\begin{figure}[tbhp]
  \centering
  \includegraphics[width=\linewidth]{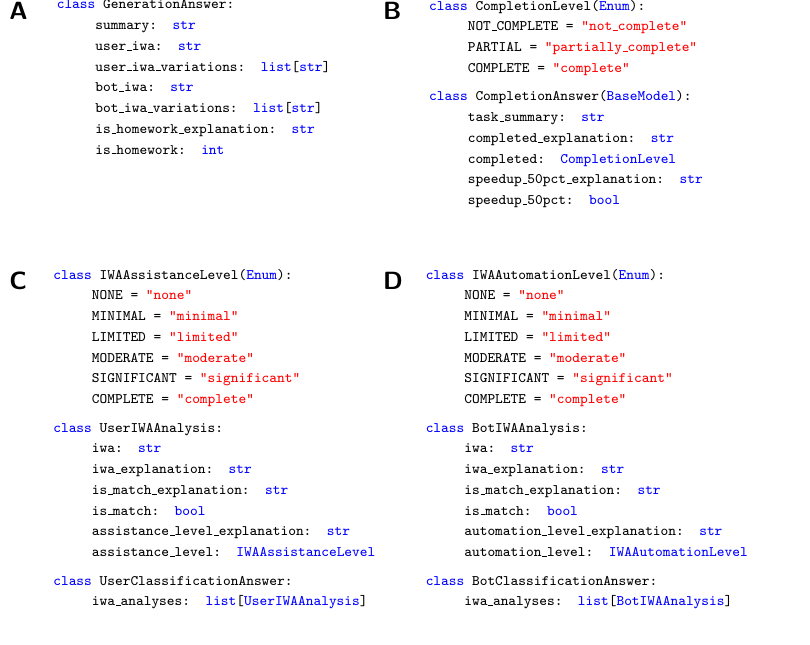}

  \caption{\textbf{Structured output formats for our four conversation-level LLM prompts.} (\textbf{A}) Generation prompt. (\textbf{B}) Completion. (\textbf{C}) Classify, user goals. (\textbf{D}) Classify, AI actions.}\label{fig:structured-outputs}
\end{figure}

\Cref{fig:structured-outputs} shows the structure of each prompt's output.

\begin{prompt}{Generate prompt (note: the \texttt{is\_homework} field is an unused remnant from early testing)}{}
<|Instruction|>
# Task overview
You will be given a conversation between a User and an AI chatbot.
You have two primary goals:
(1) summarize the main goal that the user is trying to accomplish in the style of an O*NET Intermediate Work Activity (IWA).
(2) summarize the action that the bot is performing in the conversation in the style of an O*NET IWA.
For example, if the user asks for help with a computer issue and the bot provides suggestions to resolve the issue, the user's IWA is "Resolve computer problems" and the bot's IWA is "Advise others on the design or use of technologies."
Sometimes, the user intent and bot action may be the same. 
For instance, if the user asks the bot to spellcheck a research paper and the bot corrects a few misspelled words, the user's IWA is "Edit written materials or documents" and the bot's IWA is also "Edit written materials or documents"
For both the user and bot IWA summaries, you will generate several variations of the summary to capture the same intent using different wordings.
To aid your analysis, you will also summarize the conversation.
Finally, you will also determine whether the User is a student trying to do homework. 

# Task details
Your task is to fill out the following fields: 
summary: Summarize User's queries in 3 sentences or fewer in **English**.
user_iwa: Summarize the task the user is trying to accomplish in the style of an O*NET IWA. Ensure that the summary accurately describes the goal of the User as directly evidenced in the conversation. Ensure that the summary matches the level of generality of an O*NET IWA: it should general enough to be an activity performed in a large number of occupations across multiple job families, but specific enough to capture the essence of the User's goal. Provide exactly one succinct IWA-style summary.
user_iwa_variations: Generate 4 variations of the user IWA summary that capture the same intent using different wordings.
bot_iwa: Summarize the task that the bot is performing in the style of an O*NET IWA. Ensure that the summary matches the level of generality of an O*NET IWA: it should general enough to be an activity performed in a large number of occupations across multiple job families, but specific enough to capture the essence of the bot's actions. Provide exactly one succinct IWA-style summary.
bot_iwa_variations: Generate 4 variations of the bot IWA summary that capture the same action using different wordings.
is_homework_explanation: Determine whether the User is a student trying to do homework. This may be obvious if they have pasted in assignment instructions, or it may be clear from the type of question they are asking. Explain in one sentence.
is_homework: Based on your explanation, provide the label 0 (not homework) or 1 (homework).

# Hints
Provide your answers in **English** using the given structured output format. 
<|end Instruction|>

<|Conversation between User and AI|>
{convo}
<|end Conversation|>

<|end of prompt|>
 \end{prompt}

\begin{prompt}{Classify user prompt}{}
<|Instruction|>
# Task overview
You will be given a conversation between a User and an AI chatbot as well as a summary of the conversation and a list of Candidate Intermediate Work Activity (IWA) statements from O*NET.
The IWAs will be numbered with numerical IDs to help you reference them in your responses.
Your primary task is to determine for each of the Candidate IWAs whether the user is trying to perform that IWA, according to the meaning of the IWA in the context of O*NET. The conversation must provide direct evidence that the user is themself trying to accomplish the IWA.
For example, a user asking for tech support does not match a IWA about providing tech support, but does match a IWA about resolving technical issues.
As another example, a user seeking information about a product does not match a IWA about providing product information, but does match a IWA about researching product information.
Additionally, you will determine the level of assistance that the bot provides to the user in the conversation for each matching IWA.

# Task details
Your reply to iwa_analyses should be a list of UserIWAAnalysis objects, one for each Candidate IWA in the order below. For each Candidate IWA, you will analyze the user's intent relative to that IWA and fill out the fields of UserIWAAnalysis as follows:
iwa (str): Copy the current Candidate IWA verbatim into this field. All of the following fields will be based on this IWA.
iwa_explanation (str): Explain in one sentence what the IWA means in the context of O*NET and what kinds of occupations perform this IWA.
is_match_explanation (str): Explain in one sentence whether the user is seeking to perform an activity described by the IWA, according to the meaning of the IWA in O*NET. To be considered a match, the user's intent must be to perform the action themselves, so if the IWA mentions or implies assisting clients or customers, for instance, there must be evidence in their query that the user is seeking to assist a client or customer.
is_match (bool): Based on your explanation, provide the label True if the user is seeking to perform an activity described by the IWA, according to the meaning of the IWA in O*NET, and False otherwise. To be considered a match, the user's intent must be to perform the action themselves.
assistance_level_explanation (str): Consider the full scope of the work performed under this IWA across all occupations. What fraction of this work can the bot assist users with by applying only the capability it demonstrates in this conversation? Pay careful attention to the fact that the IWA might encompass many more subtasks than represented in this conversation. Explain in one sentence, or reply N/A if the IWA does not match the user's intent (i.e., when is_match is False).
assistance_level (IWAAssistanceLevel): Based on your explanation, label the bot's capability to assist with the IWA using the IWAAssistanceLevel enum, which has the following options:
- none: The user is not seeking to perform the IWA, or the conversation does not indicate that the bot is capable of assisting with the IWA.
- minimal: With this demonstrated capability, the bot can assist with a minimal portion of the work in the IWA. 
- limited: With this demonstrated capability, the bot can assist with a limited portion of the work in the IWA. 
- moderate: With this demonstrated capability, the bot can assist with a moderate portion of the work in the IWA. 
- significant: With this demonstrated capability, the bot can assist with a significant portion of the work in the IWA. 
- complete: With this demonstrated capability, the bot can assist with all of the work in the IWA.

# Hints
- Provide your answers in **English** using the given structured output format.
<|end Instruction|>

<|Conversation between User and AI|>
{convo}
<|end Conversation|>

<|Conversation Summary|>
{summary}
<|end Conversation Summary|>

<|Candidate IWAs|>
{iwas}
<|end Candidate IWAs|>

<|end of prompt|>   
\end{prompt}

\begin{prompt}{Classify bot prompt}{}
<|Instruction|>
# Task overview
You will be given a conversation between a User and an AI chatbot as well as a summary of the conversation and a list of Candidate Intermediate Work Activity (IWA) statements from O*NET.
The IWAs will be numbered with numerical IDs to help you reference them in your responses.
Your task is to determine for each of the Candidate IWAs whether the bot is performing that IWA in the conversation, based on the meaning of the IWA in the context of O*NET. 
For example, if the user asks for help with a computer issue and the bot provides suggestions to resolve the issue, this matches an IWA about providing tech support, as that is the task that the bot is performing.
However, if the user asks the bot to spellcheck a research paper and the bot corrects a few misspelled words, this does not match an IWA about writing research papers: while the **user's** overarching goal may be writing research papers, that does not match the **bot's** task in the conversation.
Additionally, you will assess whether this conversation demonstrates the bot's ability to automate each matching IWA in the conversation.

# Task details
Your reply to iwa_analyses should be a list of BotIWAAnalysis objects, one for each candidate IWA in the order below. For each candidate IWA, you will analyze the bot's actions relative to that IWA and fill out the fields of BotIWAAnalysis as follows:
iwa (str): Copy the current Candidate IWA verbatim into this field. All of the following fields will be based on this IWA.
iwa_explanation (str): Explain in one sentence what the IWA means in the context of O*NET and what kinds of occupations perform this IWA.
is_match_explanation (str): Explain in one sentence whether the action that the bot is performing in the conversation is an example of a work activity described by the IWA, given the meaning of the IWA in the context of O*NET.
is_match (bool): Based on your explanation, provide the label True if the action that the bot is performing in the conversation is an example of a work activity described by the IWA, given the meaning of the IWA in the context of O*NET, and False otherwise.
automation_level_explanation (str): Consider the full scope of the work performed under this IWA across all occupations. What fraction of this work can the bot perform by applying only the capability it demonstrates in this conversation? Pay careful attention to the fact that the IWA might encompass many more subtasks than represented in this conversation. Explain in one sentence, or reply N/A if the IWA does not match the bot's action (i.e., when is_match is False).
automation_level (IWAAutomationLevel): Based on your explanation, label the bot's capability to perform the IWA using the IWAAutomationLevel enum, which has the following options:
- none: The bot does not perform the IWA, or the conversation does not indicate that the bot is capable of performing the IWA.
- minimal: With this demonstrated capability, the bot can perform a minimal portion of the work in the IWA. 
- limited: With this demonstrated capability, the bot can perform a limited portion of the work in the IWA. 
- moderate: With this demonstrated capability, the bot can perform a moderate portion of the work in the IWA. 
- significant: With this demonstrated capability, the bot can perform a significant portion of the work in the IWA. 
- complete: With this demonstrated capability, the bot can perform all of the work in the IWA.

# Hints
- Provide your answers in **English** using the given structured output format. 
<|end Instruction|>

<|Conversation between User and AI|>
{convo}
<|end Conversation|>

<|Conversation Summary|>
{summary}
<|end Conversation Summary|>

<|Candidate IWAs|>
{iwas}
<|end Candidate IWAs|>

<|end of prompt|>
\end{prompt}

\section{Completion classifier}\label{app:completion}
While the \thumbs{} dataset tells us which work activities receive the most positive user feedback, thumbs feedback may not reflect the success of AI across tasks, as not all types of users give feedback at the same rate (e.g., suppose users who perform some tasks are inherently more critical than those who perform others). To supplement the thumbs feedback data, we therefore also perform task completion classification with an LLM. For each conversation, we ask GPT-4o-mini (version \texttt{gpt-4o-mini-2024-07-18}, API version \texttt{2024-08-01-preview}, temperature 0) \footnote{This task is much simpler than the difficult and ambiguous IWA classification task, hence our use of the smaller model.} if the AI completed the user's task in the conversation. We treat conversations as completed if they are labeled \texttt{complete}, and not completed if they are labeled \texttt{partially\_complete} or \texttt{not\_complete}. For comparison with the E1 measure of Eloundou et al.~\cite{eloundou2024gpts}, we also ask if the AI reduced the time it takes to complete the task by at least 50\%, though we did not end up using that. 

\begin{prompt}{Completion prompt}{}
<|Instruction|>
# Task overview
You will be given a conversation between a User and an AI chatbot.
You will summarize the main task that the user is trying to accomplish in the conversation.
You will also determine whether the AI chatbot is able to complete the task, and if so, whether it reduced the time it takes to complete the task with equivalent quality by at least half.

# Task details
Your task is to fill out the following fields: 
task_summary: Summarize the task the User is trying to accomplish in **English**.
completed_explanation: Explain in one sentence whether the AI chatbot is able to complete the User's task, based on the conversation.
completed: Based on your explanation, provide one of the following labels:
- not_complete: The AI chatbot did not make substantive progress towards completing the User's task.
- partially_complete: The AI chatbot made progress towards completing the User's task, but did not complete it.
- complete: The AI chatbot completed the User's task.
speedup_50pct_explanation: Explain in one sentence whether the AI chatbot reduced the time it takes to complete the task with equivalent quality by at least half. This includes tasks that can be reduced to:
- Writing and transforming text and code according to complex instructions,
- Providing edits to existing text or code following specifications,
- Writing code that can help perform a task that used to be done by hand,
- Translating text between languages,
- Summarizing medium-length documents,
- Providing feedback on documents,
- Answering questions about a document, or
- Generating questions a user might want to ask about a document.
Assume the user is a worker with an average level of expertise in their role trying to complete the given task.
speedup_50pct: Based on your explanation, provide the label True if the AI chatbot reduced the time it takes to complete the task with equivalent quality by at least half, and False otherwise.

# Hints
Provide your answers in **English** using the given structured output format. 
<|end Instruction|>

<|Conversation between User and AI|>
{convo}
<|end Conversation|>

<|end of prompt|>
\end{prompt}

\section{Aggregating from IWAs to occupations.}\label{app:iwa_weights}
To compute IWA importance weights, we first compute a weight for every \emph{task} using the \emph{importance} (average 1--5 importance score given to a task by surveyed incumbents) and \emph{relevance} (fraction of surveyed incumbents who said the task was relevant to their job) scores in O*NET. 
For each task $k$ in SOC occupation $i$, we say $\text{weight}_{ik} = 2^{\text{importance}_{ik}} \cdot \text{relevance}_{ik}$. 
The exponential weighting of importance captures our intuition that the five-point scale spans an extremely broad range that is unreasonably compressed if treated as linear. 
We sum weights over tasks mapping to the same IWA. When a task maps to multiple IWAs, the task weight is applied to each IWA equally.  If an occupation has no ratings for any of its tasks, we assign them all weight $w_{ik} = 1$. If an occupation has ratings for only some of its tasks, we ignore the tasks with missing ratings. Dividing IWA weights by the total weight for an occupation then gives us a proxy measure for how much of a job consists of each of its IWAs. We use $w_{ij}$ to denote this normalized weight of IWA $j$ in job $i$.

In addition, we also compute a second IWA weight $w_{ij}^\text{nonphys}$ to account for the fact that some IWAs like \iwa{Provide general assistance} map to both physical and nonphysical tasks, but applicability to the IWA does not imply AI action applicability to physical tasks. To compute $w_{ij}^\text{nonphys}$, we use GPT-5 to label every O*NET task according to whether it requires touching or moving people or objects, validated by two human annotators ($\kappa>0.8$; prompt and validation details in \appcref{app:physical-tasks}). When summing task weights $w_{ik}$ to get $w_{ij}^\text{nonphys}$, we only include those tasks labeled as nonphysical, but normalize by the weight of all tasks; this makes $w_{ij}^\text{nonphys}$ an estimate of the fraction of a job $i$'s work that is both nonphysical and described by IWA $j$.

\begin{figure}[tbhp]
    \centering
    \includegraphics[width=0.48\linewidth]{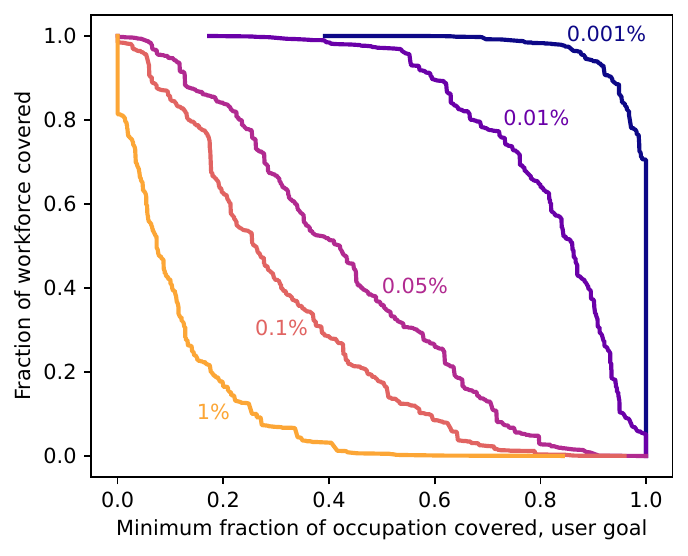}    
    \caption{\textbf{Effect of activity share threshold on absolute impact estimates.} The share of workers who have at least $x\%$ of their work in covered IWAs for different definitions of an IWA being covered ($.001\%, \ldots, 1\%$ of user chat activity).  The resulting numbers depend significantly on the selected threshold, making relative statements more meaningful than absolute coverage numbers. }
    \label{fig:depth-vs-threshold}
\end{figure}

Our approach contrasts with other research that uses a measurement~\cite{handa2025economic} or prediction~\cite{eloundou2024gpts} of the fraction of occupations or of the workforce that have at least $x$\% of their tasks impacted by AI. Such measurements cannot be made reliably from usage data alone, as the selected threshold for usage has a significant impact on the resulting numbers, whose apparent straightforwardness belies this issue. \Cref{fig:depth-vs-threshold} shows that by picking different usage thresholds, we can conclude that either $\sim0$\% or $\sim100\%$ of the workforce has 50\% of its importance-weighted tasks represented in our data, depending on whether we require 1\% of chat activity for a task to be covered or only .01\% of activity. As such, we believe it is much more meaningful to make relative statements about different kinds of occupations, which is what our AI applicability score is designed to do.

\section{Physical tasks classifier}\label{app:physical-tasks}
We used \texttt{GPT-5-2025-08-07}\footnote{Conversation-level classifications were performed before GPT-5 was released; here, we can afford to use a larger, slower, more expensive model as the number of task-level classifications is small compared to the conversation-level classifiers.} to label whether each of 18796 O*NET tasks requires touching or moving physical objects or people. We also had two human annotators label a random sample of 200 tasks. The two annotators agreed with the model on 183 and 184 of the 200 tasks (Cohen's $\kappa$ 0.82 and 0.84), while they agreed with each other on 181 of the tasks (Cohen's $\kappa$ 0.80).

\begin{prompt}{Physical task prompt}{}
# Instructions
You will be given an occupation and task from the O*NET database. Determine whether performing that task requires physical action, defined as touching or moving objects or people.
Interacting with a computer by typing on a keyboard or using a mouse does not count as physical action.
To help illustrate the task, we have provided some examples below.

# Examples
Occupation: Electrician
Task: Work from ladders, scaffolds, or roofs to install, maintain, or repair electrical wiring, equipment, or fixtures.
Is physical action required? Yes, this requires touching and moving objects.

Occupation: Software Developer
Task: Modify existing software to correct errors, adapt it to new hardware, or upgrade interfaces and improve performance.
Is physical action required? No, this does not require touching or moving objects or people.

Occupation: Chief Executives
Task: Direct, plan, or implement policies, objectives, or activities of organizations or businesses to ensure continuing operations, to maximize returns on investments, or to increase productivity.
Is physical action required? No, this does not require touching or moving objects or people.

Occupation: Nurse Practitioners
Task: Perform primary care procedures such as suturing, splinting, administering immunizations, taking cultures, and debriding wounds.
Is physical action required? Yes, this requires touching people.

# Output format
Provide your answer using the given structured output, filling out the fields as follows:
- is_physical_explanation (str): A brief explanation of your reasoning about whether the task requires touching or moving objects or people.
- is_physical (bool): true if physical action is required, false otherwise.

# Your task
Occupation: {occupation}
Task: {task}
Is physical action, defined as touching or moving objects or people, required?
\end{prompt}

\section{Contributions of AI applicability score components}\label{app:ai-app-components}
Here, we discuss how each component of AI applicability score contributes to the final occupational scores. We thank the anonymous reviewer who suggested this analysis.\

The primary signal from AI applicability comes from coverage: the Kendall tau correlation between a coverage-only occupation ranking and the AI applicability ranking is 0.88. Adding in scope brings it up to 0.95, with completion providing the last 0.05.

The pairwise swaps induced by completion and scope are important, though concentrated on a smaller number of occupations. The occupations whose scores are most changed by including completion and scope are artistic (e.g., Special Effects Artists and Animators, Craft Artists, Graphic Designers, Mechanical Drafters). By coverage alone, these are all in the top 3\% of occupations--but due to the low completion of visual design work activities, their AI applicability scores end up ranked in the middle (top 25-55\%), a big shift. (Note that other AI tools developed since 2024 may have different success rates in this area.) In the other direction, completion and scope also increase the rankings of some occupations with particularly successful work activities: Telemarketers (\#69 to \#8), Tax Examiners (\#168 to \#32), Customer Service Representatives (\#35 to \#7), and Public Relations Specialists (\#100 to \#18). These largely involve highly successful communication-based work activities.

\begin{figure}
    \centering
    \includegraphics[width=0.49\linewidth]{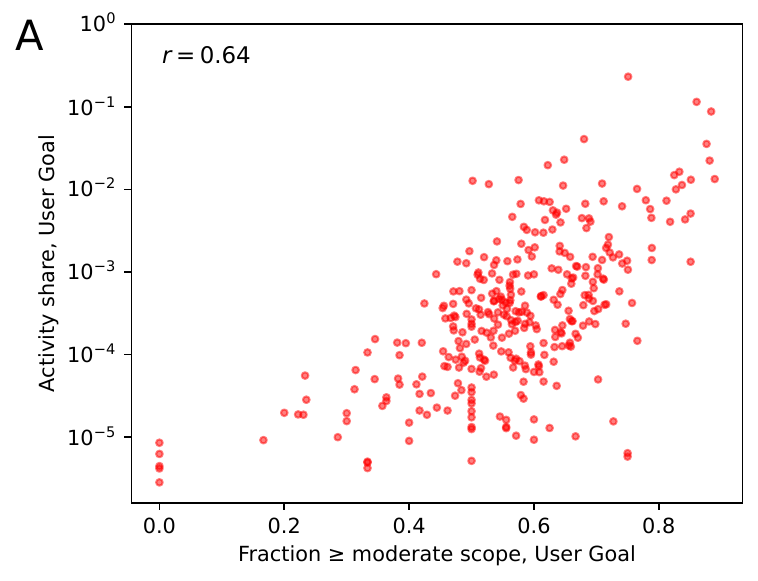}
        \includegraphics[width=0.49\linewidth]{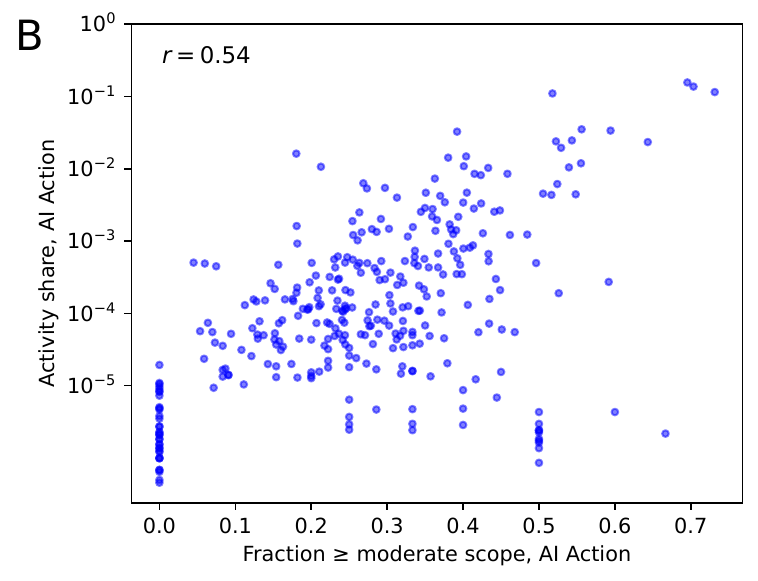}
    \caption{ \textbf{Correlation between scope and activity share in \unif{} for each IWA.} (\textbf{A}) Scatterplot of user goal IWA activity share and the fraction of conversations in which the scope is classified moderate or higher. (\textbf{B}) The same scatterplot for AI action IWAs. On both sides, impact scope is a good predictor of what activities people seek AI assistance with (better than completion or satisfaction).}
    \label{fig:scope-vs-activity-share}
\end{figure}

One of the reasons that usage is such a strong signal is that, at scale, it already captures some element of success: if many users continue to use Copilot for some work activity, then it is likely helpful for that work activity. This is reflected in the correlation between scope and activity share (\Cref{fig:scope-vs-activity-share}). This is also reflected in the Kendall tau correlation of 0.64 between a scope-only ranking and the AI applicability ranking.

\section{Information work classifier}\label{app:information-work}
\begin{figure}[htbp]
    \centering
    \includegraphics[width=0.95\linewidth]{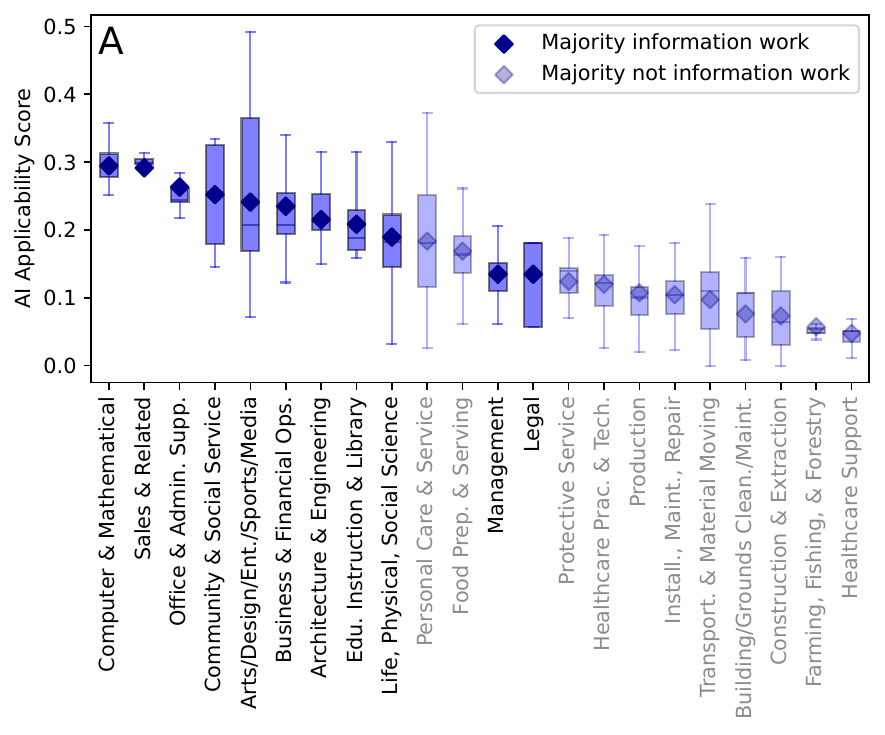}
    \caption{\textbf{AI applicability score for major occupation groups.} The employment-weighted mean, median, and inter-quartile range of the AI applicability score for each major occupation group. Highlighted occupational groups have a majority of workers in occupations classified as information work. 
    }
    \label{fig:soc-major}  
\end{figure}

To highlight information work occupations in \Cref{tab:soc-minor-group-all} and \Cref{fig:soc-major}, we used \texttt{GPT-5-2025-08-07} to label whether each SOC occupation consisted primarily of information work and/or knowledge work. See the prompt below for the definitions we used. As a robustness check, we also repeated the classification with the explicit definitions replaced by an instruction to use ``standard economic definitions'' of information and knowledge work. Agreement between the two classification runs was 86\% and 91\% for information work and knowledge work, respectively.

\begin{prompt}{Information work prompt}{}
# Instructions
You will be given a detailed SOC occupation. You tasks are to determine if that occupation primarily consists of information work and, separately, whether it primarily consists of knowledge work.
Information work is labor in which workers create, process, or communicate information rather than produce physical goods or provide physical services.
Knowledge work is labor in which workers apply knowledge, expertise, and judgment in non-routine and creative ways to solve complex problems, make decisions, or generate new ideas.

# Output format
Provide your answer using the given structured output, filling out the fields as follows:
- is_information_work_explanation (str): A brief explanation of your reasoning about whether the occupation primarily consists of information work.
- is_information_work (bool): true if the occupation primarily consists of information work, false otherwise.
- is_knowledge_work_explanation (str): A brief explanation of your reasoning about whether the occupation primarily consists of knowledge work.
- is_knowledge_work (bool): true if the occupation primarily consists of knowledge work, false otherwise.

# Task
The occupation in question: {soc_code} {occupation}    
\end{prompt}

\section{Employment weighting and socioeconomic correlates}\label{app:employment-weighting}

Most prior work did not weight occupations by employment when examining the relationship between AI exposure and wage. Since occupations vary a lot in size and the boundaries are somewhat subjective (e.g., Cooks get separate occupations for Short Order, Restaurant, Institution and Cafeteria, and Fast Food, but Maids and Housekeeping Cleaners are one category), results weighted by employment better answer the research question about overall workforce relationship between wages and occupational applicability of AI.
Nonetheless, \Cref{fig:wage-vs-occupation-coverage} shows the correlation between AI applicability score and average occupation wage without employment weighting, which increases the correlation to 0.17 for user goals and 0.32 for AI actions. This higher correlation without employment weighting indicates that prior findings may be due in part to a lack of employment weighting.
The difference between the weighted and unweighted results is primarily due to high-employment Sales and Office and Administrative Support occupations that have relatively low wages, but high AI applicability. 
There is a lot of variation across occupations and some occupations have much higher AI applicability than others, but the overall relationship between wage and AI applicability is weak. 

\begin{figure}
    \centering
    \includegraphics[width=0.49\linewidth]{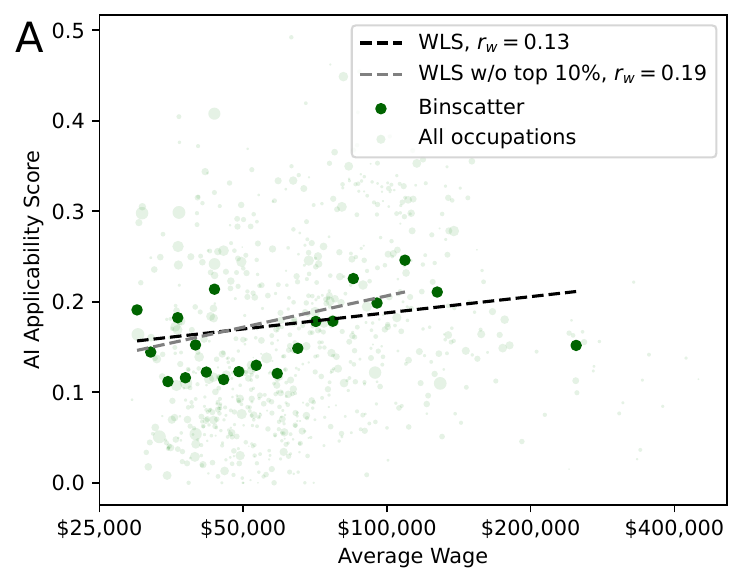}   
        \includegraphics[width=0.49\linewidth]{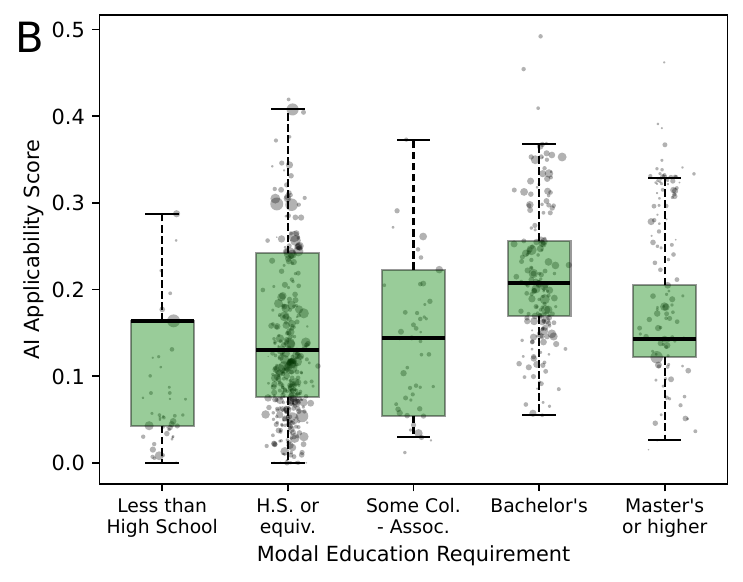}   
    \caption{\textbf{AI applicability scores by average wage and educational requirement.}
    (A) AI applicability score by average occupational wage. There is a point for each occupation and darker points for the employment-weighted average of each ventile. Binscatter conditional means and least squares fits are weighted by employment. 
    (B) AI applicability score by modal education requirement reported to O*NET. Boxplot medians and quartiles are weighted by employment (e.g., they show the score of the median worker rather than that of the median occupation).
    }
    \label{fig:wages-education}
\end{figure}

\begin{figure}[htbp]
    \centering
    \includegraphics[width=0.49\linewidth]{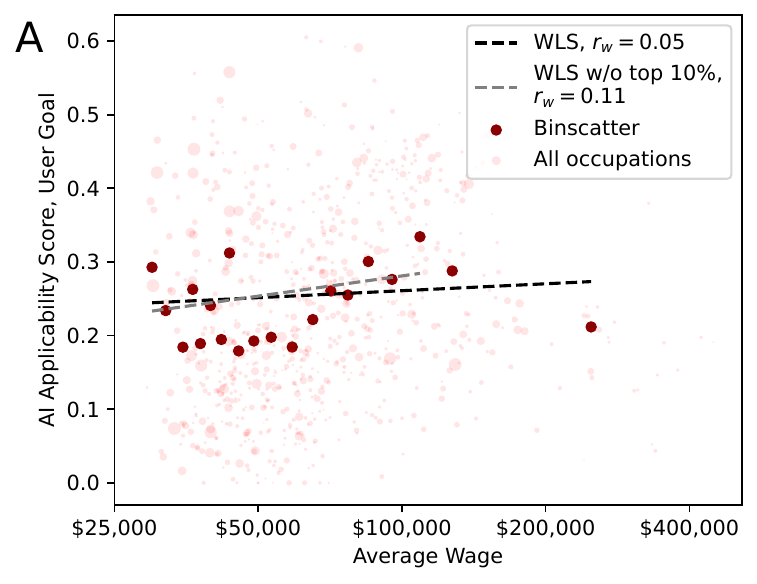}
        \includegraphics[width=0.49\linewidth]{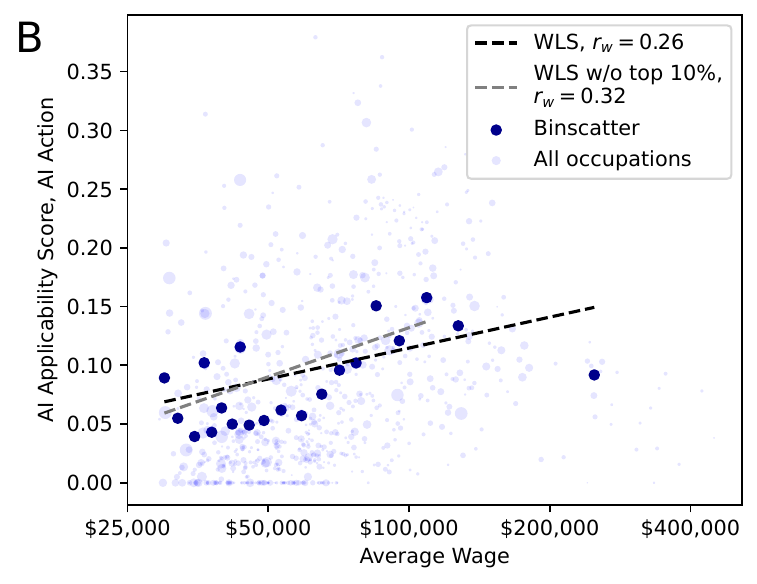}
    \caption{\textbf{AI applicability score by wage, separating out user goals and AI actions.} (\textbf{A}) Wage plot as in \Cref{fig:wages-education}A, but showing user goal AI applicability score only. (\textbf{B}) AI action applicability score only. The correlation is higher on the AI side.}
    \label{fig:wage-vs-occupation-coverage-weighted}
\end{figure}

\begin{figure}[h!]
    \centering
    \includegraphics[width=0.49\linewidth]{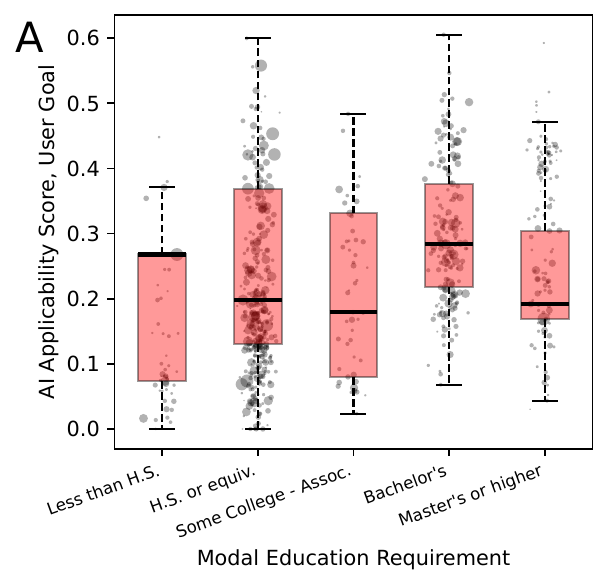}
        \includegraphics[width=0.49\linewidth]{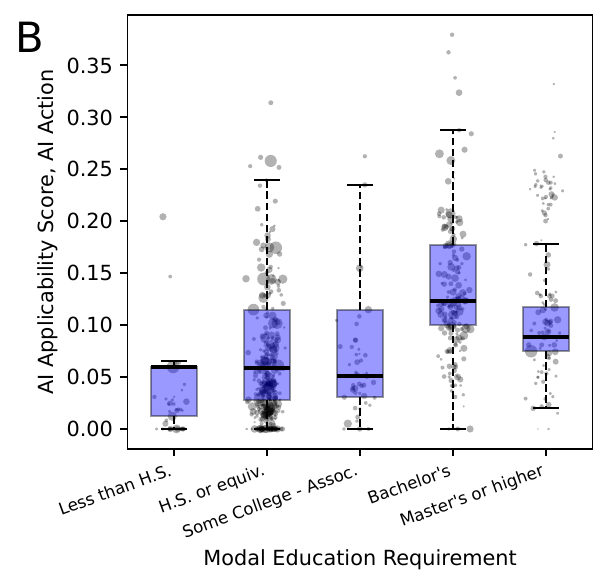}    \caption{\textbf{AI applicability score by educational requirement, separating out user goals and AI actions.} (\textbf{A}) Education plot from \Cref{fig:wages-education}B, but showing only user goal AI applicability score. (\textbf{B}) Using AI action scores. As with wage, the relationship with education is  stronger on the AI side.}
    \label{fig:education-vs-coverage-weighted}
\end{figure}

\begin{figure}
    \centering
    \includegraphics[width=0.49\linewidth]{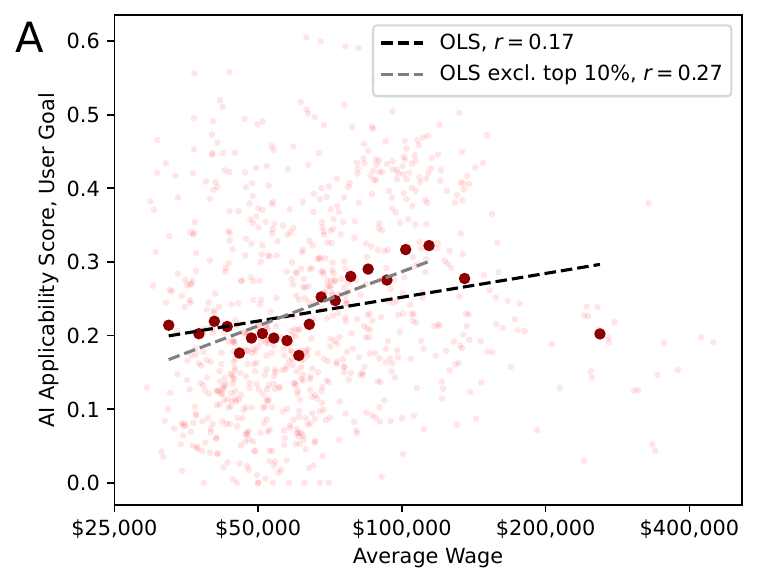}
        \includegraphics[width=0.49\linewidth]{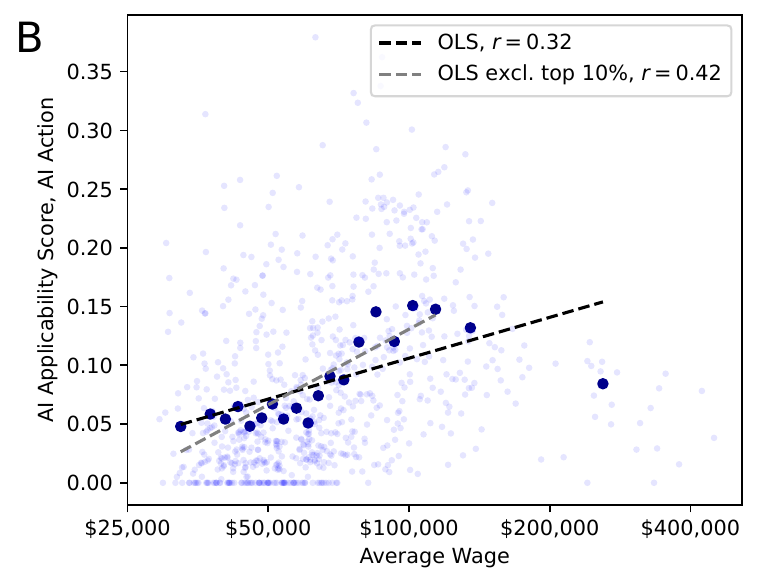}
\caption{\textbf{AI applicability score by wage, unweighted}.  This Figure is the same as \Cref{fig:wage-vs-occupation-coverage-weighted} without weighting the binscatters by employment. (\textbf{A}) Scatterplot with unweighted binscatter of each occupation's AI applicability score calculated over user goals only against the  occupation's average wage. (\textbf{B}) Using AI action AI applicability score.
    The relationship is much less noisy without employment weighting, likely because weighting by employment causes the binscatters to be influenced dramatically by a small number of high-employment occupations, increasing the variance due to noise in the coverage metric and in the mapping between occupations and IWAs.  
    Excluding the top 10\% of highest-paid workers strengthens the correlation.}
    \label{fig:wage-vs-occupation-coverage}
\end{figure}

\begin{figure}
    \centering
    \includegraphics[width=0.49\linewidth]{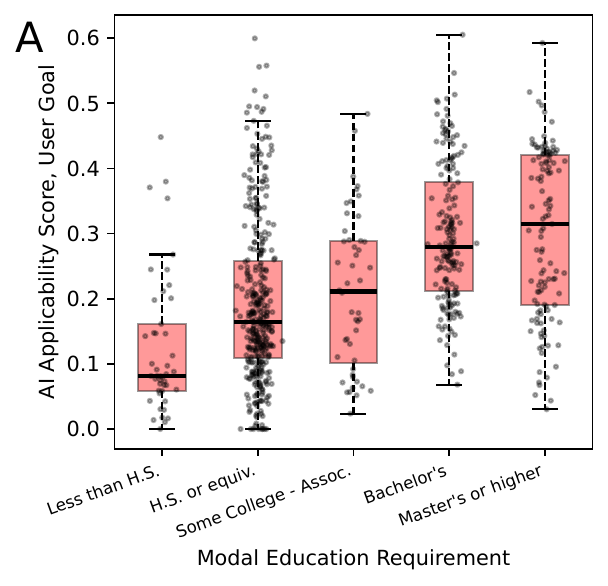}
        \includegraphics[width=0.49\linewidth]{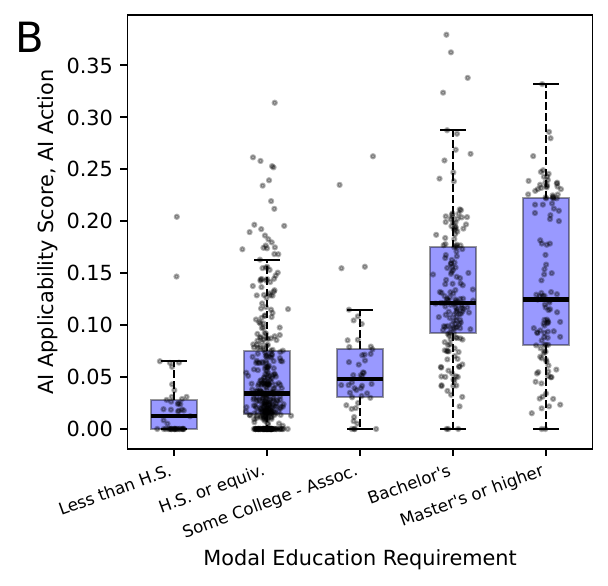}
    \caption{\textbf{AI applicability score by educational requirement, unweighted.} This Figure is the same as \Cref{fig:education-vs-coverage-weighted} without weighting by employment. (\textbf{A}) User goal AI applicability scores over occupations with different educational requirements, with boxplots over occupations (not weighting by employment). (\textbf{B}) Using AI action scores. As in \Cref{fig:wage-vs-occupation-coverage}, this reduces the noise in the relationship.}
    \label{fig:education-vs-coverage}
\end{figure}

The same pattern also occurs with educational requirements: without employment weighting, the relationship between AI applicability score and higher education requirements appears stronger (\Cref{fig:education-vs-coverage}).
\Cref{fig:wage-vs-occupation-coverage-weighted,fig:education-vs-coverage-weighted} separate out the (employment-weighted) wage and education plots from \Cref{fig:wages-education} into user goals and AI actions; the patterns are similar on each side, although the relationships are slightly stronger on the AI side.

\section{Broader impacts}\label{app:broader-impacts}
We consider broader impacts of our work on users of Microsoft Bing Copilot whose usage comprises our datasets and on society more broadly.

Regarding Bing Copilot users, we took several steps to ensure that their privacy was maintained: (1) obtaining IRB approval for our analyses, (2) automatically scrubbing conversation texts of personally identifiable information, (3) storing and processing conversation data in secure computing environments, and (4) reporting only highly aggregated metrics that do not reveal anything about individual users. Given these steps, we believe the risk of harm to users from our research is low.

Regarding society more broadly, our work provides visibility into how a generative AI system was used, data that are usually proprietary and opaque. We believe it is beneficial for society to have a better understanding of how this important technology is being used and how it may affect different types of work. More transparency and research in this area will allow the public to make more informed decisions about their education, career, and usage of AI. It also enables more informed decisions by policymakers. 

However, the high societal importance of our research also carries a heavy burden of responsibility. If our analysis contains errors or is  misleading or unclear, this could misinform rather than inform the public and policymakers, leading to worse decisions at a critical time in the development of this new technology. Due to this risk, we have sought to adhere to the highest level of rigor, clarity, and transparency possible. On balance, we believe the benefits of this knowledge outweigh the risks.

\section{Supplementary figures}\label{app:plots}

\begin{figure}[h!]
    \centering
    \includegraphics[width=\linewidth]{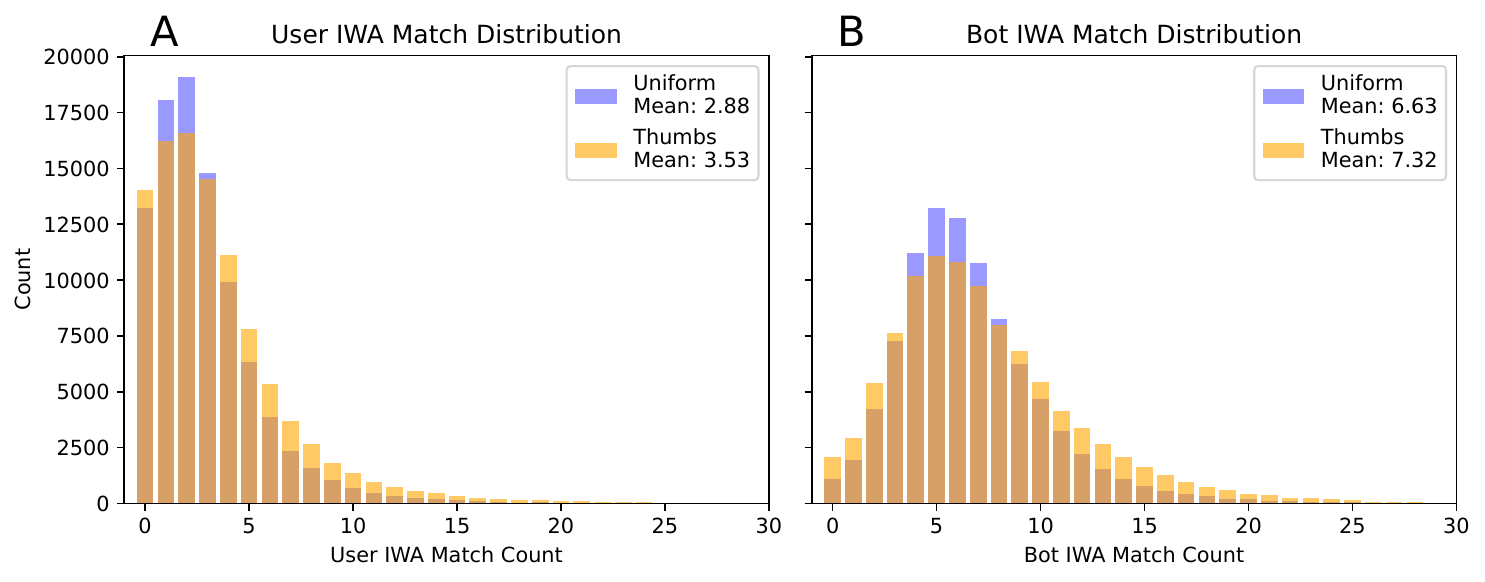}
    \caption{\textbf{Distribution of IWAs per conversation.} (\textbf{A}) The distribution of the number of IWAs to which a conversation is matched for user goal in both the uniform and thumbs Copilot datasets. (\textbf{B}) The same distributions for AI action IWAs.}
    \label{fig:per-convo-matches}
\end{figure}

\begin{figure}[h!]
    \centering
    \includegraphics[width=\linewidth]{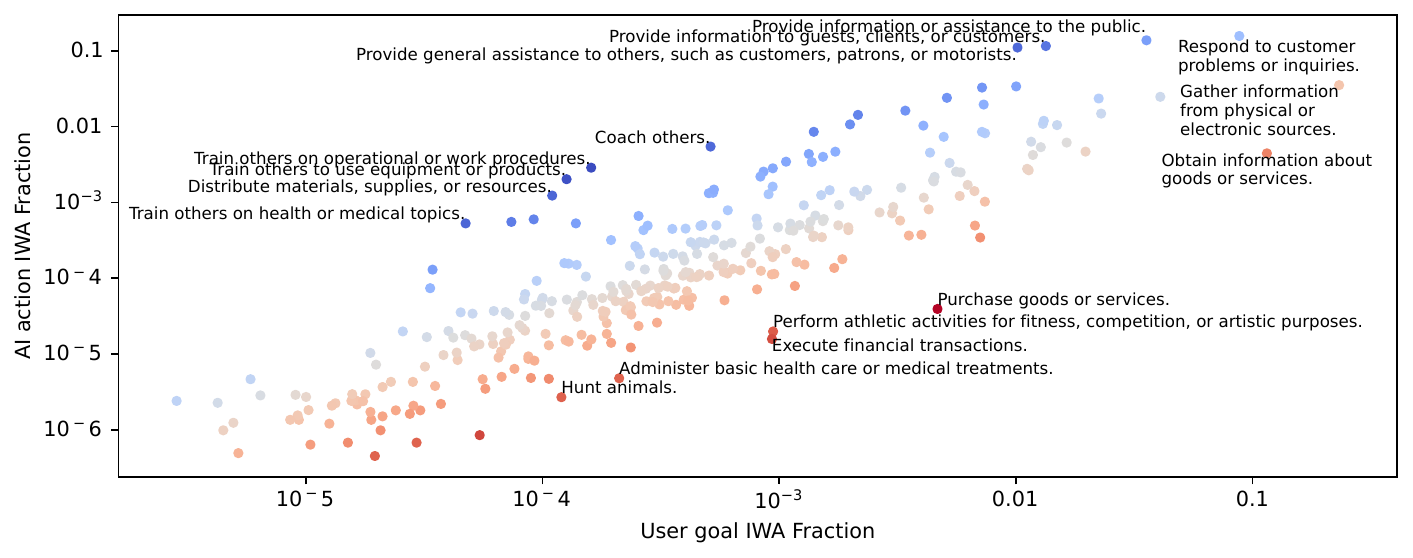}
    \caption{\textbf{IWA frequency in AI actions and user goals.} For each IWA, this Figure plots the fraction of user intents ($x$ axis) and AI actions ($y$ axis) described by that IWA. Outliers show which IWAs are more likely to be performed (blue) or assisted (red) by Bing Copilot. When a conversation is labeled with multiple IWAs, that share of Copilot activity is evenly distributed among the IWAs so that the sum of all IWA fractions is 1. \Cref{tab:user-bot-ratios} lists the relative frequencies for the most off-diagonal IWAs. }
    \label{fig:user-bot-counts}
\end{figure}

\begin{figure}[h]
    \centering
    \includegraphics[width=\linewidth]{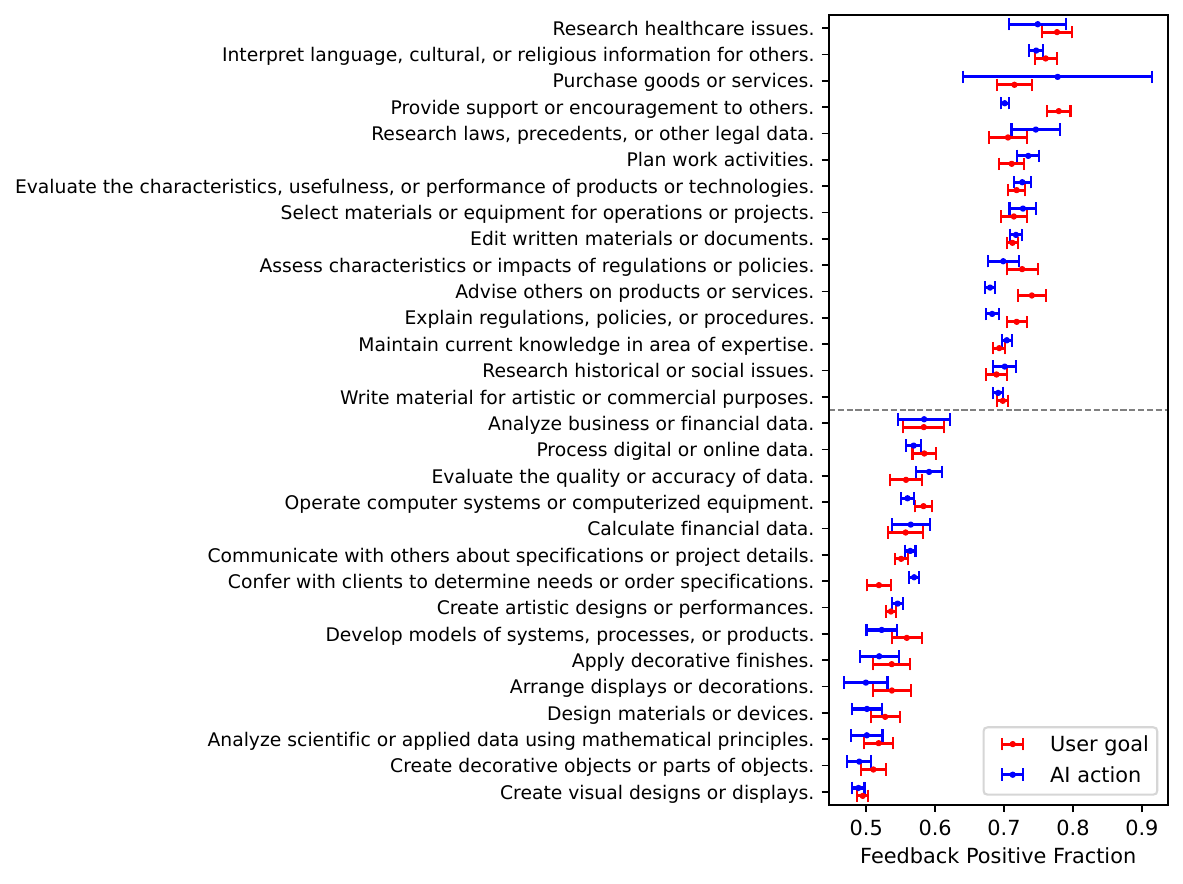}
    \caption{\textbf{IWAs with the highest and lowest shares of positive feedback.}
     This Figure shows the top and bottom 15 IWAs by the share of positive feedback,  filtered to common IWAs matched in at least 1\% of conversations in our feedback dataset,  with bootstrapped 95\% confidence intervals. The common IWAs with highest positive feedback share include two about writing and editing; four about evaluating or purchasing goods and services; and six about researching information about health, culture, law, policy, and society. Meanwhile, the common IWAs with the lowest positive feedback share include five visual design and five data analysis IWAs. }
    \label{fig:thumb-rate-top-bottom-iwas}
\end{figure}

\begin{figure}[htbp]
    \centering
    \includegraphics[width=\linewidth]{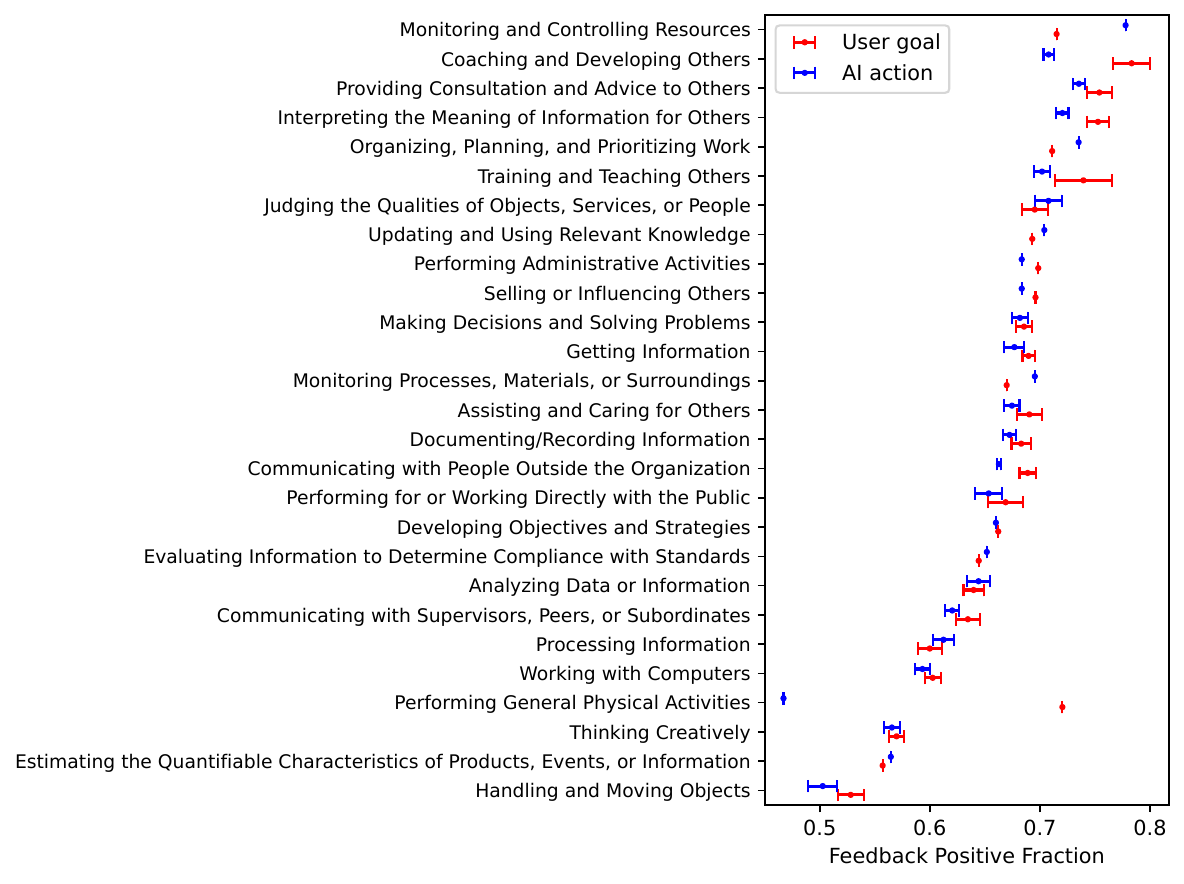}
    \caption{\textbf{Share of positive feedback by GWA.} This Figure plots the positive feedback share for each GWA, aggregating common IWAs into their GWAs and with bootstrapped 95\% confidence intervals; any IWA appearing in less than 1\% of our feedback data is ignored. 14 GWAs have no common IWAs and are thus excluded from this plot, including those relating to operating vehicles, repairing equipment, and management tasks like hiring, negotiation, and guiding subordinates.}
    \label{fig:thumb-rate-top-bottom-gwas}
\end{figure}

\begin{figure}[htbp]
    \centering
    \includegraphics[width=0.49\linewidth]{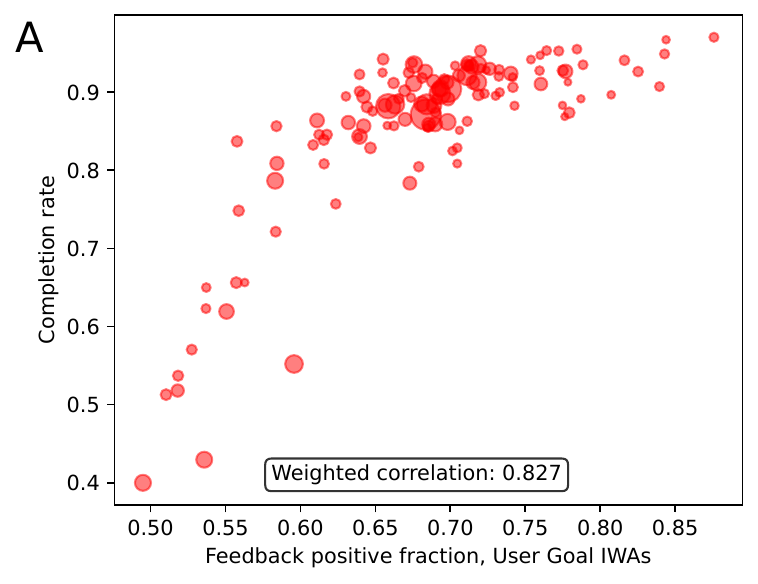}
        \includegraphics[width=0.49\linewidth]{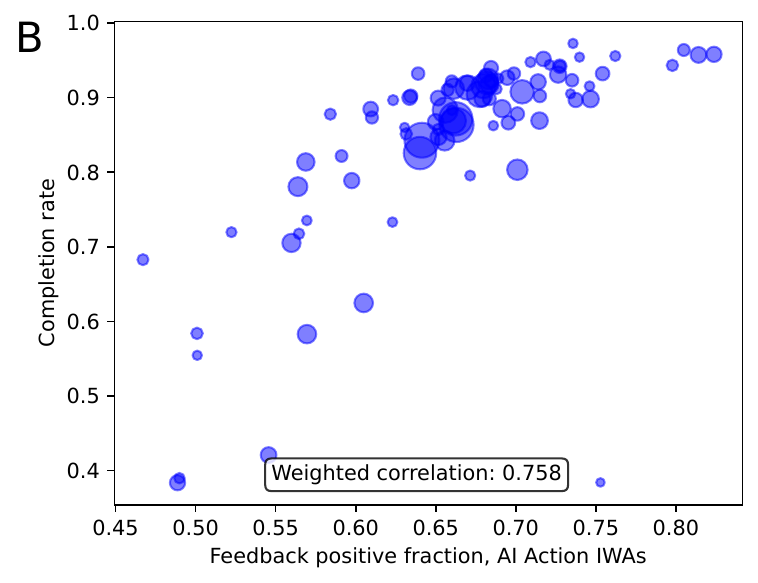}
    \caption{\textbf{Relationship between thumbs feedback and task completion rate for each IWA.} (\textbf{A}) Scatterplot of user goal IWA positive feedback share (measured in \thumbs{}) and completion rate (measured in \unif{}). (\textbf{B}) The same scatterplot for AI action IWAs.  There is a strong correlation between thumbs feedback rate  and GPT-4o-mini task completion rate  for each IWA, indicating that both are capturing real signal about AI success in assisting or performing an IWA. Point size proportional to square root of IWA match count in \unif{}. Weighted correlations also weighted by match count in \unif{}.}
    \label{fig:thumbs-vs-completion}
\end{figure}

\begin{figure}[htbp]
    \centering
    \includegraphics[width=0.49\linewidth]{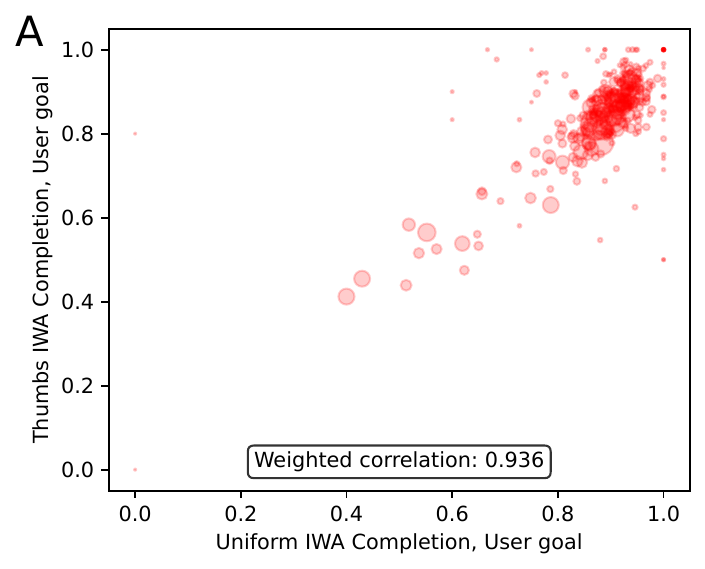}
        \includegraphics[width=0.49\linewidth]{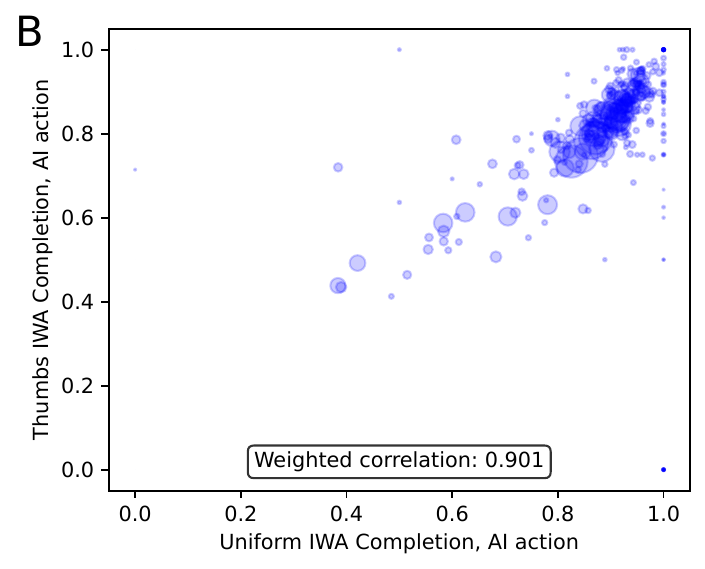}
    \caption{\textbf{Correlation between IWA completion rates in \unif{} and \thumbs{}.} (\textbf{A}) Scatterplot of user goal IWA completion rate in our two datasets. (\textbf{B}) The same scatterplot for AI action IWAs.  IWA-level completion rates are consistent between \unif{} and \thumbs{}. Point size proportional to square root of IWA match count in \unif{}. Weighted correlations also weighted by match count in \unif{}.}
    \label{fig:completion-uniform-vs-thumbs}
\end{figure}

\begin{figure}[htbp]
    \centering
    \includegraphics[width=0.9\linewidth]{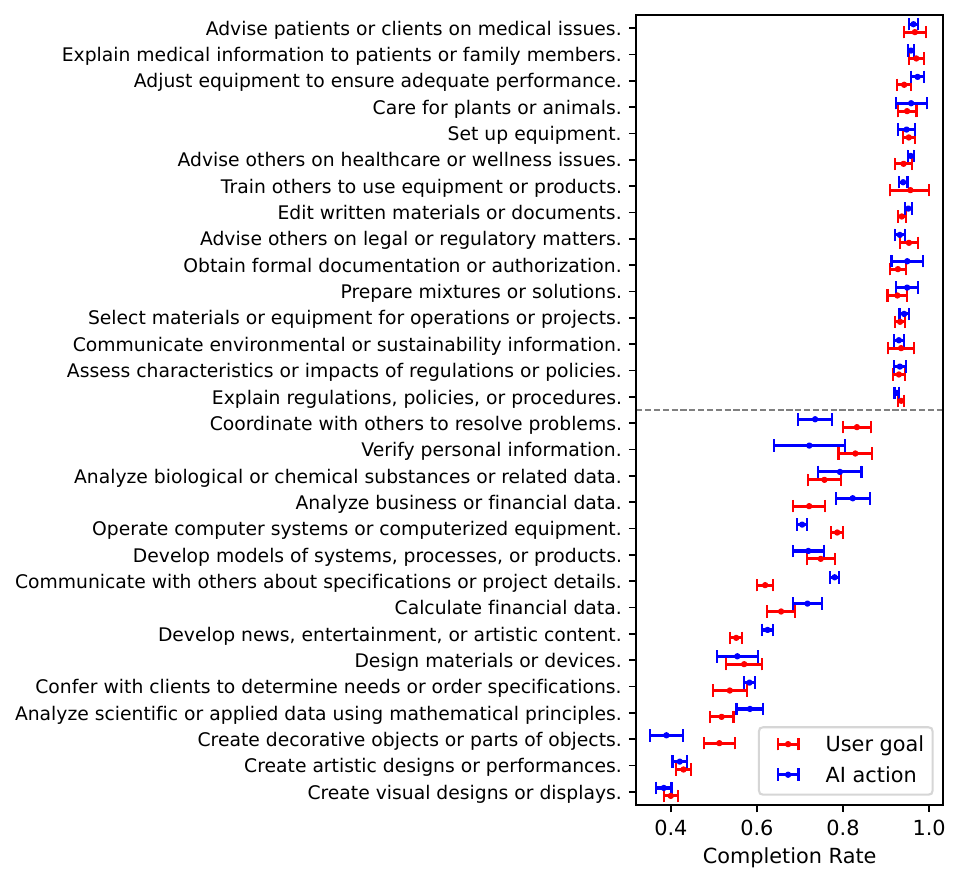}
    \caption{\textbf{IWAs with the highest and lowest completion rates.} This Figure shows the top and bottom 15 IWAs by completion rate,  filtered to `common' IWAs with activity share at least 0.1\% in \unif{},  with 95\% confidence intervals (using the normal approximation to binomial confidence intervals). Task completion shows some of the same patterns as positive feedback fraction (\Cref{fig:thumb-rate-top-bottom-iwas}), with visual design and data analysis on the low end and writing on the high end. One notable difference is that the highest completion rate IWAs include 7 about explaining, advising, or training.}
    \label{fig:completion-rate-top-bottom-iwas}
\end{figure}

\begin{figure}[htbp]
    \centering
    \includegraphics[width=0.9\linewidth]{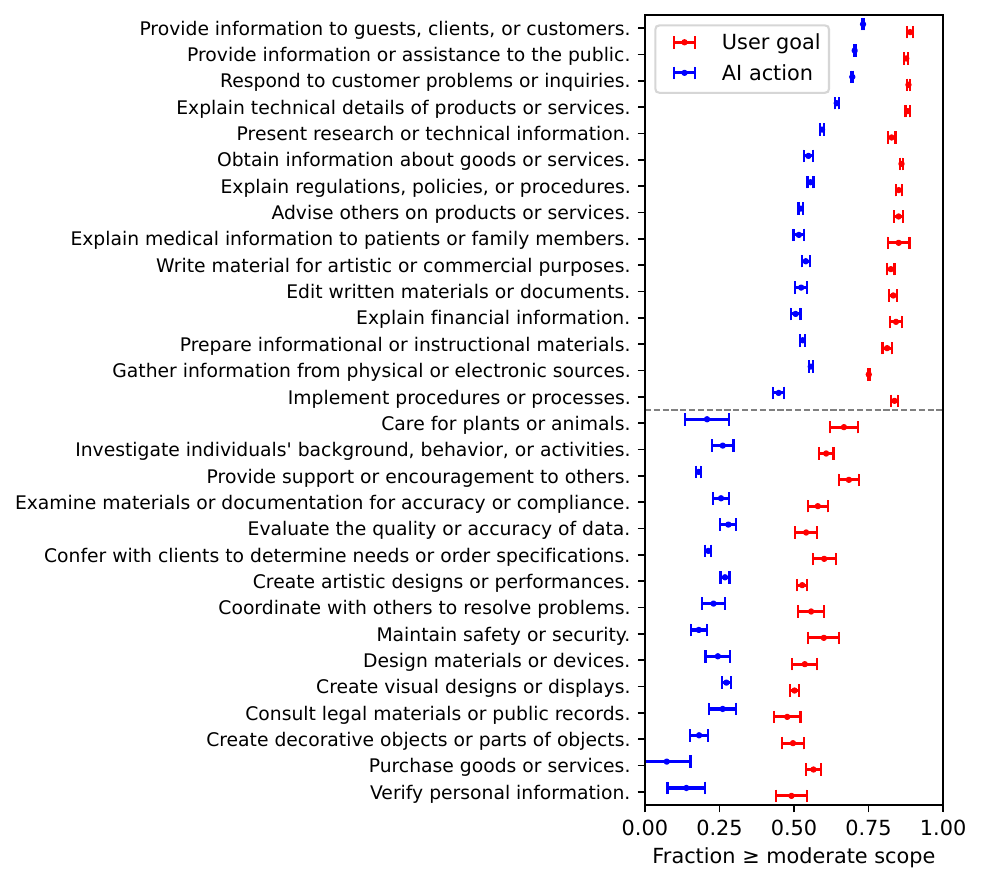}
    \caption{\textbf{IWAs with the highest and lowest fraction of conversations at moderate or higher  scope.} This Figure shows the top and bottom 15 IWAs by how often they are assigned scope of impact at least moderate,  filtered to ``common'' IWAs with activity share at least 0.1\% in \unif{},  with 95\% confidence intervals (using the normal approximation to binomial confidence intervals). Some patterns mirror those of thumbs feedback and completion, with research and writing IWAs having high scope, and data analysis and visual design IWAs having low scope. AI performance consistently has lower impact scope than user assistance.}
    \label{fig:scope-top-bottom-iwas}
\end{figure}

\begin{figure}
    \centering
    \includegraphics[width=0.49\linewidth]{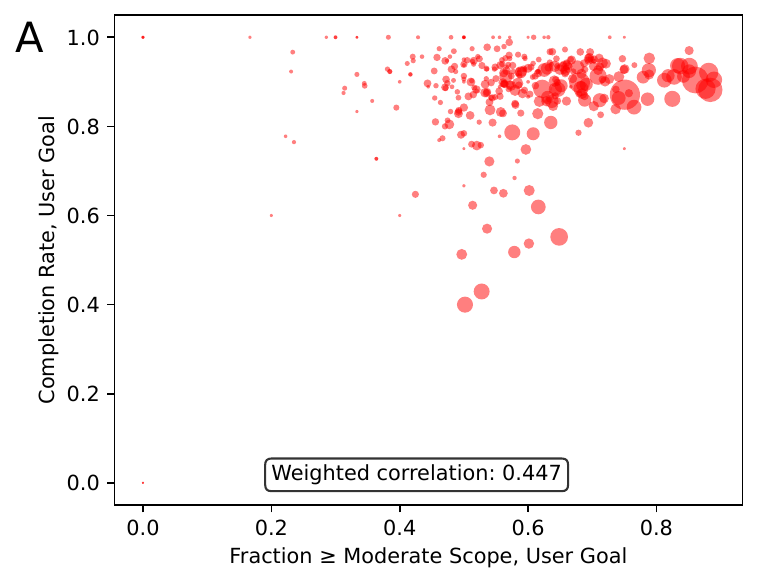}
        \includegraphics[width=0.49\linewidth]{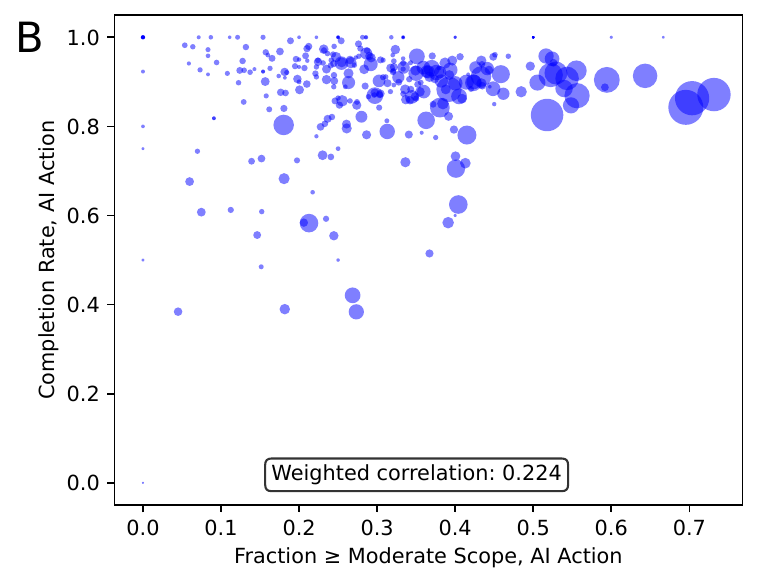}
    \caption{\textbf{Relationship between scope and task completion rate for each IWA.} (\textbf{A}) Scatterplot of user goal IWA moderate or higher scope share and completion rate. (\textbf{B}) The same scatter plot for AI action IWAs. The relationship between scope and completion is much weaker than between completion and thumbs feedback, as they are capturing different questions. Point size proportional to square root of IWA match count in \unif{}. Weighted correlations also weighted by match count in \unif{}.}
    \label{fig:scope-vs-completion}
\end{figure}

\begin{figure}[htbp]
    \centering
    \includegraphics[width=0.95\linewidth]{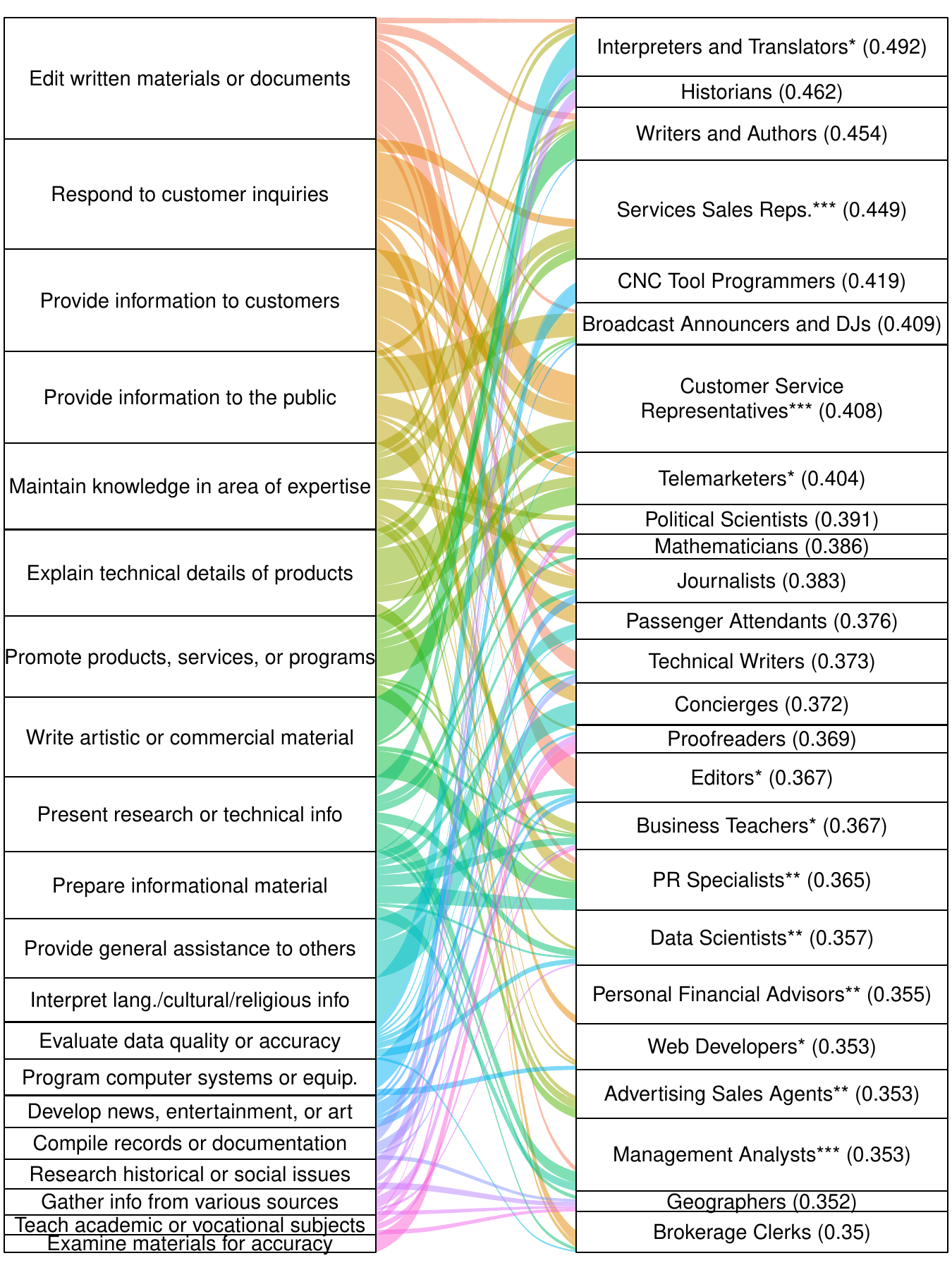}
    \caption{\textbf{Top 25 occupations by AI applicability score and their contributing IWAs.} Similar to \Cref{fig:sankey}, the 25 SOC occupations with the highest AI applicability scores (right) along with the 20 IWAs that provide the greatest contributions to those scores (left). Asterisks indicate 2023 U.S.\ occupational employment (none: up to 50k, one: 50k to 100k, two: 100k to 500k, three: more than 500k); the heights of each stratum indicate the same. Occupations' applicability scores are in parentheses, and occupations are sorted by their score in decreasing order. 
    Portions of the occupational strata not connected to an IWA by a colored flow represent IWAs not present in the Figure that still contribute to the occupation's applicability score. Both occupation and IWA titles have been shortened for space.}
    \label{fig:sankey25}
\end{figure}

\begin{figure}
    \centering
    \includegraphics[width=\linewidth]{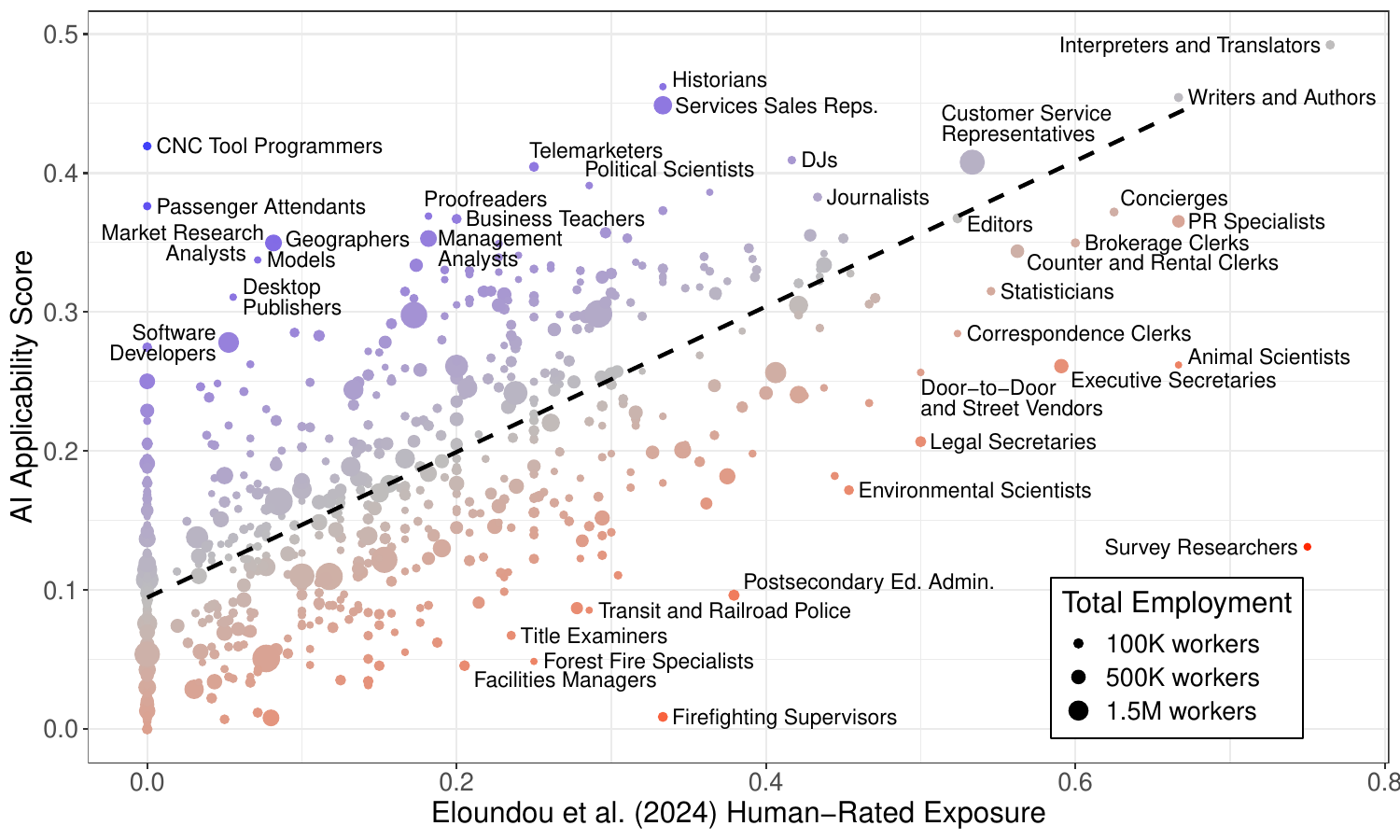}   
    \caption{\textbf{AI applicability scores by predicted exposure}
    AI applicability score versus E1 exposure from Eloundou et al.~\cite{eloundou2024gpts}, the fraction of an occupation's tasks for which human raters predicted LLMs could reduce completion time by 50\%. Occupations are colored by their distance from the employment-weighted regression line: blue points have higher AI applicability scores than would be expected from their E1 exposure.
    }
    \label{fig:prediction}
\end{figure}

\begin{figure}
    \centering
    \includegraphics[width=\linewidth]{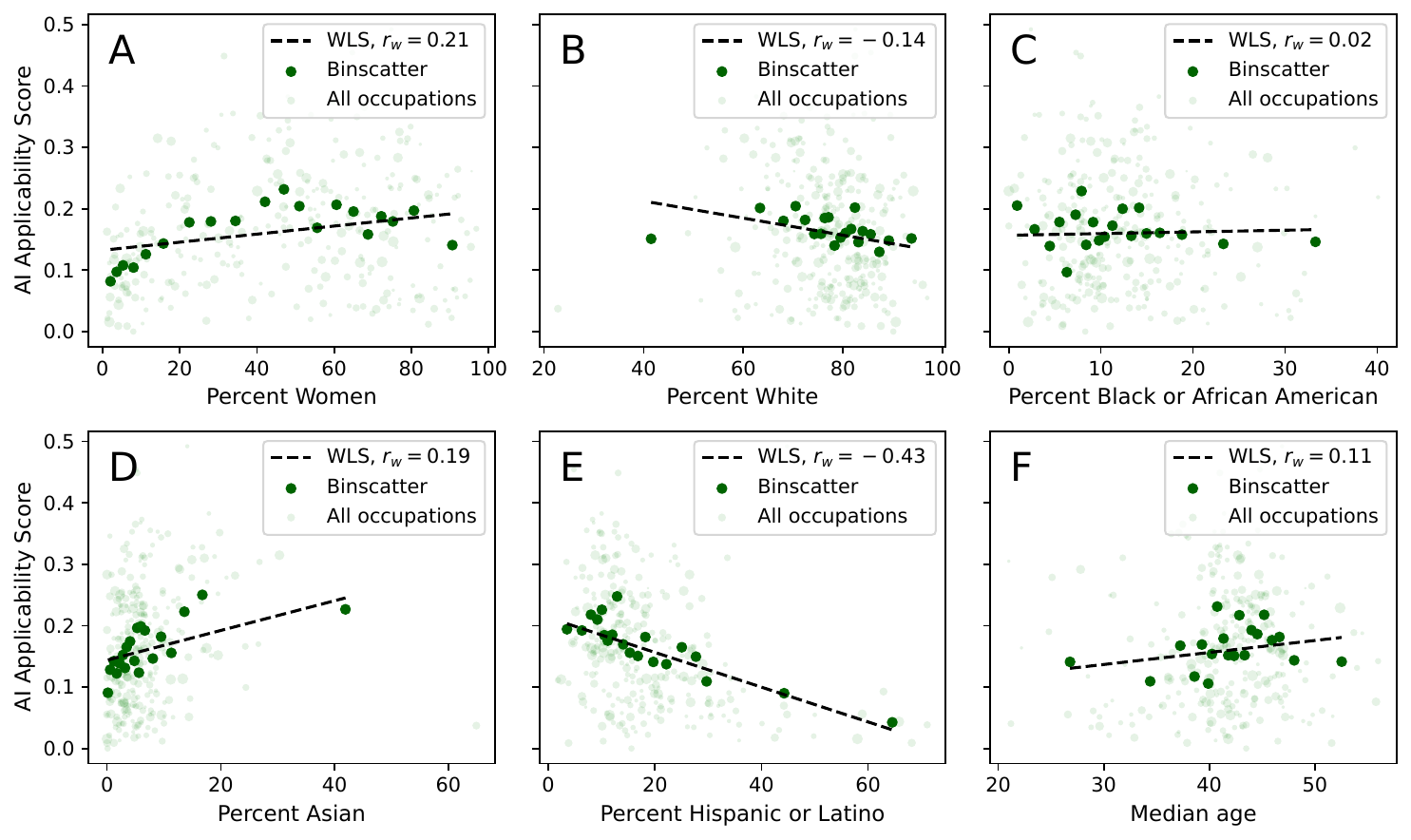}
    \caption{\textbf{Correlations between AI applicability score and 2024 Current Population Survey occupational demographics,  weighted by employment.} (\textbf{A}) Correlation with percent women. (\textbf{B}) Correlation with percent white. (\textbf{C}) Correlation with percent black or African American. (\textbf{D}) Correlation with percent Asian. (\textbf{E}) Correlation with percent Hispanic or Latino. (\textbf{F}) Correlation with median age. For each panel, we drop occupation codes with missing demographic data in the 2024 CPS Annual Averages.}
        \label{fig:demographics-ai-app}

\end{figure}

\begin{figure}[tbhp]
    \centering
    \includegraphics[width=0.48\linewidth]{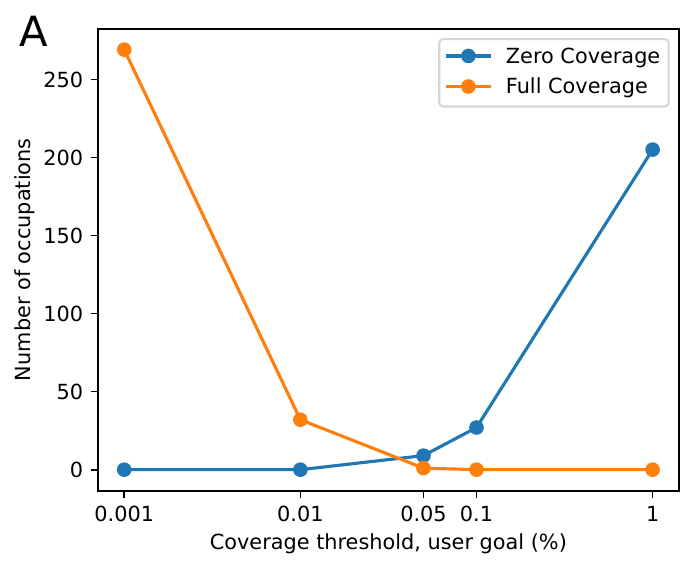}
    \includegraphics[width=0.48\linewidth]{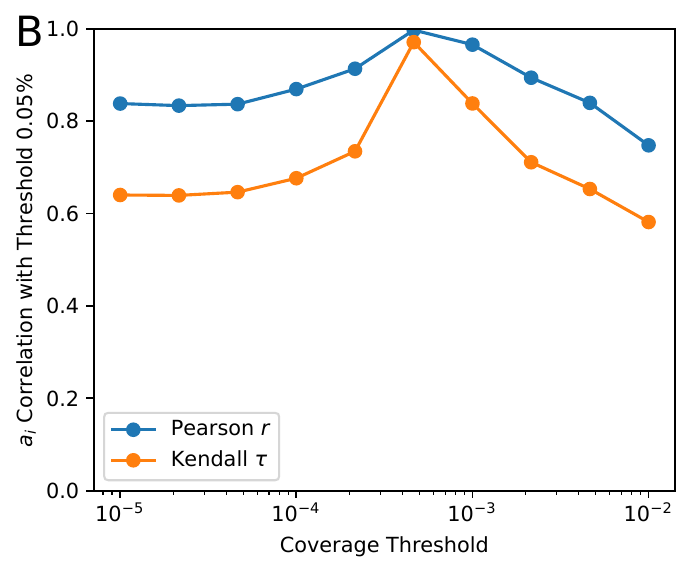}

    \caption{\textbf{Rationale for our chosen coverage threshold.} (A) Effects of coverage threshold on number of occupations with user goal coverage 0 and 1. Our threshold of 0.05\% approximately minimizes the number of occupations assigned user goal coverage 0 or 1. (B) The correlation between AI applicability scores defined using different coverage thresholds and the one we report with threshold 0.05\%. Across thresholds spanning three orders of magnitude, we get strongly correlated rankings of occupations by AI applicability. Contrast this robustness of relative applicability score with the absolute measure in \Cref{fig:depth-vs-threshold}, which is highly sensitive to arbitrary choice of threshold.}
    \label{fig:threshold-selection}
\end{figure}

\begin{figure}[tbhp]
    \centering
    \includegraphics[width=0.49\linewidth]{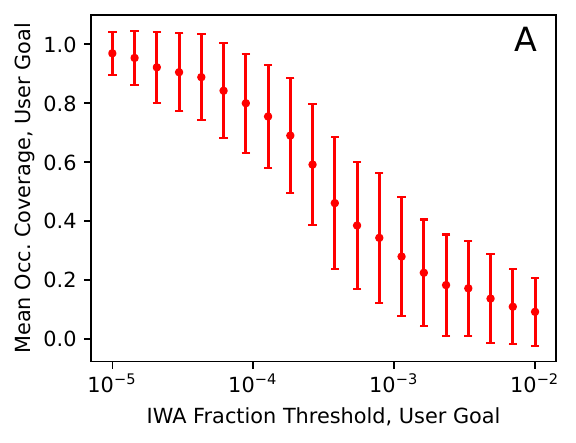}
        \includegraphics[width=0.49\linewidth]{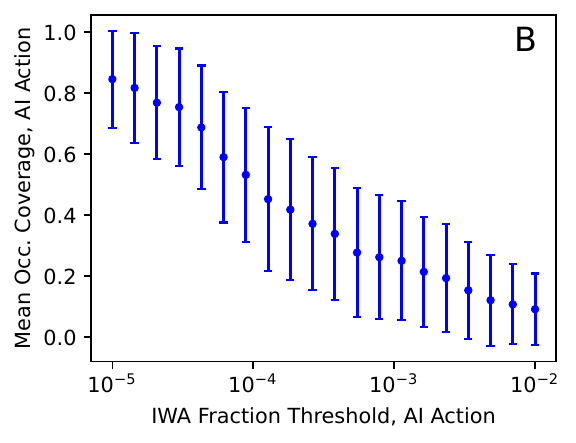}
    \caption{\textbf{Mean and standard deviation of occupation coverage by threshold.} (\textbf{A}) The averages and standard deviations of occupation user goal coverage for different thresholds for the share of chat activity an IWA must have to ``be done'' with the LLM. (\textbf{B}) The same means and standard deviations for coverage derived from AI action IWAs. We use a threshold of 0.0005, which results in 127 and 87 IWAs being considered covered (of 332) on the user goal and AI action sides, respectively.}
    \label{fig:coverage-by-threshold}
\end{figure}

\begin{figure}[tbhp]
    \centering
    \includegraphics[width=\linewidth]{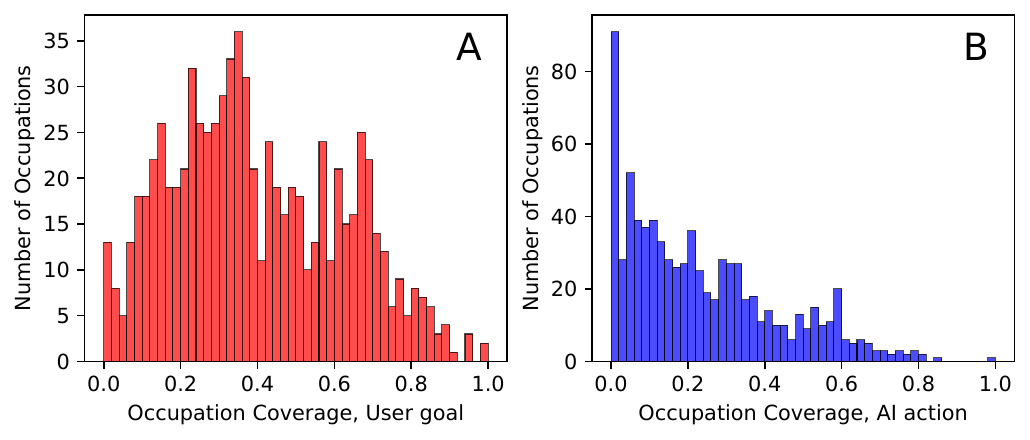}
    \caption{\textbf{Distribution of occupation coverage.} (\textbf{A}) The distribution of occupation user goal coverage scores (fraction of importance-weighted work in an occupation with user goal activity share at least 0.05\% in \unif{}). (\textbf{B}) The same distribution for AI action coverage.}
    \label{fig:soc-coverage-hists}
\end{figure}

\begin{figure}
    \centering
    \includegraphics[width=\linewidth]{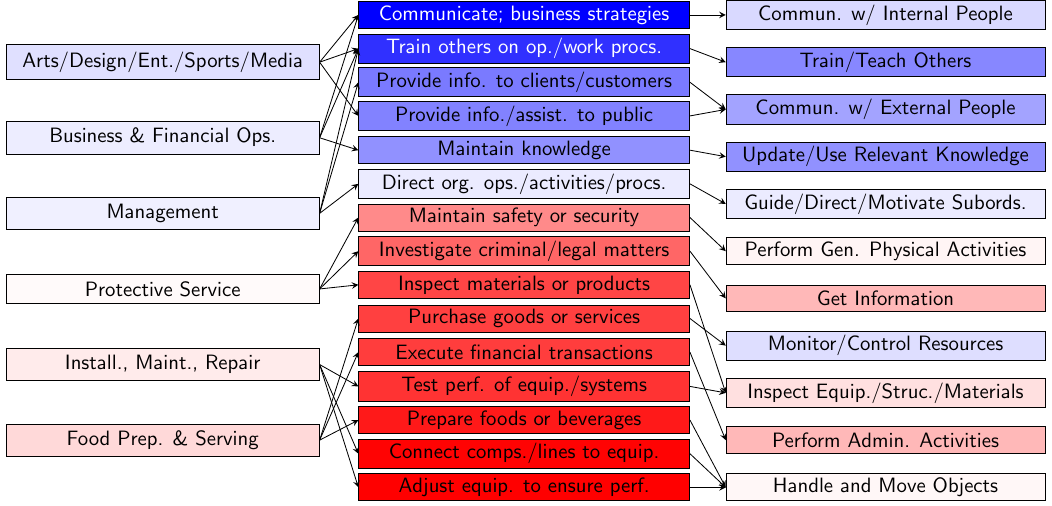}
    \caption{\textbf{IWAs contributing to user goal vs.\ AI action asymmetry.} This tripartite graph shows six prominent occupation groups that are skewed toward user goal or AI action applicability (left), the three IWAs contributing the most to each group's skew (middle) ordered by the extent to which those IWAs skew more to the AI action side, and the GWAs those IWAs belong to (right). Blue indicates a larger skew toward AI action applicability, while red indicates a larger skew toward user goal applicability. White is neutral. To identify which IWAs contribute most to an occupation group’s skew, we take the second component magnitudes in an IWA-level PCA (where we perform PCA on IWA-level scope $\times$ completion $\times$ coverage indicator), multiply them by IWA weights $w_{ij}$ for each occupation, and then take an employment-weighted average to get a score for each occupational-group--IWA pair. IWA-level scores are averaged to get GWA-level scores.}
    \label{fig:tripartite}
\end{figure}

\begin{figure}
    \centering
    \includegraphics[width=\linewidth]{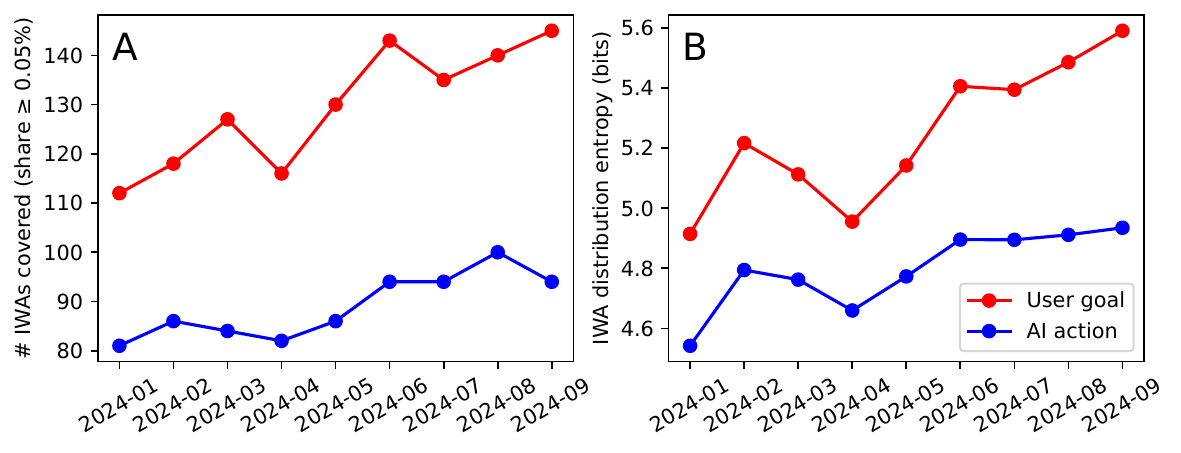}
    \caption{\textbf{Monthly IWA breadth metrics.} (\textbf{A}) Number of IWAs whose activity share crossed the 0.05\% coverage threshold within each month. (\textbf{B}) Entropy of the IWA distribution in each month. Both metrics show an increasing variety in user goals and AI actions over the data period, indicating users performing more varied tasks over time.}
    \label{fig:monthly-coverage-entropy}
\end{figure}

\begin{figure}
    \centering
    \includegraphics[width=0.8\linewidth]{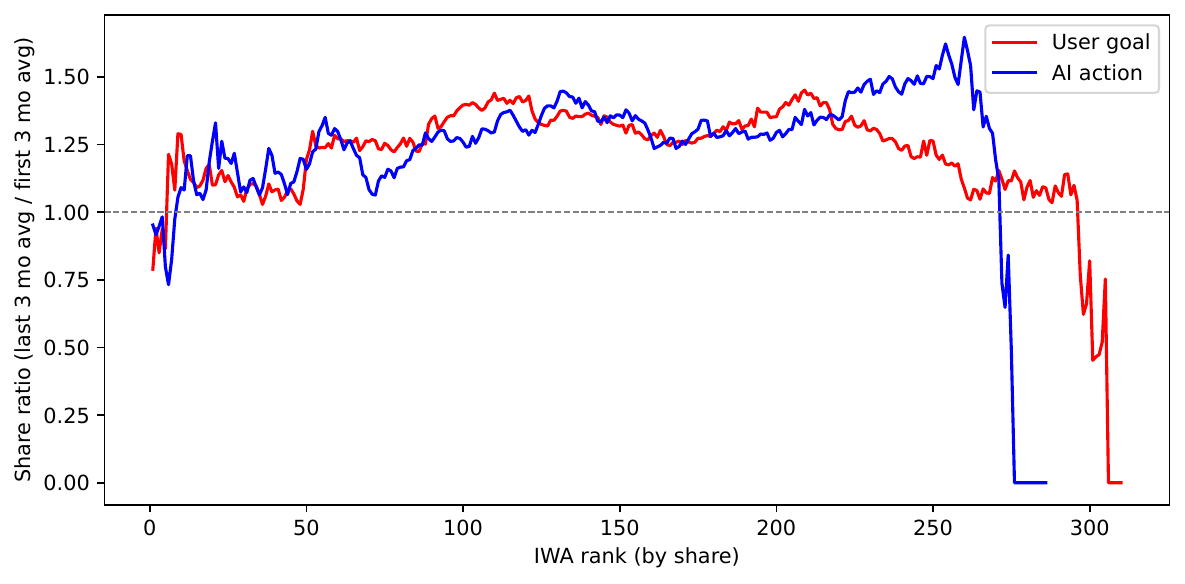}
    \caption{\textbf{Ratio of IWA activity share by rank between last and first three months.} Users performed tail IWAs (ranks $\sim 20 - 250$) more often at the end of the data period than at the beginning, reinforcing the trends from \Cref{fig:monthly-coverage-entropy}.}
    \label{fig:monthly-share-ratio}
\end{figure}

\begin{figure}
    \centering
    \includegraphics[width=\linewidth]{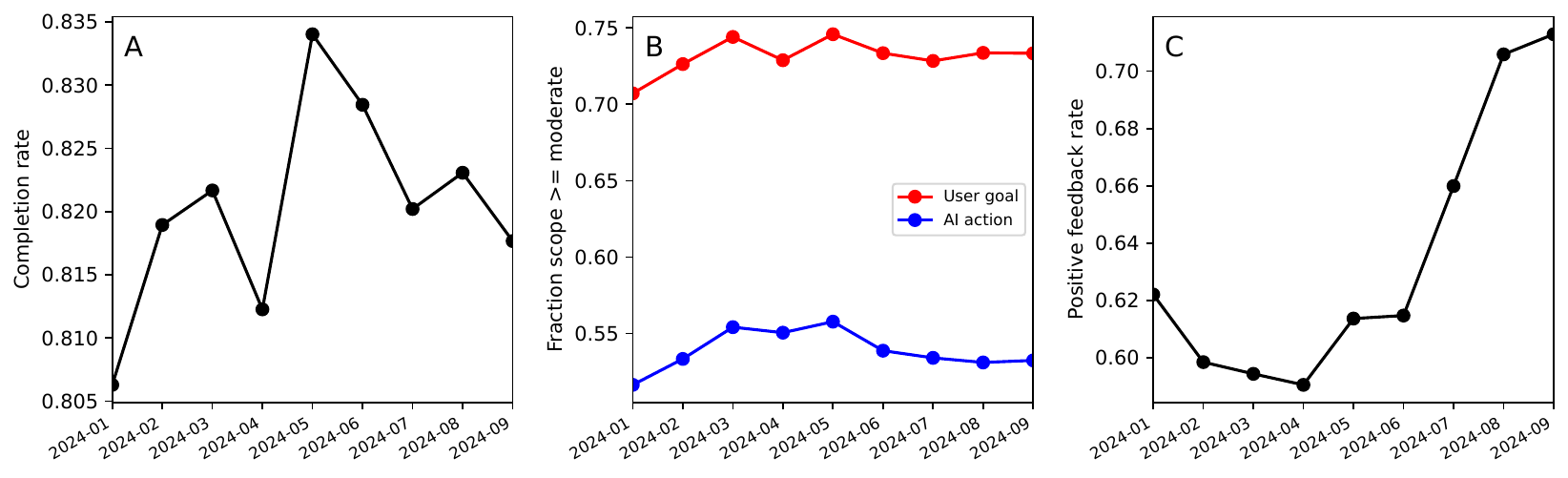}
    \caption{\textbf{Monthly conversation-level metrics.} (\textbf{A}) Fraction of conversations in each month labeled as task completed. (\textbf{B}) Fraction of IWA matches on each side labeled with scope moderate or higher. (\textbf{C}) Share of thumbs feedback that was positive by month.}
    \label{fig:monthly-convo-level}
\end{figure}

\clearpage

\section{Supplementary tables}\label{app:tables}

\begin{table}[htbp]

\centering

\begin{threeparttable}
    
\caption{\textbf{Work activities with largest ratios between user goal and AI action activity share.}}\label{tab:user-bot-ratios}

    \footnotesize
\begin{tabular}{ll}
\toprule
 \bfseries More often assisted by AI                                                                     & \bfseries More often performed by AI                                                       \\
\midrule
Purchase goods or services. (118.4x)                                                & Train others on operational or work procedures. (17.9x)                                 \\
 Execute financial transactions. (58.8x)                                             & Train others to use equipment or products. (16.0x)                                      \\
 Perform athletic activities. (47.3x) & Distribute materials, supplies, or resources. (11.2x)                                   \\
 Obtain information about goods or services. (25.9x)                                 & Train others on health or medical topics. (11.2x)                                       \\
 Research healthcare issues. (20.5x)                                                 & Provide general assistance to others. (10.9x) \\
 Prepare foods or beverages. (14.7x)                                                 & Coach others. (10.6x)                                                                   \\
 Research technology designs or applications. (13.5x)                                & Provide information to guests, clients, or customers. (8.6x)                            \\
 Obtain formal documentation or authorization. (12.5x)                               & Advise others on workplace health or safety issues. (7.5x)                              \\
 Operate office equipment. (11.4x)                                                   & Teach academic or vocational subjects. (6.6x)                                           \\
 Investigate incidents or accidents. (11.3x)                                         & Teach safety procedures or standards to others. (6.5x)\\
\bottomrule
\end{tabular}
\begin{tablenotes}
    \item \scriptsize \rmfamily Numbers show IWA over-representation factors.  Only includes IWAs with user or AI activity share  $\ge 0.05\%$
\end{tablenotes}
\end{threeparttable}

\end{table}

\begin{table}[htbp]
\centering

\begin{threeparttable}

\caption{\textbf{SOC major groups by AI applicability score.}}\label{tab:soc-major}
\small

\begin{tabular}{lrrrrl}
\toprule
\bfseries Major Group                                                &  \bfseries Coverage & \bfseries  Completion &  \bfseries Scope &  \bfseries Score &\bfseries Employment   \\
\midrule
 Computer \& Mathematical        &       0.64 &         0.86 &    0.48 &    0.29 & 5,177,390    \\
 Sales \& Related                &       0.56 &         0.89 &    0.51 &    0.29 & 13,266,370   \\
 Office \& Admin. Supp.          &       0.56 &         0.89 &    0.49 &    0.26 & 18,163,760   \\
 Community \& Social Service     &       0.51 &         0.88 &    0.44 &    0.25 & 2,216,930    \\
 Arts/Design/Ent./Sports/Media  &       0.59 &         0.80 &    0.49 &    0.24 & 2,039,830    \\
 Business \& Financial Ops.      &       0.49 &         0.89 &    0.47 &    0.23 & 10,087,850   \\
 Architecture \& Engineering     &       0.49 &         0.84 &    0.46 &    0.22 & 2,523,090    \\
 Edu. Instruction \& Library     &       0.46 &         0.89 &    0.46 &    0.21 & 8,328,920    \\
 Life, Physical, Social Science &       0.39 &         0.88 &    0.46 &    0.19 & 1,381,930    \\
 Personal Care \& Service        &       0.39 &         0.90 &    0.45 &    0.18 & 2,959,620    \\
 Food Prep. \& Serving           &       0.32 &         0.91 &    0.43 &    0.17 & 13,142,870   \\
 Management                     &       0.27 &         0.90 &    0.45 &    0.13 & 10,445,050   \\
 Legal                          &       0.33 &         0.89 &    0.42 &    0.13 & 1,196,870    \\
 Protective Service             &       0.33 &         0.84 &    0.40 &    0.12 & 3,484,710    \\
 Healthcare Prac. \& Tech.       &       0.25 &         0.91 &    0.39 &    0.12 & 9,251,930    \\
 Production                     &       0.23 &         0.91 &    0.41 &    0.11 & 8,419,460    \\
 Install., Maint., Repair       &       0.22 &         0.92 &    0.41 &    0.10 & 5,979,150    \\
 Transport. \& Material Moving   &       0.21 &         0.92 &    0.38 &    0.10 & 13,664,940   \\
 Building/Grounds Clean./Maint. &       0.15 &         0.94 &    0.38 &    0.08 & 4,403,350    \\
 Construction \& Extraction      &       0.16 &         0.92 &    0.40 &    0.07 & 6,188,720    \\
 Farming, Fishing, \& Forestry   &       0.11 &         0.92 &    0.39 &    0.06 & 422,740      \\
 Healthcare Support             &       0.13 &         0.90 &    0.38 &    0.05 & 7,063,540    \\
\bottomrule
\end{tabular}
\begin{tablenotes}
    \item \scriptsize Coverage, completion, scope, and score are first calculated at the occupation-level and then averaged across occupations in the major group, weighted by employment; each is the mean of user goal and AI action. Coverage, completion and scope are all importance- and relevance-weighted averages of the metrics for IWAs done by an occupation. Coverage is whether the IWA has activity share of  at least $.05\%$. Completion is the share of conversations in an IWA where the LLM completed the user's task. Scope is the share of conversations where the scope was at least moderate. Score is the AI applicability score (see~\Cref{eq:ai-score}). Employment is the total employment for occupations in that group in 2023 OEWS data. 
\end{tablenotes}

\end{threeparttable}

\end{table}

\begin{table}[htbp]
\caption{\textbf{Top 40 occupations with highest AI applicability score.} See note in \Cref{tab:soc-major} for column definitions.}
\label{tab:top-occupations}
\centering
\scriptsize
\begin{tabular}{lrrrrr}
\toprule
 \bfseries Job Title (Abbrv.)                                           &   \bfseries Coverage &   \bfseries Completion &  \bfseries Scope &   \bfseries Score\ & \bfseries Employment   \\
\midrule
Interpreters and Translators                                                                     &      0.980 &        0.883 &          0.568 &   0.492 & 51,560       \\
 Historians                                                                                       &      0.906 &        0.852 &          0.564 &   0.462 & 3,040        \\
 Writers and Authors                                                                              &      0.850 &        0.841 &          0.603 &   0.454 & 49,450       \\
 Sales Representatives of Services &      0.842 &        0.902 &          0.573 &   0.449 & 1,142,020    \\
 Computer Numerically Controlled Tool Programmers                                                 &      0.899 &        0.869 &          0.528 &   0.419 & 28,030       \\
 Broadcast Announcers and Radio Disc Jockeys                                                      &      0.745 &        0.841 &          0.605 &   0.409 & 25,070       \\
 Customer Service Representatives                                                                 &      0.720 &        0.901 &          0.591 &   0.408 & 2,858,710    \\
 Telemarketers                                                                                    &      0.656 &        0.893 &          0.603 &   0.404 & 81,580       \\
 Political Scientists                                                                             &      0.774 &        0.873 &          0.528 &   0.391 & 5,580        \\
 Mathematicians                                                                                   &      0.905 &        0.744 &          0.538 &   0.386 & 2,220        \\
 News Analysts, Reporters, and Journalists                                                        &      0.815 &        0.811 &          0.560 &   0.383 & 45,020       \\
 Passenger Attendants                                                                             &      0.798 &        0.875 &          0.616 &   0.376 & 20,190       \\
 Technical Writers                                                                                &      0.826 &        0.816 &          0.538 &   0.373 & 47,970       \\
 Concierges                                                                                       &      0.699 &        0.880 &          0.556 &   0.372 & 41,020       \\
 Proofreaders and Copy Markers                                                                    &      0.911 &        0.860 &          0.486 &   0.369 & 5,490        \\
 Editors                                                                                          &      0.780 &        0.818 &          0.537 &   0.367 & 95,700       \\
 Business Teachers, Postsecondary                                                                 &      0.700 &        0.904 &          0.517 &   0.367 & 82,980       \\
 Public Relations Specialists                                                                     &      0.627 &        0.900 &          0.600 &   0.365 & 275,550      \\
 Data Scientists                                                                                  &      0.765 &        0.863 &          0.506 &   0.357 & 192,710      \\
 Personal Financial Advisors                                                                      &      0.689 &        0.875 &          0.524 &   0.355 & 272,190      \\
 Web Developers                                                                                   &      0.733 &        0.858 &          0.506 &   0.353 & 85,350       \\
 Advertising Sales Agents                                                                         &      0.660 &        0.901 &          0.526 &   0.353 & 108,100      \\
 Management Analysts                                                                              &      0.676 &        0.896 &          0.540 &   0.353 & 838,140      \\
 Geographers                                                                                      &      0.771 &        0.828 &          0.478 &   0.352 & 1,460        \\
 Brokerage Clerks                                                                                 &      0.742 &        0.892 &          0.574 &   0.350 & 48,060       \\
 Market Research Analysts and Marketing Specialists                                               &      0.707 &        0.897 &          0.517 &   0.350 & 846,370      \\
 Economics Teachers, Postsecondary                                                                &      0.676 &        0.899 &          0.512 &   0.349 & 12,210       \\
 Public Safety Telecommunicators                                                                  &      0.660 &        0.878 &          0.534 &   0.346 & 97,820       \\
 Counter and Rental Clerks                                                                        &      0.622 &        0.900 &          0.523 &   0.344 & 390,300      \\
 Telephone Operators                                                                              &      0.804 &        0.859 &          0.567 &   0.342 & 4,600        \\
 Library Science Teachers, Postsecondary                                                          &      0.654 &        0.904 &          0.512 &   0.341 & 4,220        \\
 Tax Examiners and Collectors, and Revenue Agents                                                 &      0.534 &        0.889 &          0.557 &   0.340 & 50,250       \\
 Political Science Teachers, Postsecondary                                                        &      0.653 &        0.898 &          0.503 &   0.339 & 17,090       \\
 Philosophy and Religion Teachers, Postsecondary                                                  &      0.654 &        0.897 &          0.509 &   0.338 & 20,320       \\
 Models                                                                                           &      0.642 &        0.888 &          0.533 &   0.337 & 3,090        \\
 Mathematical Science Occupations, All Other                                                      &      0.646 &        0.882 &          0.527 &   0.336 & 4,320        \\
 Computer User Support Specialists                                                                &      0.720 &        0.875 &          0.475 &   0.334 & 689,700      \\
 Criminal Justice and Law Enforcement Teachers, Postsecondary                                     &      0.648 &        0.898 &          0.506 &   0.334 & 13,390       \\
 Child, Family, and School Social Workers                                                         &      0.691 &        0.878 &          0.469 &   0.334 & 352,160      \\
 Foreign Language and Literature Teachers, Postsecondary                                          &      0.638 &        0.899 &          0.509 &   0.333 & 20,820       \\
\bottomrule
\end{tabular}
\end{table}

\begin{table}[htbp]\renewcommand{\arraystretch}{0.5} 
\caption{\textbf{Bottom 40 occupations with lowest AI applicability score.} See note in \Cref{tab:soc-major} for column definitions}
\label{tab:bottom-occupations}
\centering
\scriptsize
\begin{tabular}{lrrrrr}
\toprule
 \bfseries Job Title (Abbrv.)                                           &   \bfseries Coverage &   \bfseries Cmpltn. &   \bfseries Scope &   \bfseries Score & \bfseries Empl.   \\
\midrule
 Helpers--Extraction Workers                                    &      0.089 &        0.956 &          0.336 &   0.028 & 7,360        \\
 Helpers--Painters, Paperhangers, Plasterers, and Stucco Masons &      0.042 &        0.957 &          0.376 &   0.027 & 7,700        \\
 Plant and System Operators, All Other                          &      0.053 &        0.927 &          0.375 &   0.026 & 15,370       \\
 Oral and Maxillofacial Surgeons                                &      0.053 &        0.894 &          0.340 &   0.026 & 4,160        \\
 Embalmers                                                      &      0.066 &        0.545 &          0.220 &   0.026 & 3,380        \\
 Automotive Glass Installers and Repairers                      &      0.045 &        0.930 &          0.342 &   0.025 & 16,890       \\
 Ship Engineers                                                 &      0.050 &        0.918 &          0.386 &   0.025 & 8,860        \\
 Tire Repairers and Changers                                    &      0.044 &        0.947 &          0.352 &   0.023 & 101,520      \\
 Phlebotomists                                                  &      0.057 &        0.952 &          0.286 &   0.022 & 137,080      \\
 Hazardous Materials Removal Workers                            &      0.045 &        0.953 &          0.347 &   0.022 & 49,960       \\
 Helpers--Production Workers                                    &      0.037 &        0.925 &          0.362 &   0.021 & 181,810      \\
 Highway Maintenance Workers                                    &      0.032 &        0.963 &          0.319 &   0.021 & 150,860      \\
 Medical Equipment Preparers                                    &      0.038 &        0.961 &          0.309 &   0.021 & 66,790       \\
 Packaging and Filling Machine Operators and Tenders            &      0.037 &        0.914 &          0.389 &   0.020 & 371,600      \\
 Machine Feeders and Offbearers                                 &      0.046 &        0.889 &          0.360 &   0.019 & 44,500       \\
 Dishwashers                                                    &      0.029 &        0.947 &          0.301 &   0.018 & 463,940      \\
 Cement Masons and Concrete Finishers                           &      0.033 &        0.921 &          0.391 &   0.015 & 203,560      \\
 Prosthodontists                                                &      0.099 &        0.896 &          0.288 &   0.015 & 570          \\
 Industrial Truck and Tractor Operators                         &      0.031 &        0.944 &          0.281 &   0.013 & 778,920      \\
 Massage Therapists                                             &      0.102 &        0.908 &          0.321 &   0.012 & 92,650       \\
 Surgical Assistants                                            &      0.029 &        0.784 &          0.288 &   0.012 & 18,780       \\
 Tire Builders                                                  &      0.028 &        0.927 &          0.399 &   0.012 & 20,660       \\
 Helpers--Roofers                                               &      0.016 &        0.943 &          0.368 &   0.010 & 4,540        \\
 Gas Compressor and Gas Pumping Station Operators               &      0.015 &        0.957 &          0.472 &   0.010 & 4,400        \\
 Roofers                                                        &      0.018 &        0.942 &          0.376 &   0.009 & 135,140      \\
 First-Line Supervisors of Firefighting and Prevention Workers  &      0.040 &        0.883 &          0.385 &   0.009 & 84,120       \\
 Roustabouts, Oil and Gas                                       &      0.013 &        0.952 &          0.392 &   0.009 & 43,830       \\
 Maids and Housekeeping Cleaners                                &      0.022 &        0.935 &          0.336 &   0.008 & 836,230      \\
 Ophthalmic Medical Technicians                                 &      0.037 &        0.890 &          0.329 &   0.007 & 73,390       \\
 Paving, Surfacing, and Tamping Equipment Operators             &      0.009 &        0.958 &          0.293 &   0.007 & 43,080       \\
 Logging Equipment Operators                                    &      0.012 &        0.955 &          0.364 &   0.005 & 23,720       \\
 Motorboat Operators                                            &      0.007 &        0.934 &          0.387 &   0.003 & 2,710        \\
 Orderlies                                                      &      0.000 &        0.761 &          0.181 &   0.000 & 48,710       \\
 Floor Sanders and Finishers                                    &      0.000 &        0.937 &          0.337 &   0.000 & 5,070        \\
 Pile Driver Operators                                          &      0.000 &        0.976 &          0.236 &   0.000 & 3,010        \\
 Rail-Track Laying and Maintenance Equipment Operators          &      0.000 &        0.962 &          0.271 &   0.000 & 18,770       \\
 Foundry Mold and Coremakers                                    &      0.000 &        0.952 &          0.361 &   0.000 & 11,780       \\
 Water and Wastewater Treatment Plant and System Operators      &      0.000 &        0.919 &          0.439 &   0.000 & 120,710      \\
 Bridge and Lock Tenders                                        &      0.000 &        0.931 &          0.386 &   0.000 & 3,460        \\
 Dredge Operators                                               &      0.000 &        0.986 &          0.223 &   0.000 & 940          \\
\bottomrule
\end{tabular}
\end{table}

\begin{table}[htbp]
    \centering

\begin{threeparttable}
    
{
    \caption{\textbf{Occupations with the largest difference in user goal and AI action applicability score percentiles.}}    \label{tab:user-ai-diff-occs}
\scriptsize
    \begin{tabular}{ll}
\toprule
\bfseries AI assistance, not performance                     &  \bfseries AI performance, not assistance                          \\
\midrule
Cooks, Fast Food (83, 6)                                                          & Exercise Trainers and Group Fitness Instructors (17, 75)                     \\
 Meat  Cutters and Trimmers (79, 6)                              & Choreographers (34, 81)                                                      \\
 Butchers and Meat Cutters (83, 11)                                                & Training and Development Managers (45, 88)                                   \\
 Animal Breeders (76, 6)                                                           & Coaches and Scouts (43, 80)                                                  \\
 Cooks, Private Household (97, 32)                                                 & Actuaries (42, 76)                                                           \\
 Cooks, Restaurant (76, 12)                                                        & Environmental Engineers (55, 88)                                             \\
 Photographers (95, 33)                                                            & Human Resources Specialists (53, 86)                                         \\
 Archivists (98, 45)                                                               & First-Line Supervisors of Non-Retail Sales Workers (48, 77)                  \\
 Lighting Technicians (82, 35)                                                     & First-Line Sups. of Office and Admin.\ Supp.\ Workers (66, 92) \\
 Door-to-Door Sales Workers and Related (94, 49) & Air Traffic Controllers (50, 75)                                             \\
\bottomrule
\end{tabular}}
\begin{tablenotes}
    \item \scriptsize \rmfamily This Table shows the 10 occupations on each side with the largest difference in their AI applicability score percentile computed from user goals and AI actions, filtered for occupations in the top quartile on their higher-ranked side. Occupation title abbreviated. Numbers in parentheses are (user goal AI applicability score percentile, AI action applicability score percentile). The occupations on the left focus on physical occupations involving cooking and working with animals, while many of the occupations on the right involve teaching, training, or coaching.
\end{tablenotes}

\end{threeparttable}

\end{table}

\clearpage
\newpage
\section*{NeurIPS Paper Checklist}

\begin{enumerate}

\item {\bf Claims}
    \item[] Question: Do the main claims made in the abstract and introduction accurately reflect the paper's contributions and scope?
    \item[] Answer: \answerYes{} %
    \item[] Justification: We provide clear descriptions of our data and the extent of our results, without over-claiming.

\item {\bf Limitations}
    \item[] Question: Does the paper discuss the limitations of the work performed by the authors?
    \item[] Answer: \answerYes{} %
    \item[] Justification: We articulate the limitations of our datasets (consumer data from a single generative AI platform) as well as our classifiers and how our metrics should be interpreted.

\item {\bf Theory assumptions and proofs}
    \item[] Question: For each theoretical result, does the paper provide the full set of assumptions and a complete (and correct) proof?
    \item[] Answer: \answerNA{} %
    \item[] Justification: We make no theoretical claims.

    \item {\bf Experimental result reproducibility}
    \item[] Question: Does the paper fully disclose all the information needed to reproduce the main experimental results of the paper to the extent that it affects the main claims and/or conclusions of the paper (regardless of whether the code and data are provided or not)?
    \item[] Answer: \answerYes{} %
    \item[] Justification: We provide all prompts and model details used for classification, as well as a detailed description of our classification pipeline. This would enable reproduction of our methods on a different dataset (we cannot release the Copilot conversation texts due to legal and privacy obligations).

\item {\bf Open access to data and code}
    \item[] Question: Does the paper provide open access to the data and code, with sufficient instructions to faithfully reproduce the main experimental results, as described in supplemental material?
    \item[] Answer: \answerNo{} %
    \item[] Justification: The raw conversation dataset we analyze is highly sensitive and cannot be released due to legal and privacy obligations. We have released all IWA- and occupation-level metrics at \url{https://github.com/microsoft/working-with-ai}. 

\item {\bf Experimental setting/details}
    \item[] Question: Does the paper specify all the training and test details (e.g., data splits, hyperparameters, how they were chosen, type of optimizer) necessary to understand the results?
    \item[] Answer: \answerNA{} %
    \item[] Justification: The paper does not include ML experiments in the traditional sense, but we provide all details of the models and parameters used for classification.  

\item {\bf Experiment statistical significance}
    \item[] Question: Does the paper report error bars suitably and correctly defined or other appropriate information about the statistical significance of the experiments?
    \item[] Answer: \answerYes{} %
    \item[] Justification: Our plots include error bars or other indications of variance where appropriate.

\item {\bf Experiments compute resources}
    \item[] Question: For each experiment, does the paper provide sufficient information on the computer resources (type of compute workers, memory, time of execution) needed to reproduce the experiments?
    \item[] Answer: \answerNA{} %
    \item[] Justification: As we cannot release the data, reproducing our classifications on our data is not possible. 
    
\item {\bf Code of ethics}
    \item[] Question: Does the research conducted in the paper conform, in every respect, with the NeurIPS Code of Ethics \url{https://neurips.cc/public/EthicsGuidelines}?
    \item[] Answer: \answerYes{} %
    \item[] Justification: We received IRB approval for our use of human data and  ensured we maintained user privacy in our analysis.

\item {\bf Broader impacts}
    \item[] Question: Does the paper discuss both potential positive societal impacts and negative societal impacts of the work performed?
    \item[] Answer: \answerYes{} %
    \item[] Justification: We discuss broader impacts in \Cref{app:broader-impacts}.
    
\item {\bf Safeguards}
    \item[] Question: Does the paper describe safeguards that have been put in place for responsible release of data or models that have a high risk for misuse (e.g., pre-trained language models, image generators, or scraped datasets)?
    \item[] Answer: \answerNA{} %
    \item[] Justification: We are not releasing models or data with high risk of misuse.

\item {\bf Licenses for existing assets}
    \item[] Question: Are the creators or original owners of assets (e.g., code, data, models), used in the paper, properly credited and are the license and terms of use explicitly mentioned and properly respected?
    \item[] Answer: \answerYes{} %
    \item[] Justification: We credit the creators of the models we use for classification. Our metrics are released under a CC BY 4.0 license at \url{https://github.com/microsoft/working-with-ai}.

\item {\bf New assets}
    \item[] Question: Are new assets introduced in the paper well documented and is the documentation provided alongside the assets?
    \item[] Answer: \answerYes{} %
    \item[] Justification: Our GitHub data release includes detailed documentation (\url{https://github.com/microsoft/working-with-ai}).

\item {\bf Crowdsourcing and research with human subjects}
    \item[] Question: For crowdsourcing experiments and research with human subjects, does the paper include the full text of instructions given to participants and screenshots, if applicable, as well as details about compensation (if any)? 
    \item[] Answer: \answerNA{} %
    \item[] Justification: Our data do not involve human subjects in the laboratory sense. Copilot data was used in accordance with the Microsoft Privacy Statement and IRB approval. Annotators for validating classifiers were authors.

\item {\bf Institutional review board (IRB) approvals or equivalent for research with human subjects}
    \item[] Question: Does the paper describe potential risks incurred by study participants, whether such risks were disclosed to the subjects, and whether Institutional Review Board (IRB) approvals (or an equivalent approval/review based on the requirements of your country or institution) were obtained?
    \item[] Answer: \answerYes{} %
    \item[] Justification: We state in the paper that we received IRB approval for our use of  human data. More detail is provided in \Cref{app:broader-impacts}.

\item {\bf Declaration of LLM usage}
    \item[] Question: Does the paper describe the usage of LLMs if it is an important, original, or non-standard component of the core methods in this research? Note that if the LLM is used only for writing, editing, or formatting purposes and does \emph{not} impact the core methodology, scientific rigor, or originality of the research, declaration is not required.
    \item[] Answer: \answerYes{} %
    \item[] Justification: LLMs are a vital part of our classification approach, which we describe in detail.

\end{enumerate}

\end{document}